\documentclass{article}

\usepackage[preprint]{neurips_2026}

\usepackage[utf8]{inputenc}
\usepackage[T1]{fontenc}
\usepackage{graphicx}
\usepackage{float}
\usepackage{booktabs}
\usepackage{longtable}
\usepackage{amsmath}
\usepackage{amssymb}
\usepackage{subcaption}
\usepackage{hyperref}
\usepackage{url}
\usepackage{amsfonts}
\usepackage{nicefrac}
\usepackage{microtype}
\usepackage{xcolor}
\usepackage{tikz}
\usetikzlibrary{arrows.meta, positioning, calc, fit, backgrounds}

\title{A New Framework to Analyse the Distributional Robustness of Deep Neural Networks}

\author{
  Divij Khaitan\thanks{Work done while at Ashoka University} \\
  Microsoft \\
  Bengaluru, India \\
  \texttt{t-dkhaitan@microsoft.com} \\
  \And
  Subhashis Banerjee \\
  Ashoka University \\
  Sonipat, India \\
  \texttt{suban@ashoka.edu.in}
}
\begin{document}

\maketitle

\begin{abstract}
        Deep neural networks have achieved impressive performance on a variety of tasks, but their brittleness to distributional shifts remains a significant barrier to real-world deployment. In this paper, we propose a framework to analyse and quantify the distributional robustness of neural networks by studying the interactions between layer weights and activations. We model these interactions using Bernoulli distributions, using the separation between classes as a diagnostic proxy for robustness. We demonstrate the usefulness of this framework through models trained on CIFAR-10 and ImageNet. We show that our proposed metrics can distinguish between networks that have memorised their training data and those that have not. We also perform analogous experiments in the activation space and find that the same properties do not hold up. Additionally, we investigate the behaviour of our metrics under various distribution shifts and show that these shifts reduce separation under our path-based diagnostics. Our results suggest that this framework provides useful model-level diagnostics of representation structure and robustness. 
\end{abstract}

\section{Introduction}
        Despite the impressive performance of deep neural networks (DNNs) on various tasks, their brittleness to shifts in distribution remains a significant barrier to their real-world deployment. Adversarial examples are perhaps the most extreme form of this, where input perturbations imperceptible to humans trigger misclassifications. While much work has gone into the detection of distribution shifts at a sample level, a principled means of analysing distributional robustness remains an open problem. The goal of this paper is to propose a diagnostic framework that allows us to understand and quantify the distributional robustness of neural networks; it is not intended as a replacement for standalone OOD detectors.
    \section{Related Work}
        \begin{enumerate}
            \item 
                Adversarial Robustness: Adversarial Examples were first identified by Szegedy et al. \cite{szegedy2014intriguingpropertiesneuralnetworks}, who showed that imperceptible perturbations to images could cause misclassifications. Since then, a plethora of attack and defense mechanisms have been proposed \cite{goodfellow2015explainingharnessingadversarialexamples, madry2019deeplearningmodelsresistant}. Additionally, there has been some work on identifying the causes of adversarial vulnerability, for example  \cite{shafahi2020adversarialexamplesinevitable} argues, using the concentration of measure in high dimensional spaces, that every classification task has a fundamental limit on the robustness that it can achieve. However, it does not prove that they are always easy to find. It also suggests that allowing models to reject predictions may be strong workaround to adversarial examples. Additionally, there is a large body of literature discussing various forms of adversarial attacks \cite{madry2019deeplearningmodelsresistant, moosavi2016deepfool,goodfellow2015explainingharnessingadversarialexamples} and defenses against them \cite{raghunathan2018certified, hayase2021spectre, 10.5555/3327757.3327896}. 
            \item 
                Out of Distribution Detection: This is the problem that tries to detect whether the given input is from a different distribution than the training data. These techniques can broadly be classified into post-hoc and training-based methods. Post-hoc methods involve analysing or modifying the outputs of a trained network, hypothesising that OOD samples have some characteristic difference from in-distribution samples, such as lower output probabilities \cite{hendrycks2018baselinedetectingmisclassifiedoutofdistribution}, a large distance from the distributions of all the classes \cite{lee2018simpleunifiedframeworkdetecting}, or different feature representations \cite{sun2021reactoutofdistributiondetectionrectified}. Training based methods involve a post-training step with a modified loss function to encourage the model to discriminate between in-distribution and OOD samples \cite{sharifi2024gradientregularizedoutofdistributiondetection}. A fundamental limitation of training-based methods is the need for large collections of OOD data, which may be a hurdle in practice.
            \item 
                Interpretability: A variety of methods have been proposed to assign human-understandable semantics to the latent spaces and predictions made by DNNs. On the side of interpretability, SUMMIT \cite{hohman2019summitscalingdeeplearning} and NAP \cite{bauerle2022neuralactivationpatternsnaps} both attempt to identify important neurons in an attempt to interpret the features associated with different classes. \cite{stano2020explainingpredictionsdeepneural} fits a mixture of Gaussians to each neuron independently in selected layers with the explicit goal of attributing predictions to neurons. They use at most 2 Gaussians. Similar ideas have been used for the detection of rare classes \cite{paterson2021detectionmitigationraresubclasses} as well as the detection of out-of-distribution inputs \cite{olber2023detectionoutofdistributionsamplesusing}. \cite{simonyan2014deepinsideconvolutionalnetworks, faaaf4beb6f24ca599a7ecae54e0c724, Selvaraju_2019, gautam2021lookslikethatenhancing} focus on mapping each input pixel to an importance value, while \cite{ribeiro2016whyitrustyou, lundberg2017unifiedapproachinterpretingmodel} focus on importance of each feature locally around a particular datapoint. \cite{gautam2024prototypicalselfexplainablemodelsretraining} uses clustering in the embedding space alongside PRP \cite{gautam2021lookslikethatenhancing} characterise the prediction of any model in terms of its similarity to a set of prototypes - images which are representative of the rest of the class - which is obtained by clustering embeddings in the latent space. Our work is slightly different in that we attempt to characterise the distribution of activations of each neuron across all classes, and use this to understand how different classes relate to each other.
            \item 
                Domain Adaptation: This is the problem of adapting a model trained on one distribution to perform well on another distribution. This is closely related to the problem of distributional robustness, as a model that is robust to distribution shifts should be able to adapt to new domains with minimal performance degradation. Additionally, methods from this area often involve measuring distances between distributions in the latent space, such as maximum mean discrepancy, Wasserstein distance and energy distance. The major differentiator between our work and this area is that often the goal is to map the output distribution of the new data given to the model to the output distribution of the original data, which is often highly data intensive. Our work is more focused on understanding the structure of the latent space and how OOD data relates to it, which is a more lightweight approach that provides an additional advantage of being a much richer descriptor of the distributional robustness of the model compared to the cross entropy loss or traditional accuracy metrics such as precision and recall.
        \end{enumerate}
    \section{Methodology}
        \subsection{Neuron-Weight Interactions}
            \begin{figure}[t]
                \centering
                \input{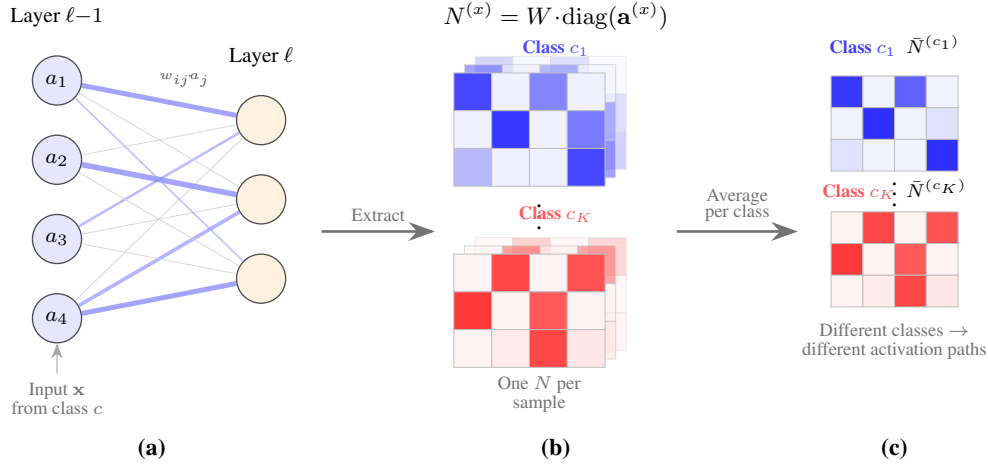}
                \caption{Overview of the proposed framework. \textbf{(a)}~Two consecutive layers of a neural network; line thickness encodes the magnitude of each weight--activation product $w_{ij}\cdot a_j$. \textbf{(b)}~The extracted interaction matrix $N = W\cdot\mathrm{diag}(\mathbf{a})$, whose entries capture per-connection contributions. \textbf{(c)}~Different classes produce different $N$ matrices, reflecting class-specific activation paths through the network.}
                \label{fig:pipeline_overview}
            \end{figure}
            In this paper, we study the interactions between the weights and activations of neurons in a trained neural network. Consider a neural network with $L$ layers, where each layer $l$ has weights $W^{(l)}$ and biases $b^{(l)}$. For an input sample $a$, the activation of layer $l$ is given by: 
            \[a^{(l)} = f(W^{(l)} a^{(l-1)} + b^{(l)})\] 
            where $f$ is the activation function (e.g., ReLU, sigmoid). We focus on the pre-activation values (i.e., before applying the activation function) to capture the raw interactions between weights and inputs. Mathematically, the matrix multiplication at layer $l$ can be expressed as:
            \[
            \begin{bmatrix}
                    w_{11} & w_{12} & \cdots & w_{1n} \\
                    w_{21} & w_{22} & \cdots & w_{2n} \\
                    \vdots & \vdots & \ddots & \vdots \\
                    w_{m1} & w_{m2} & \cdots & w_{mn}
                \end{bmatrix}
                \begin{bmatrix}
                    a_{1} \\
                    a_{2} \\
                    \vdots \\
                    a_{n}
                \end{bmatrix}
                =
                \begin{bmatrix}
                    \sum_{j=1}^{n} w_{1j} a_{j} \\
                    \sum_{j=1}^{n} w_{2j} a_{j} \\
                    \vdots \\
                    \sum_{j=1}^{n} w_{mj} a_{j}
            \end{bmatrix} 
            \]
            where $w_{ij}$ represents the weight connecting the $j^{th}$ neuron in layer $l-1$ to the $i^{th}$ neuron in layer $l$, and $a_j$ is the activation from the previous layer. Now, factoring out an all 1s vector from the input, we can re-write the above as:
            \[
                N = 
                \begin{bmatrix}
                    w_{11} & w_{12} & \cdots & w_{1n} \\
                    w_{21} & w_{22} & \cdots & w_{2n} \\
                    \vdots & \vdots & \ddots & \vdots \\
                    w_{m1} & w_{m2} & \cdots & w_{mn}
                \end{bmatrix}
                \begin{bmatrix}
                    a_{1} & 0 & \cdots & 0 \\
                    0 & a_2 & \cdots & 0 \\
                    \vdots & \vdots & \ddots & \vdots \\
                    0 & 0 & \cdots & a_{n}
                \end{bmatrix}
                =
                \begin{bmatrix}
                    w_{11} a_1 & w_{12} a_2 & \cdots & w_{1n} a_n \\
                    w_{21} a_1 & w_{22} a_2 & \cdots & w_{2n} a_n \\
                    \vdots & \vdots & \ddots & \vdots \\
                    w_{m1} a_1 & w_{m2} a_2 & \cdots & w_{mn} a_n
                \end{bmatrix}
            \]
            This formulation is a much more granular view of the interactions between weights and activations, allowing us to study how each weight contributes to the final pre-activation value based on the input activations. We hypothesise that classes take similar paths through the network. From these, a significance matrix can be constructed as such 
            \[
                S_{ij} = I \{|N_{ij}| > n * \sum_{k} N_{ik}\}
            \]
            where $I$ is the indicator function and $n$ is the input size to the layer. For a randomly chosen matrix (such as a layer at initialisation) the significance matrix would be dense and unstructured since the sum is expected to be zero. We expect the model to discover some structure in the data and encode it into the matrix, causing the significance matrix to become sparse. The layer needs a few paths to be associated with any single class, with the rest of the paths being insignificant. However, if the entire layer has significant activations then that would suggest a high degree of entanglement between classes. Additionally, these can be used to associate a collection of Bernoulli distributions with each class 
            \[\mathcal{B}_{ij}(p) = \left \{ \begin{array}{ll}
                S_{ij} = 1 & \text{with probability } p \\
                S_{ij} = 0 & \text{with probability } 1-p
            \end{array} \right \} \]
            which allows us to compare them and possibly make claims about the robustness of the classifier. Large inter-class distances would indicate robustness, as would small intra-class entropy. For the distance measure, we use the coordinate-wise KL divergence between the distributions. Given two Bernoulli distribution matrices $\mathcal{B}^{1}$ and $\mathcal{B}^{2}$, the KL divergence is given by:
            \[ KL(\mathcal{B}^{1} || \mathcal{B}^{2}) = \sum_{(i, j)} \mathcal{B}_{ij}^{1} \log \frac{\mathcal{B}_{ij}^{1}}{\mathcal{B}_{ij}^{2}} + (1 - \mathcal{B}_{ij}^{1}) \log \frac{(1 - \mathcal{B}_{ij}^{1})}{(1 -\mathcal{B}_{ij}^{2})} \]

        \subsection{Memorisation and Shifts in Distribution}
            Zhang et al. \cite{zhang2017understandingdeeplearningrequires} showed that even InceptionV3 is capable of fitting all of ImageNet \cite{russakovsky2015ImageNetlargescalevisual} with its labels shuffled randomly. Memorisation is the most extreme case of overfitting, where the network does not learn any generalisable features and learns statistical artifacts that allow it to map each input to the correct label. In our framework, this should reflect as inter-class distances staying static over the course of training. To this end, we reproduce both experiments from the paper, training a modified version of alexnet on CIFAR10 and InceptionV3 on ImageNet with both normal and shuffled labels. We additionally provide results for the ResNet-50 \cite{he2016deep} and ViT-B/32 \cite{dosovitskiy2020image} architectures on ImageNet, however we omit training them on random labels due to computational constraints.

            Additionally, we want to investigate the behaviour of our proposed metrics under distribution shifts. To this end, we use the ImageNet-r dataset \cite{hendrycks2021facesrobustnesscriticalanalysis}, which contains stylised versions of images corresponding to classes from ImageNet. For CIFAR10, we use the CIFAR10.1 \cite{recht2018cifar10classifiersgeneralizecifar10}, SVHN \cite{37648} and TinyImageNet \cite{Le2015TinyIV} as OOD datasets. SVHN does not have a correspondence with the classes in CIFAR10, and a mapping for TinyImageNet is created as outlined in the Appendix. The class distributions for in and out-of-distribution datasets are compared to analyse and quantify the shifts in the distribution of representations.

            For InceptionV3, ResNet50 and ViT-B/32 the publically available checkpoint on the pytorch hub is used as the trained model, with the InceptionV3 model fitting random labels trained from scratch for 250 epochs using the same hyperparameters outlined in \cite{szegedy2015rethinkinginceptionarchitecturecomputer}. For the CIFAR10 experiments, the modified version of alexnet outlined in \cite{zhang2017understandingdeeplearningrequires} is used, training for 250 epochs with the same hyperparameters.

    \section{Results}
        We report all analyses on the trained and random-label checkpoints described in Section 3.2. Alongside the heatmaps, we provide scalar summaries of separation and entropy changes computed from these existing runs.
        
        \subsection{Results for ImageNet}
\begin{figure}[hbtp]
                \centering
                \begin{subfigure}{0.32\textwidth}
                    \includegraphics[width=\linewidth]{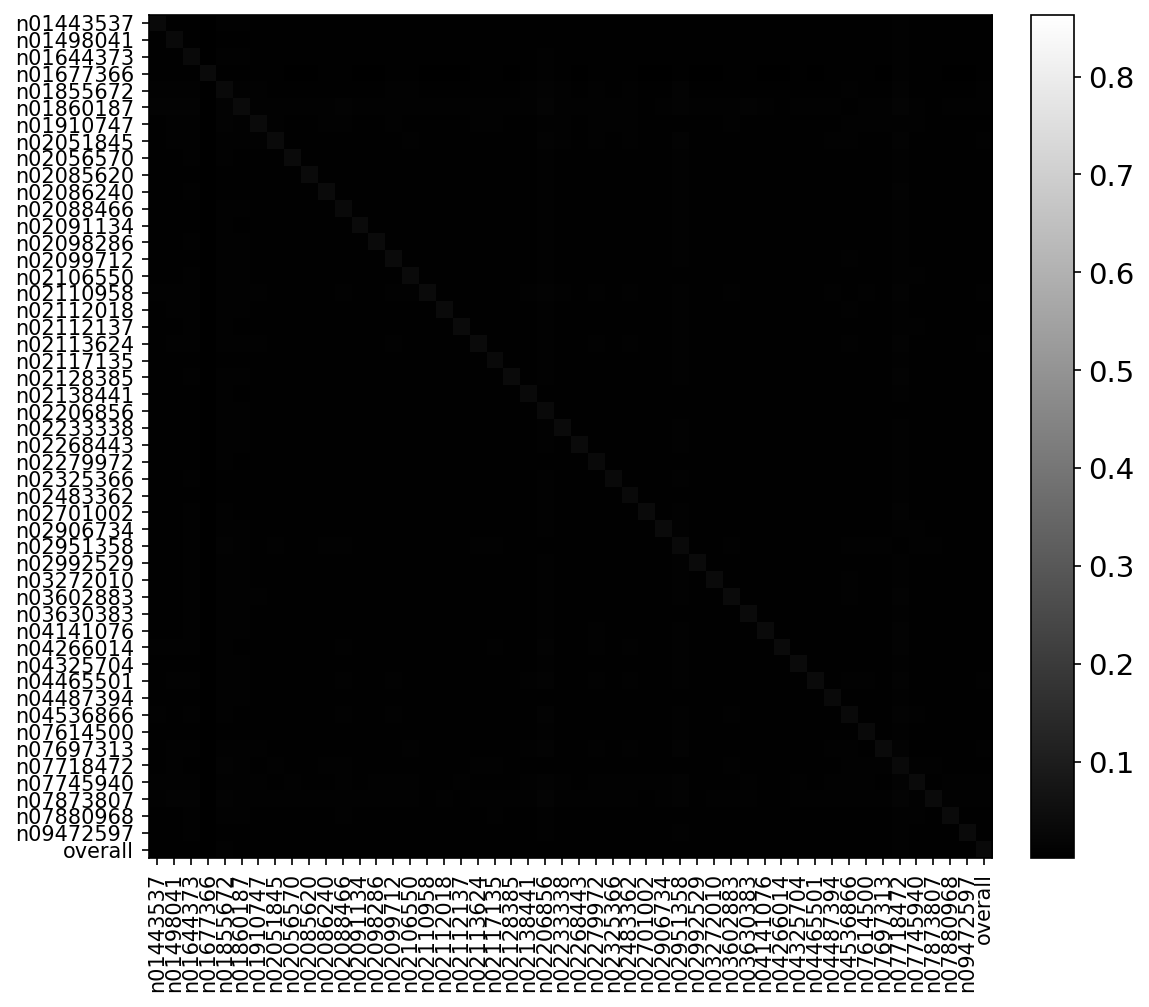}
                    \caption{Initialisation}
                \end{subfigure}
                \begin{subfigure}{0.32\textwidth}
                    \includegraphics[width=\linewidth]{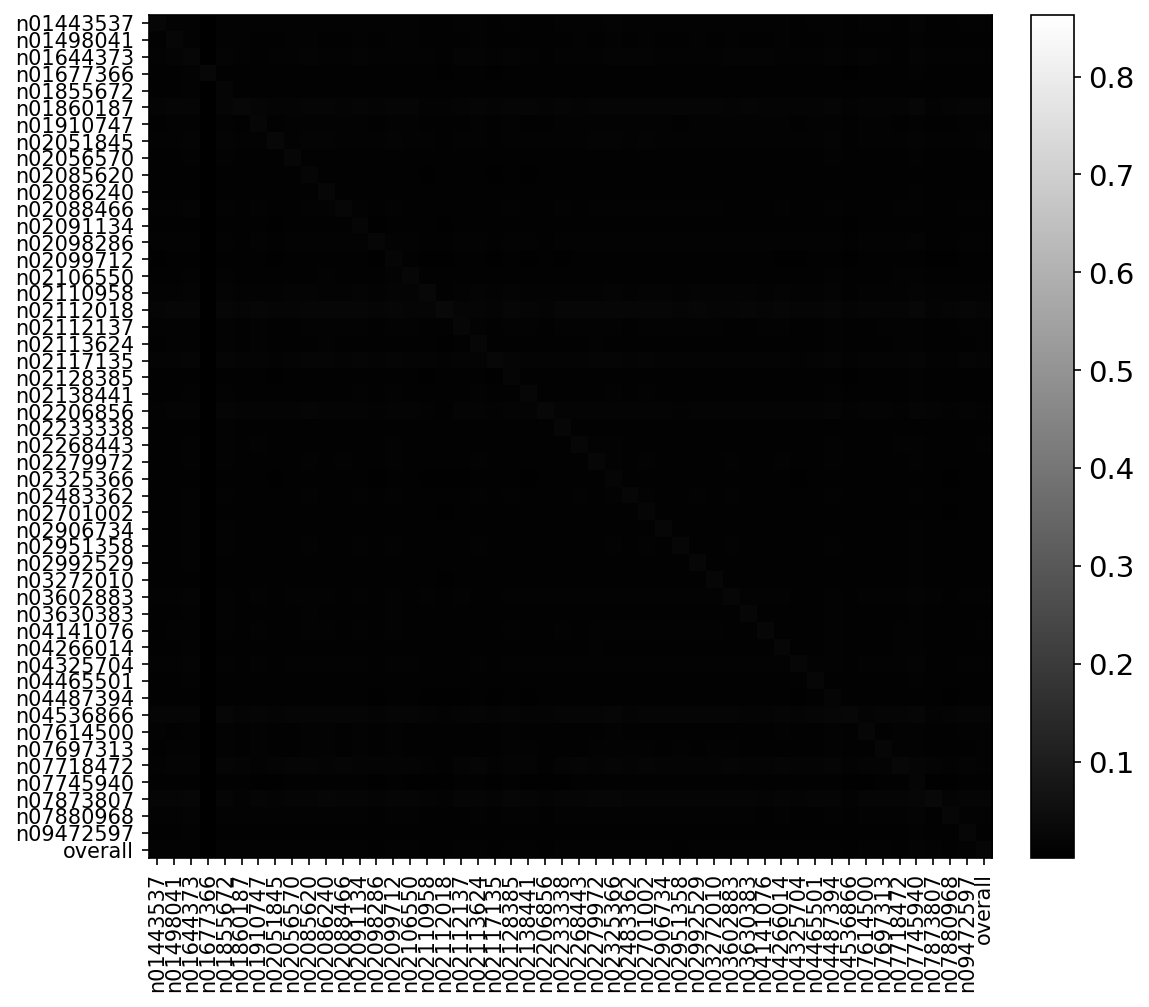}
                    \caption{Random Labels}
                \end{subfigure}
                \begin{subfigure}{0.32\textwidth}
                    \includegraphics[width=\linewidth]{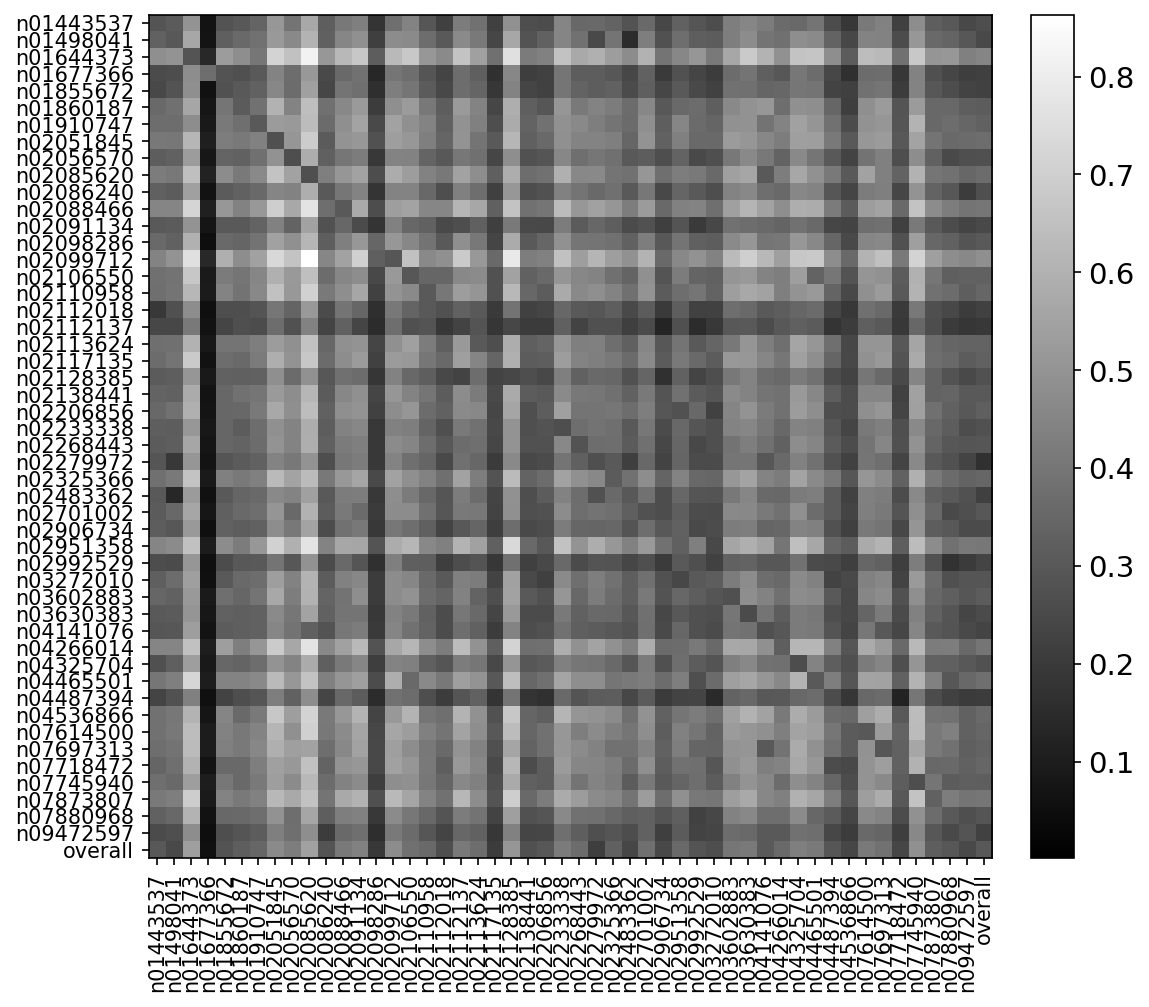}
                    \caption{Normal Labels}
                \end{subfigure}
                \begin{subfigure}{0.32\textwidth}
                    \includegraphics[width=\linewidth]{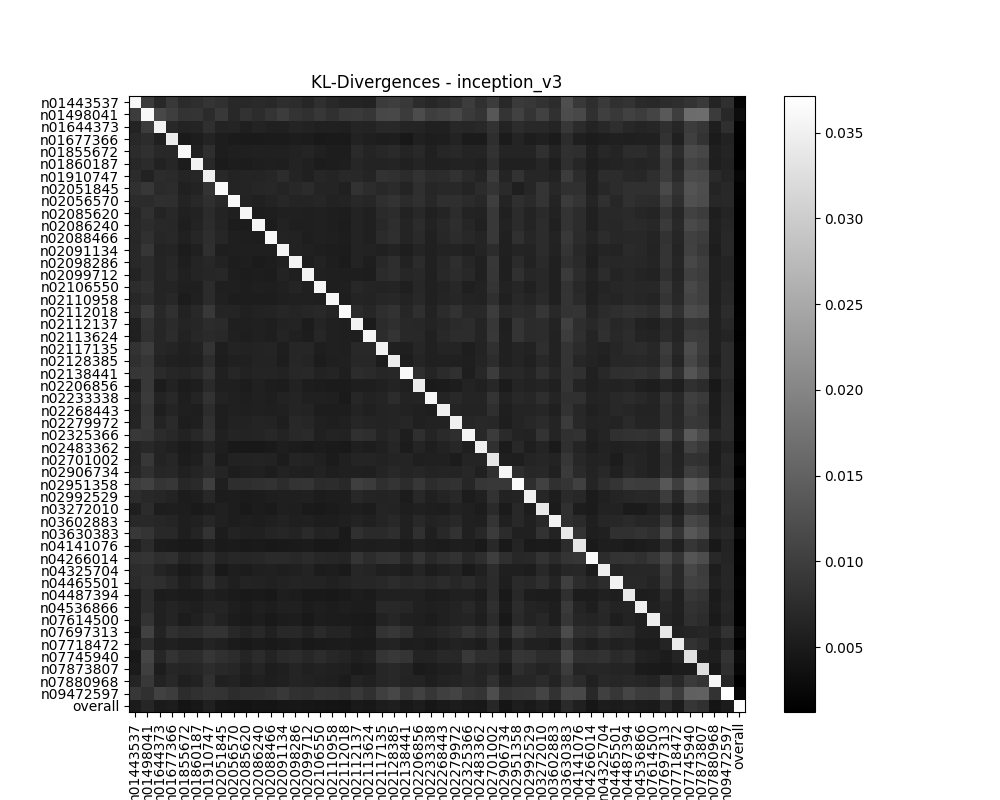}
                    \caption{Unnormalised Initialisation}
                \end{subfigure}
                \begin{subfigure}{0.32\textwidth}
                    \includegraphics[width=\linewidth]{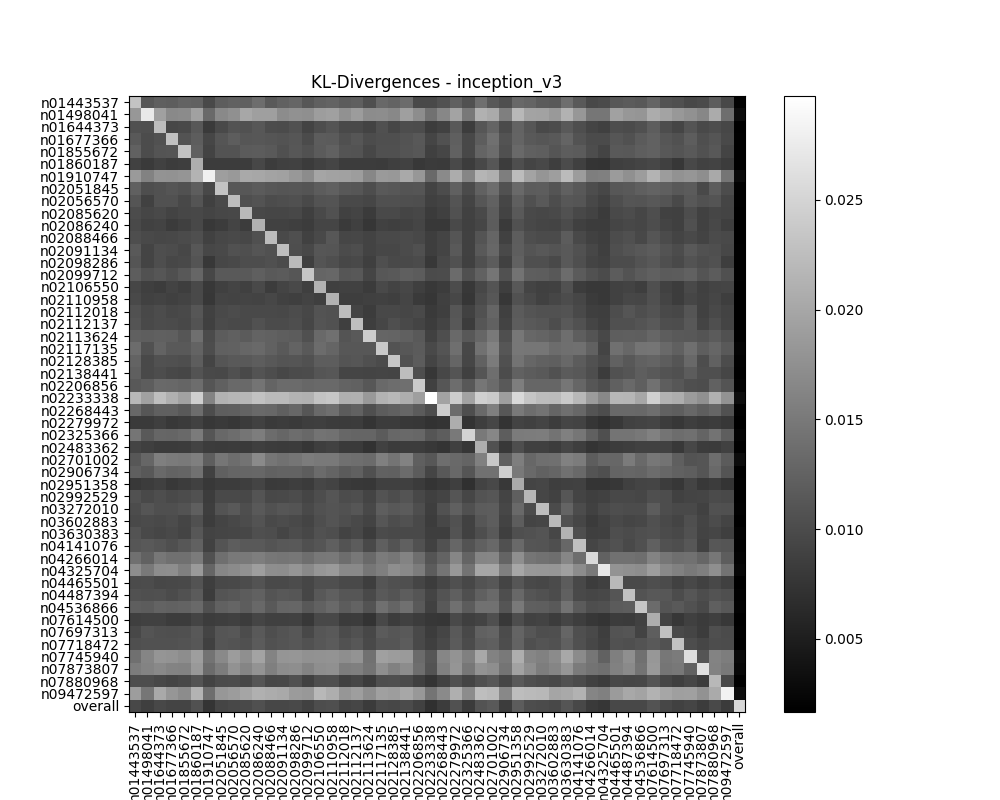}
                    \caption{Unnormalised Random Labels}
                \end{subfigure}
                \begin{subfigure}{0.32\textwidth}
                    \includegraphics[width=\linewidth]{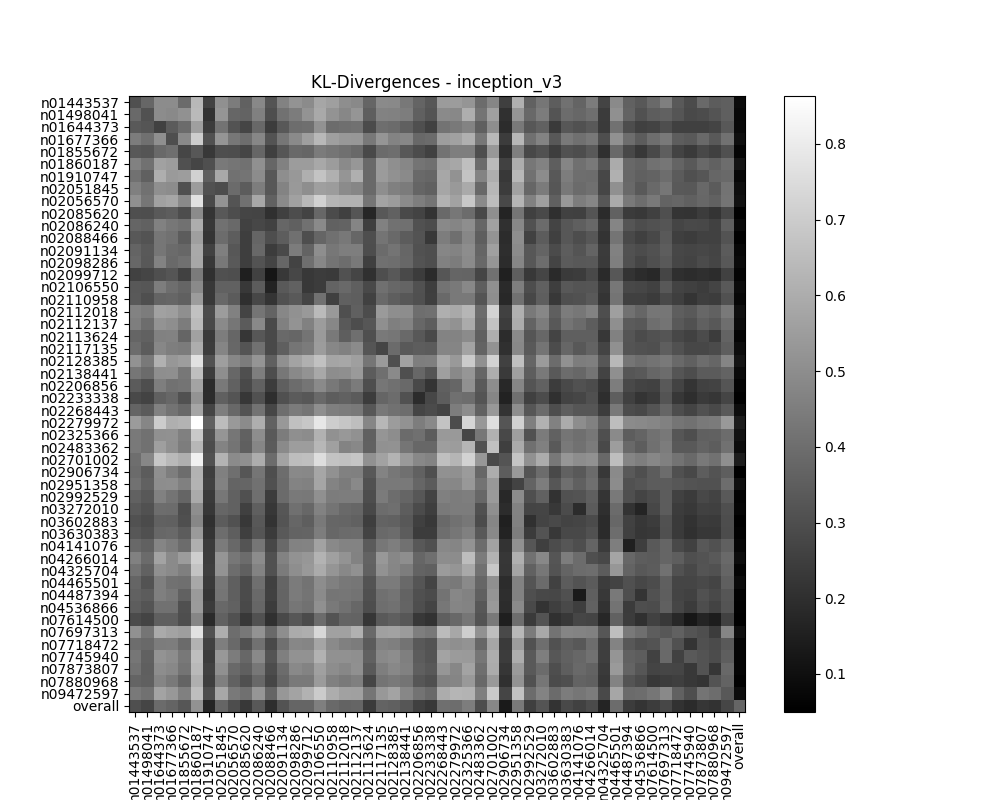}
                    \caption{Unnormalised Normal Labels}
                \end{subfigure}
                \caption{Pairwise KL Divergence Heatmaps for InceptionV3 on ImageNet. Top row contains the heatmaps normalised to the same scale, while the bottom row contains the unnormalised heatmaps. The random labels case shows almost no change from the initialisation, while the normal labels case shows a significant increase in inter-class distances and a decrease in class entropy}
                \label{fig:imagenet_results}
            \end{figure}

            Heatmaps containing the pairwise KL divergence between classes i.) at initialisation, ii.) at convergence when fitting random labels and iii.) at convergence when training with normal labels are shown in figure \ref{fig:imagenet_results}. For ease of viewing, the a sample of 50 classes is shown in the heatmaps with the tables containing statistics from the full dataset. The last row and column contains the same numbers for the overall class-agnostic distrbution across the full test set. The diagonals of the maps are replaced with the entropy of the class distribution instead. The random labels have almost exactly the same divergence structure as the network at initialisation, showing us that no meaningful learning has taken place. In the case with the normal labels, while certain regions remain dark, notably including the distribution for the entire dataset, we see an increase in inter-class distances, as well a marked decrease in the entropy of the classes.
            To complement the visual comparisons, Table \ref{tab:imagenet_Bernoulli_summary} reports scalar summaries from the same matrices: the mean inter-class KL and the mean intra-class KL (diagonal entry). Across all three ImageNet architectures, the trained models exhibit much larger class separation than their untrained counterparts, while the random-label InceptionV3 checkpoint remains close to the initialised network. The untrained ViT-B/32 is an interesting case, as all of the examples register exactly half of the paths as significant and the other half as insignificant. (first row, third subsection of \ref{tab:imagenet_Bernoulli_summary}). 

            \begin{table}[hbtp]
                \centering
                \captionsetup{aboveskip=5pt}
                \begin{tabular}{llcc}
                    \toprule
                    Model & Condition & Mean inter-class KL & Mean Class Entropy \\
                    \midrule
                    InceptionV3 & Untrained & 0.0081 & 0.0426 \\
                    InceptionV3 & Random labels & 0.0115 & 0.0234 \\
                    InceptionV3 & Normal labels & 0.3650 & 0.2993 \\
                    
                    \midrule
                    
                    ResNet50 & Untrained & 0.0086 & 0.0342 \\
                    ResNet50 & Normal labels & 0.2271 & 0.4582 \\

                    \midrule

                    ViT-B/32 & Untrained & 0.0000 & 0.0000 \\
                    ViT-B/32 & Normal labels & 0.1211 & 0.1552 \\
                    \bottomrule
                \end{tabular}
                \caption{Summary statistics for ImageNet Bernoulli KL divergences. Inter-class separation increases with training, even relative to entropy}
                \label{tab:imagenet_Bernoulli_summary}
            \end{table}

\begin{figure}[hbtp]
                \centering
                \begin{subfigure}{0.32\textwidth}
                    \includegraphics[width=\linewidth]{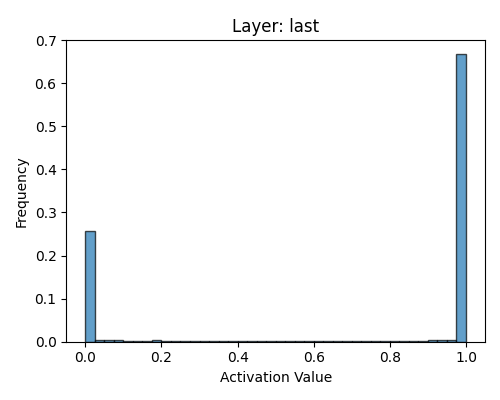}
                    \caption{Initialisation}
                \end{subfigure}
                \begin{subfigure}{0.32\textwidth}
                    \includegraphics[width=\linewidth]{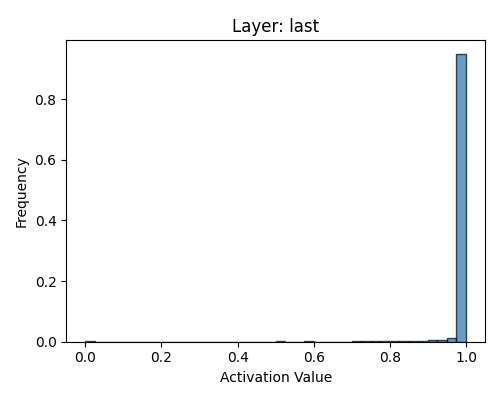}
                    \caption{Random Labels}
                \end{subfigure}
                \begin{subfigure}{0.32\textwidth}
                    \includegraphics[width=\linewidth]{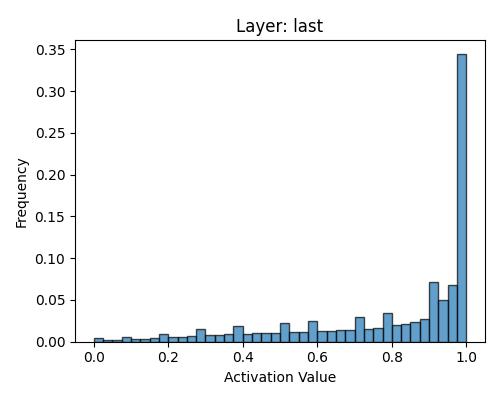}
                    \caption{Normal Labels}
                \end{subfigure}
                \begin{subfigure}{0.32\textwidth}
                    \includegraphics[width=\linewidth]{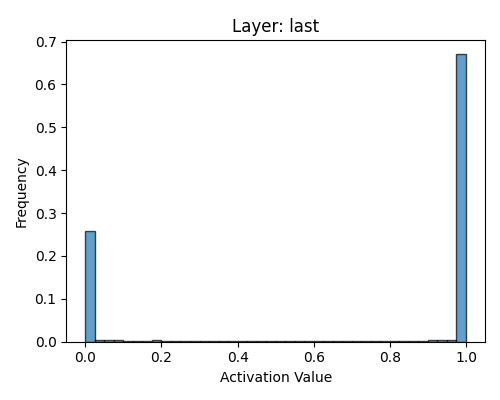}
                    \caption{Initialisation}
                \end{subfigure}
                \begin{subfigure}{0.32\textwidth}
                    \includegraphics[width=\linewidth]{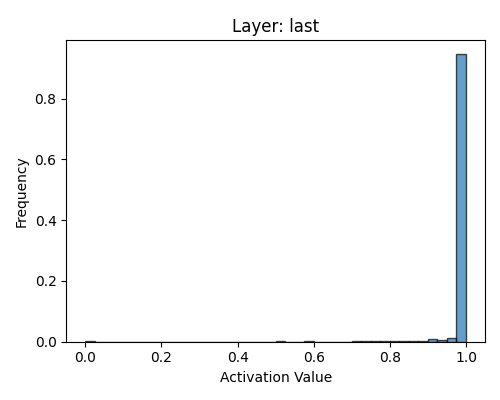}
                    \caption{Random Labels}
                \end{subfigure}
                \begin{subfigure}{0.32\textwidth}
                    \includegraphics[width=\linewidth]{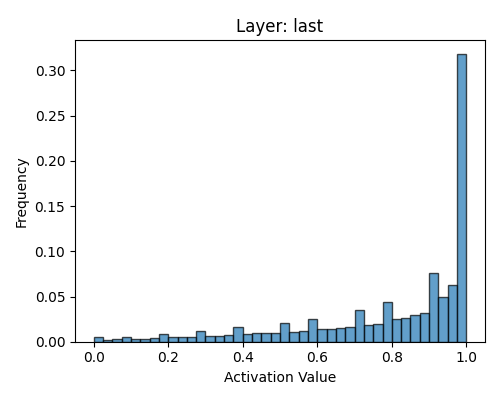}
                    \caption{Normal Labels}
                \end{subfigure}
                \caption{Activation Sparsity Histograms for InceptionV3 on ImageNet. The x-axis represents the proportion of paths that are significant for the class, while the y-axis represents the number of paths that have that proportion. Training the network to fit random labels causes a decrease in sparsity and training with normal labels causes an increase. The top row corresponds to the class 'king penguin' (n02056570) while the bottom row corresponds to the class 'cheeseburger' (n07697313). The same trends are observed for both classes, with the normal labels case showing a much higher degree of sparsity than the random labels case.}
                \label{fig:inceptionv3_sparsity}
            \end{figure}

            Looking at the behaviour of the network with regards to sparsity, we attempt to visualise a histogram of the frequency of a path being significant against the number of paths which have that frequency in Figure \ref{fig:inceptionv3_sparsity}. 

            Here, we can see that the sparsity of the network increases meaningfully over the course of training with normal labels. However, with the random labels case we can see that the activations become more dense, lending credence to the idea that the structure of the significance matrix directly relates to the robustness of the network. 
            
\begin{figure}[hbtp]
                \centering
                \begin{subfigure}{0.32\textwidth}
                    \includegraphics[width=\linewidth]{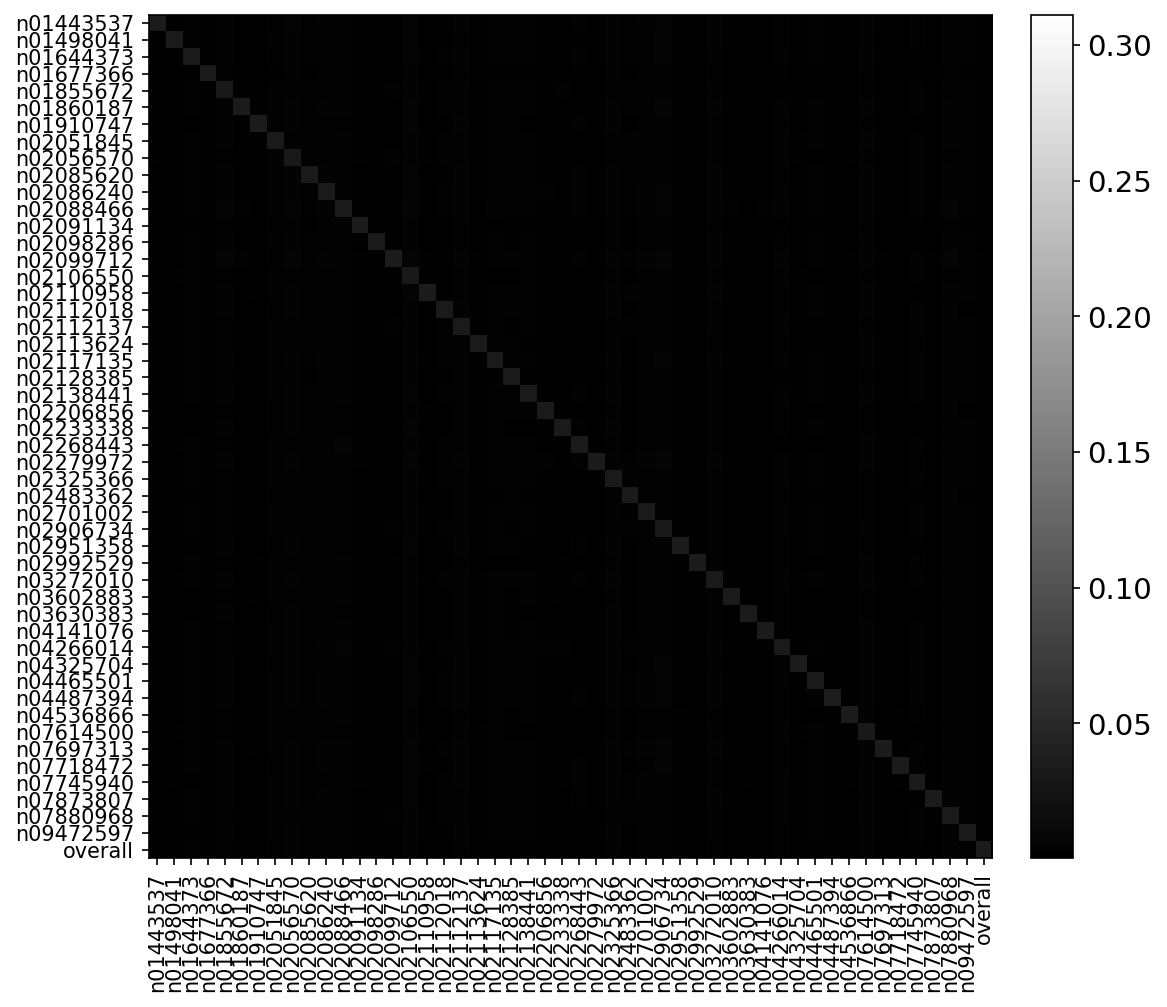}
                    \caption{Untrained}
                \end{subfigure}
                \begin{subfigure}{0.32\textwidth}
                    \includegraphics[width=\linewidth]{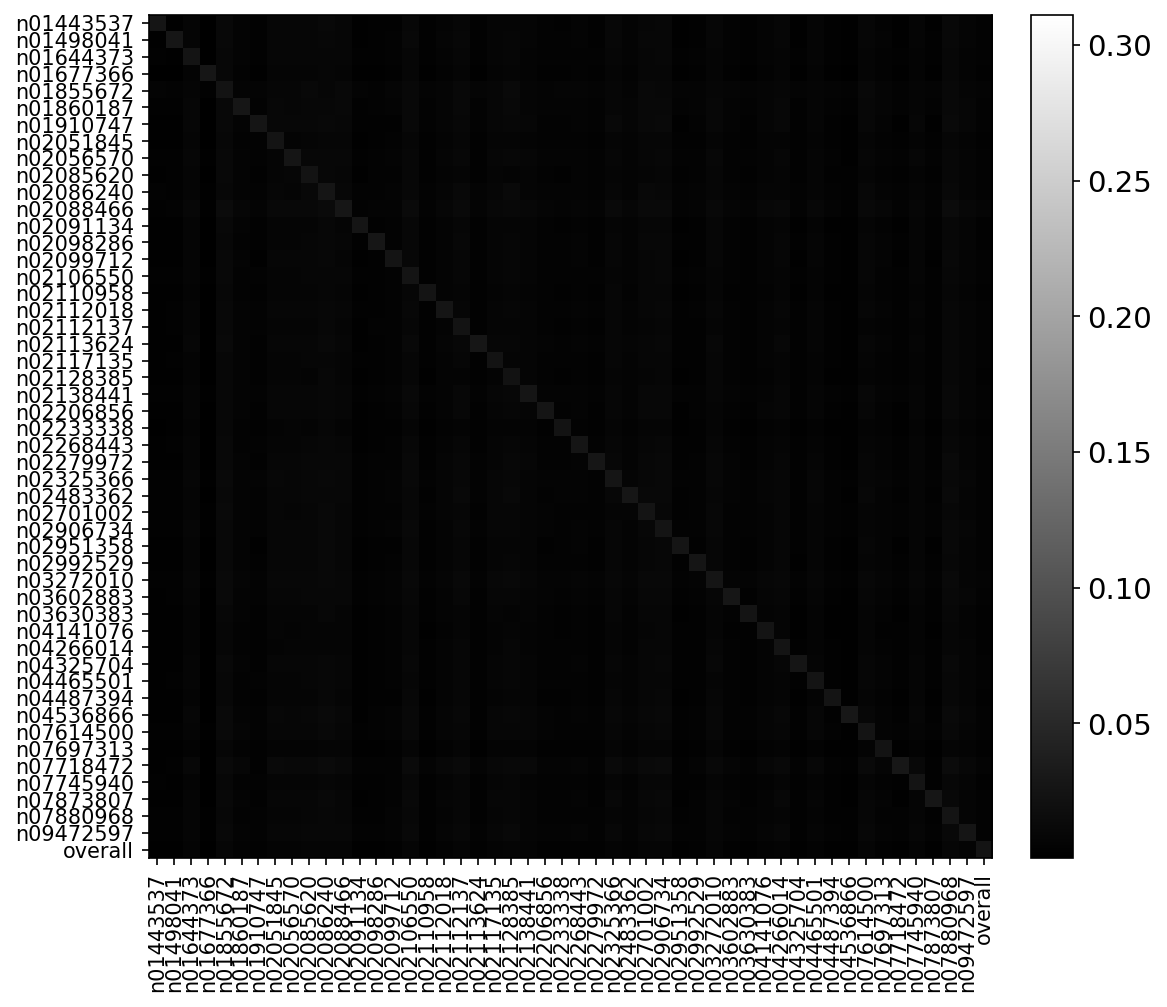}
                    \caption{Random Labels}
                \end{subfigure}
                \begin{subfigure}{0.32\textwidth}
                    \includegraphics[width=\linewidth]{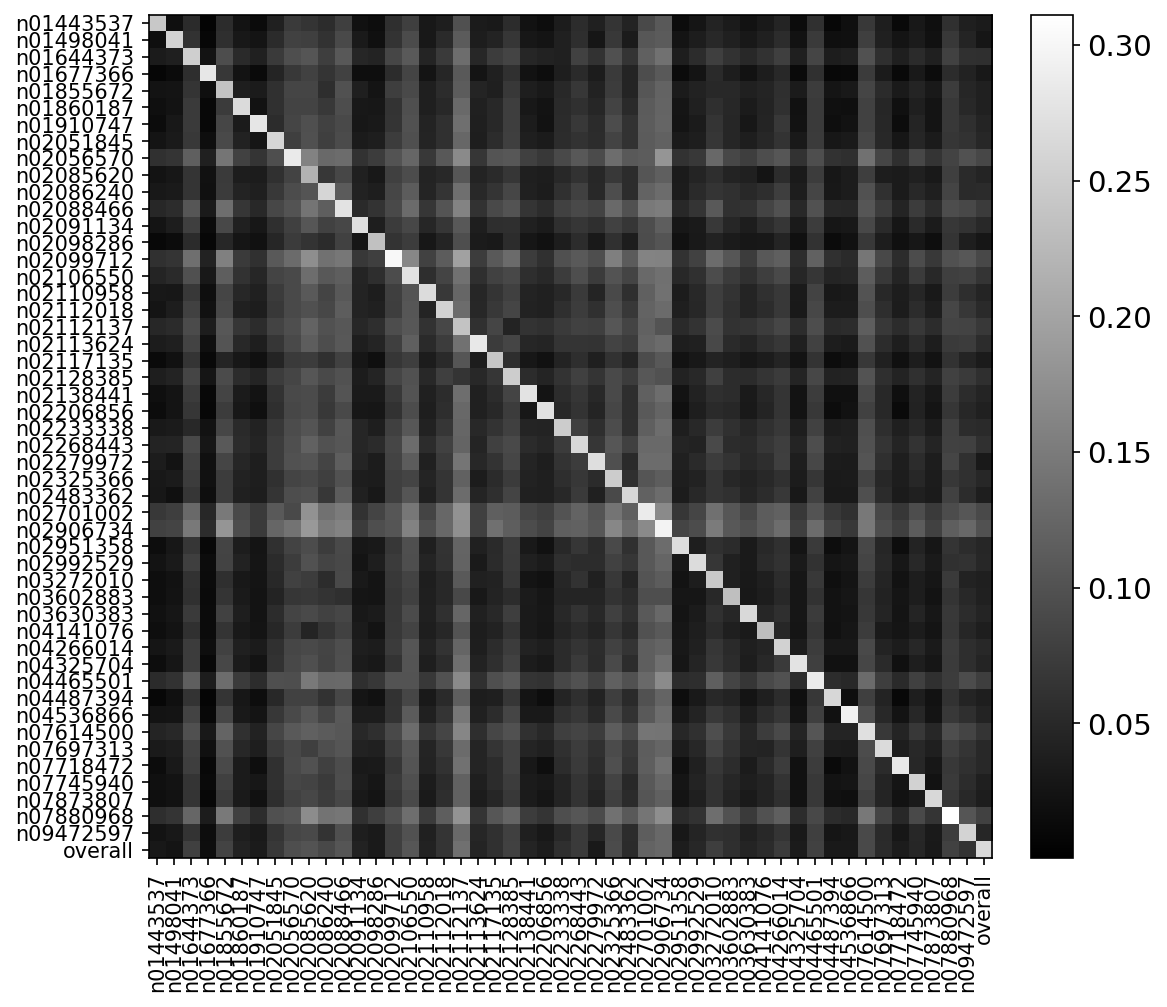}
                    \caption{Normal Labels}
                \end{subfigure}
                \caption{Pairwise KL Divergence Heatmaps for InceptionV3 on ImageNet-R. The separation betwen OOD classes is much smaller than the in-distribution case}
                \label{fig:imagenet_ood_results}
            \end{figure}

            Now, for out-of-distribution data, the inter-class distances are shown in Figure \ref{fig:imagenet_ood_results} and Table \ref{tab:imagenet_r_Bernoulli_summary}. Here, we can see that the inter-class distances for the OOD data are significantly lower than those for the in-distribution data, with many classes being very close to each other. Under our diagnostic metric, this indicates weaker class separation for OOD inputs.

            \begin{table}[hbtp]
                \centering
                \captionsetup{aboveskip=5pt}
                \begin{tabular}{llcc}
                    \toprule
                    Model & Condition & Mean inter-class KL & Mean Class Entropy \\
                    \midrule
                    InceptionV3 & Untrained & 0.026 & 0.0329 \\
                    InceptionV3 & Random labels & 0.0058 & 0.0260 \\
                    InceptionV3 & Normal labels & 0.0635 & 0.2646 \\
                    
                    \midrule
                    
                    ResNet50 & Untrained & 0.0031 & 0.0277 \\
                    ResNet50 & Normal labels & 0.0776 & 0.5765 \\

                    \midrule

                    ViT-B/32 & Untrained & 0.0000 & 0.0000 \\
                    ViT-B/32 & Normal labels & 0.0335 & 0.1909 \\
                    \bottomrule
                \end{tabular}
                \caption{Summary statistics for ImageNet-r Bernoulli KL divergences. The increases in inter-class KL are much smaller than the in-distribution case, and the class entropies are much higher, indicating weaker class separation for OOD data.}
                \label{tab:imagenet_r_Bernoulli_summary}
            \end{table}

            Here, we can see that the inter-class separation does not change meaningfully in the random label case, as expected. There are minor increases in separation between classes when training on ImageNet, but the magnitude of distances is much smaller, alongside the entropy for the OOD classes being much larger than their in-distribution counterparts.

\begin{figure}[hbtp]
                \centering
                \begin{subfigure}{0.32\textwidth}
                    \includegraphics[width=\linewidth]{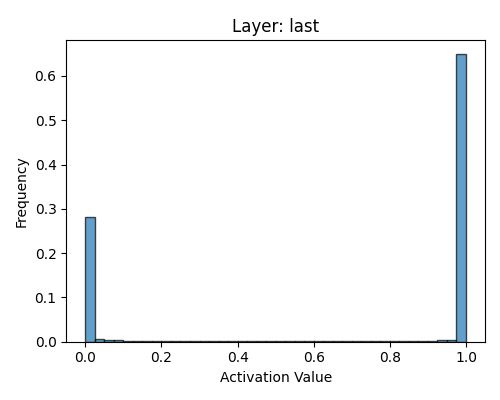}
                    \caption{Initialisation}
                \end{subfigure}
                \begin{subfigure}{0.32\textwidth}
                    \includegraphics[width=\linewidth]{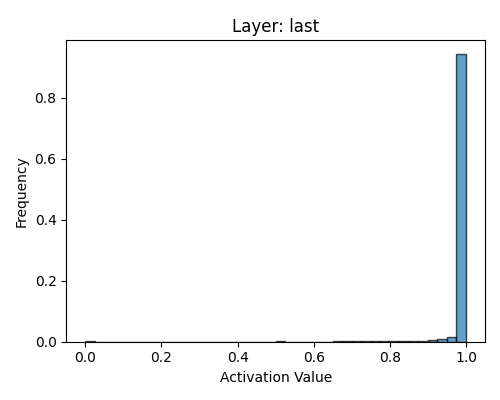}
                    \caption{Random Labels}
                \end{subfigure}
                \begin{subfigure}{0.32\textwidth}
                    \includegraphics[width=\linewidth]{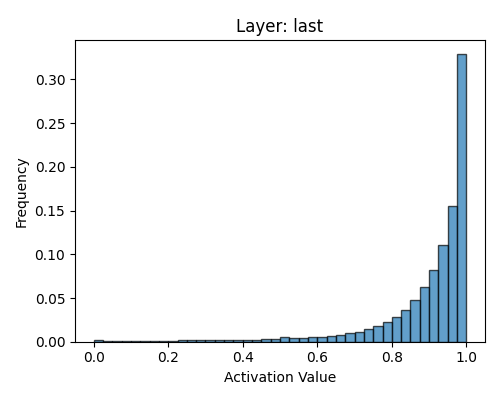}
                    \caption{Normal Labels}
                \end{subfigure}

                \caption{Activation Sparsity Histograms for InceptionV3 on ImageNet-r for the class 'king penguin' (n02056570). The network is much denser for the data compared to the ID data}
                \label{fig:inceptionv3_sparsity_king_penguin_OOD}
            \end{figure}

            The sparsity results for the OOD data are in Figures \ref{fig:inceptionv3_sparsity_king_penguin_OOD}, \ref{fig:inceptionv3_sparsity_cheeseburger_OOD}. Once again, the untrained and random label cases are exactly the same as the in-distribution case. However, in the normal labels case the activations are significantly more dense than the in-distribution data, signified by the heavier tail of the distribution. This is in line with our hypothesis that the robustness of the network is linked with the sparsity of the significance matrix. Additionally, \cite{sun2021reactoutofdistributiondetectionrectified} shows that the shift in batch normalisation statistics is the primary cause for the drop in performance. This result supports that, because improper normalisation will lead to more neurons having abnormally large magnitudes in both the positive and negative directions, which causes the number of significant neuron-weight interactions to increase drastically.  

\begin{figure}[hbtp]
                \centering
                \includegraphics[width=\linewidth]{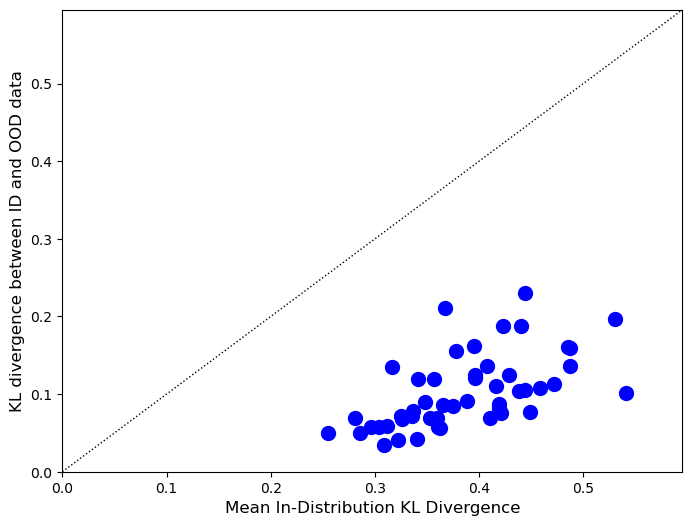}
                \caption{KL divergence for class distributions between in-distribution and OOD data for InceptionV3. 50 out of 200 common classes between ImageNet-r and ImageNet are shown. The inter-class distances for the OOD data are significantly smaller than those for the average divergence from the in-distribution data, with many classes being very close to each other. Under our diagnostic metric, this indicates weaker class separation for OOD inputs.}
                \label{fig:in_out_distances}
            \end{figure}

            The results for the distances between the distributions of in-distribution and OOD classes are in figure \ref{fig:in_out_distances}, where we see that the separation between OOD classes and their ID counterparts is much smaller than the separation between different ID classes. Within this framework, this indicates that OOD examples often occupy nearby regions of the learned class-conditional path distributions. This motivates using these statistics as robustness diagnostics and suggests that reliable OOD discrimination may require objectives beyond post-hoc scoring alone.
            
        \subsection{Ablations: Metric Choices}
            To showcase the utility of our choice of metric, we also examine another set of metrics.
            \begin{enumerate}
                \item 
                    We look at an instance of the matrix $N$, take the average value of it over the entire class and compute the row-wise KL divergence between classes. The idea is to see if the 'prototype' neuron-weight interaction for classes is meaningfully different. This is to ablate the choice of using the Bernoulli distributions and significance to model the interactions.
                \item
                    We look at the softmax of the final layer's activations, and compute the average KL divergence between collections of softmax outputs between each classes, compared. The justification for this metric is that the network is trained to differentiate inputs in a probabilitic sense, so the softmax of the final layer's activations should be a meaningful representation of the distribution of the class in the latent space. Additionally, we compute the average KL divergence within the class to measure the spread of the distribution. The goal of this is to ablate the choice of using the interactions between weights and activations, and see if the same properties hold up when looking at the activations themselves.
                \item 
                    We measure the energy distances between the distributions of activations for each class. The energy distance is a measure of distance between collections of point masses that doesn't make any distributional assumptions about the data. For the inter-class separation, we compute the expected distance over every pair of points between the two classes, and for the intra-class spread we compute the expected distance over every pair of points within the class. This is similar to the above in ablating the choice of using the interactions between weights and activations, but it also ablates the choice of using the KL divergence as a distance measure, as the energy distance is based on the Euclidean distance between points in the latent space.
            \end{enumerate}
            We present our findings in the table \ref{tab:imagenet_ablations}. The KL divergence between the significant paths is the only metric that shows a consistent pattern of a meaningful increase in separation between classes when training with normal labels, while the other three metrics show inconsistent patterns. We see cases of untrained networks having larger inter-class distances than trained networks, and cases of random label networks having larger inter-class distances than normal label networks. This suggests that the structure of the significance matrix is a more meaningful diagnostic of robustness than the other metrics we examined. 

            \begin{table}[hbtp]
                \centering
                \captionsetup{aboveskip=5pt}
                \begin{tabular}{cccc}
                    \toprule
                    Model & Condition & Mean Inter-Class Separation & Mean Intra-Class Spread \\
                    \midrule
                    \multicolumn{4}{c}{KL Divergence Between Significant Path Proportions} \\
                    \midrule
                    InceptionV3 & Untrained & 0.0081 & 0.0426 \\
                    InceptionV3 & Random labels & 0.0115 & 0.0234 \\
                    InceptionV3 & Normal labels & 0.3650 & 0.2993 \\
                    \midrule
                    ResNet50 & Untrained & 0.0086 & 0.0342 \\
                    ResNet50 & Normal labels & 0.2271 & 0.4582 \\
                    \midrule
                    ViT-B/32 & Untrained & 0.0000 & 0.0000 \\
                    ViT-B/32 & Normal labels & 0.1211 & 0.1552 \\
                    \midrule
                    \multicolumn{4}{c}{KL Divergence Between Average Neuron-Weight Interactions} \\
                    \midrule
                    InceptionV3 & Untrained & 2 $\times$ 10$^{-9}$ & 7.6251 \\
                    InceptionV3 & Random labels & 6 $\times$ 10$^{-10}$ & 7.6251 \\
                    InceptionV3 & Normal labels & 6 $\times$ 10$^{-13}$ & 7.6251 \\
                    \midrule
                    ResNet50 & Untrained & 1 $\times$ 10$^{-10}$ & 7.6251 \\
                    ResNet50 & Normal labels & 6 $\times$ 10$^{-15}$ & 7.6251 \\
                    \midrule
                    ViT-B/32 & Untrained & 0.0000 & 6.6451 \\
                    ViT-B/32 & Normal labels & 3 $\times$ 10$^{-7}$ & 6.6451 \\
                    \midrule
                    \multicolumn{4}{c}{KL Divergence Between Softmaxed Final Layer Activations} \\
                    \midrule
                    InceptionV3 & Untrained & 0.0595 & 0.0606 \\
                    InceptionV3 & Random labels & 0.03424 & 0.3499 \\
                    InceptionV3 & Normal labels & 0.0725 & 0.0743 \\
                    \midrule
                    ResNet50 & Untrained & 0.1106 & 0.1127 \\
                    ResNet50 & Normal labels & 0.1348 & 0.1350 \\
                    \midrule
                    ViT-B/32 & Untrained & 0.6593 & 0.5836 \\
                    ViT-B/32 & Normal labels & 0.3933 & 0.2485 \\
                    \midrule
                    \multicolumn{4}{c}{Energy Distance Between Distributions of Activations} \\
                    \midrule
                    InceptionV3 & Untrained & 13.6103 & 13.3820 \\
                    InceptionV3 & Random labels & 29.6274 & 29.3145 \\
                    InceptionV3 & Normal labels & 15.7113 & 15.5280 \\
                    \midrule
                    ResNet50 & Untrained & 21.0639 & 20.6740 \\
                    ResNet50 & Normal labels & 13.7042 & 13.2803 \\
                    \midrule
                    ViT-B/32 & Untrained & 30.8505 & 28.1287 \\
                    ViT-B/32 & Normal labels & 24.3184 & 18.1244 \\
                    \bottomrule
                \end{tabular}

            \caption{Summary statistics for ImageNet alternative metrics. No other metric shows a consistent pattern of an increase in inter-class separation when training with normal labels apart from the KL divergence between significant paths}
            \label{tab:imagenet_ablations}
        \end{table}
    \section{Conclusion and Future Work}
        In this paper, we present a method for analysing the robustness of neural networks based on the concept of significant paths. Across the experiments in this work, we find that the structure of the significance matrix is closely related to memorisation and robustness, with better-performing models typically exhibiting sparser and more well-separated paths. Under distribution shift, these diagnostics show reduced separation and increased entropy. We emphasise that this framework is intended as a diagnostic tool for representation analysis rather than a replacement for dedicated OOD detection methods. We hope to extend this analysis to layers that are not fully connected, such as convolutional layers or attention layers, which would allow analysis of a much larger class of architectures widely used for tasks such as segmentation and language modelling.

\bibliographystyle{plain}
\bibliography{references}

\appendix

    \section{TinyImageNet to CIFAR10 class mapping}
        \begin{tabular}{|c|c|}
            \hline
            Class in CIFAR10 & WNIDs from TinyImageNet \\
            \hline
            automobile & n03599486, n02814533, n03444034, n03100240 \\
            \hline
            bird & n02058221, n02002724 \\
            \hline
            cat & n02124075, n02125311, n02123394, n02509815, n02123045 \\
            \hline
            deer & n02423022 \\
            \hline
            dog & n02106662, n02099712, n02094433, n02085620 \\
            \hline
            frog & n01641577, n01644900 \\
            \hline
            truck & n04146614, n02917067, n04465501, n04487081 \\
            \hline
        \end{tabular}
    \section{Compute and Architectural Details}
        All experiments were run on a system with a single NVIDIA A5000 GPU, with 24GB of VRAM and an Intel Xeon W-2295 CPU (3.00GHz, 18 cores, 36 threads). The modified Alexnet and InceptionV3 were trained for 250 epochs, with the models with the highest validation accuracy being analysed. The batch size was set at 32 unless specified otherwise.
    
        The modified version of Alexnet is adapted to the CIFAR10 dataset, which is much smaller than the ImageNet dataset that Alexnet was built for. The convolutional block here is a 5 $\times$ 5 convolution, followed by a 3 $\times$ 3 max pool and a local response normalisation layer with size 5, $\alpha=0.0001$, $\beta=0.75$ and $k=2$. Two convolutional blocks are followed by a two fully-connected hidden layers with 384 and 192 neurons, with the output having 10 neurons. The output is softmax activated, while the other layers are ReLU activated.

        The optimizer used is SGD, with the initial learning rate set to 0.01, decaying by 0.95 after every epoch. No explicit regularisation such as dropout or weight decay are used. Each 32 $\times$ 32 image is cropped to a 28 $\times$ 28 resolution, with each image \textit{independently} standardised to have zero mean and unit standard deviation.

    \section{ImageNet Sampled Class List}
        \begin{longtable}{@{}ll@{}} 
            \toprule 
            \textbf{Item Name} & \textbf{Code} \\
            \midrule 
            \endhead 
            \bottomrule 
            \endfoot 
            
            leopard & n02128385 \\
            canoe & n02951358 \\
            Rottweiler & n02106550 \\
            chow & n02112137 \\
            broom & n02906734 \\
            cheeseburger & n07697313 \\
            cellular telephone & n02992529 \\
            goldfish & n01443537 \\
            volcano & n09472597 \\
            West Highland white terrier & n02098286 \\
            lab coat & n03630383 \\
            hyena & n02117135 \\
            ice cream & n07614500 \\
            toy poodle & n02113624 \\
            Pomeranian & n02112018 \\
            strawberry & n07745940 \\
            black swan & n01860187 \\
            ambulance & n02701002 \\
            cockroach & n02233338 \\
            Chihuahua & n02085620 \\
            Labrador retriever & n02099712 \\
            pug & n02110958 \\
            tractor & n04465501 \\
            wood rabbit & n02325366 \\
            stingray & n01498041 \\
            bloodhound & n02088466 \\
            sax & n04141076 \\
            pizza & n07873807 \\
            bee & n02206856 \\
            whippet & n02091134 \\
            king penguin & n02056570 \\
            pelican & n02051845 \\
            cucumber & n07718472 \\
            burrito & n07880968 \\
            common iguana & n01677366 \\
            space shuttle & n04266014 \\
            stole & n04325704 \\
            tree frog & n01644373 \\
            trombone & n04487394 \\
            goose & n01855672 \\
            joystick & n03602883 \\
            dragonfly & n02268443 \\
            jellyfish & n01910747 \\
            electric guitar & n03272010 \\
            gibbon & n02483362 \\
            violin & n04536866 \\
            monarch & n02279972 \\
            Shih-Tzu & n02086240 \\
            meerkat & n02138441 \\
            \caption{List of Items and Codes} 
            \label{tab:itemcodes} 
            \end{longtable}

        \section{Illustrative Examples from ImageNet and ImageNet-r}
\begin{figure}[hbtp]
                \centering
                \begin{subfigure}{0.40\textwidth}
                    \includegraphics[width=\linewidth]{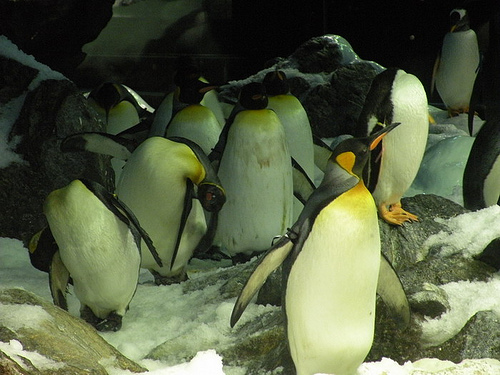}
                    \subcaption{ID (ImageNet)}
                \end{subfigure}
                \begin{subfigure}{0.40\textwidth}
                    \includegraphics[width=\linewidth]{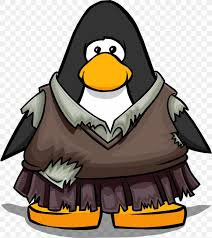}
                    \subcaption{OOD 2 (ImageNet-R)}
                \end{subfigure}
                \caption{Example of in-distribution and out-of-distribution images for the class n02056570 (penguin) from ImageNet and ImageNet-r respectively}
                \label{fig:ood_examples_penguin}
            \end{figure}

\begin{figure}[hbtp]
                \centering
                \begin{subfigure}{0.40\textwidth}
                    \includegraphics[width=\linewidth]{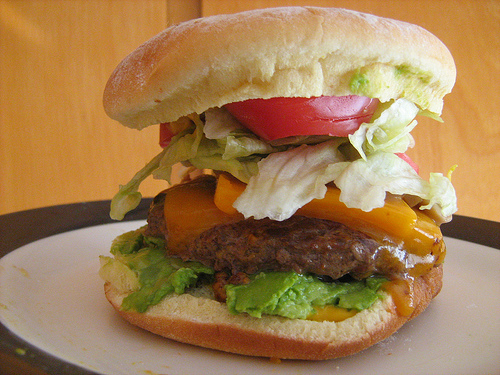}
                    \subcaption{ID (ImageNet)}
                \end{subfigure}
                \begin{subfigure}{0.40\textwidth}
                    \includegraphics[width=\linewidth]{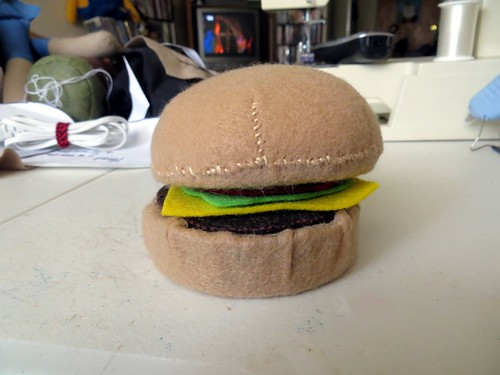}
                    \subcaption{OOD 1 (ImageNet-R)}
                \end{subfigure}
                \caption{Example of in-distribution and out-of-distribution images for the class n07697313 (burger) from ImageNet and ImageNet-r respectively}
                \label{fig:ood_examples_burger}
            \end{figure}

            We attach two pairs of examples from different classes common to ImageNet and ImageNet-r to illustrate the distribution shift between the two datasets. The first pair of examples is from the class n02056570 (king penguin), while the second pair is from the class n07697313 (cheeseburger). The in-distribution images are taken from the ImageNet validation set, while the out-of-distribution images are taken from the ImageNet-r dataset. These can be found in Figures \ref{fig:ood_examples_penguin} and \ref{fig:ood_examples_burger} respectively. The OOD images are stylised versions of the ID images, with the penguin being a cartoon version of the original image and the cheeseburger being a toy version of the original image. These examples illustrate the significant distribution shift between the two datasets, which is reflected in the reduced class separation and increased entropy for the OOD data in our diagnostics.
            
        \section{Ablations: Visualised Heatmaps for Alternative Metrics}
            Here, we present the heatmaps for the alternative metrics described in Section 4.2. The first three rows contain the energy distance heatmaps, while the last three rows contain the softmax KL divergence heatmaps. The same sample of 50 classes is shown in the heatmaps as in figure \ref{fig:imagenet_results} for ease of viewing, with the tables containing statistics from the full dataset.

            \begin{figure}[hbtp]
                \centering
                \begin{subfigure}{0.32\textwidth}
                    \includegraphics[width=\linewidth]{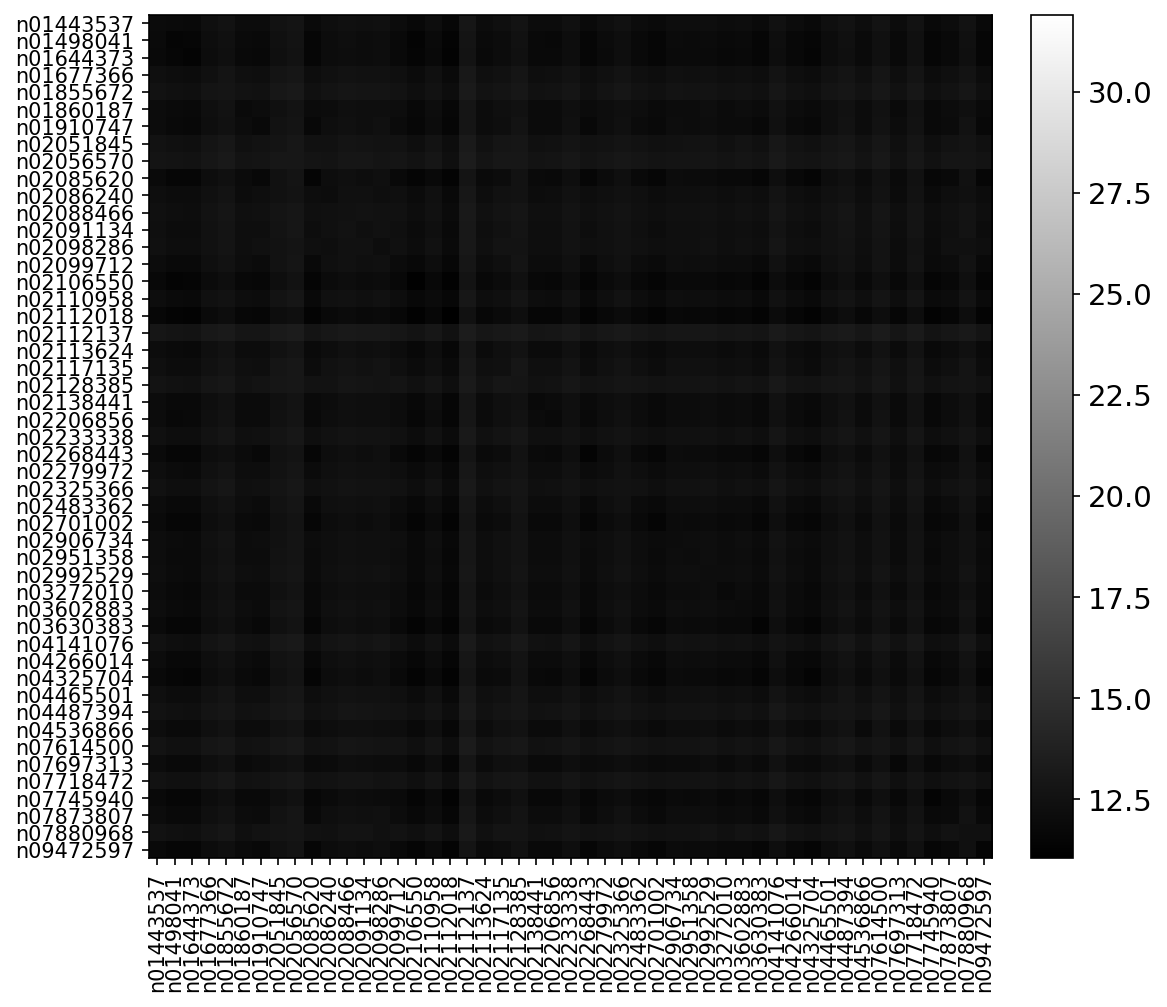}
                    \caption{Untrained}
                \end{subfigure}
                \begin{subfigure}{0.32\textwidth}
                    \includegraphics[width=\linewidth]{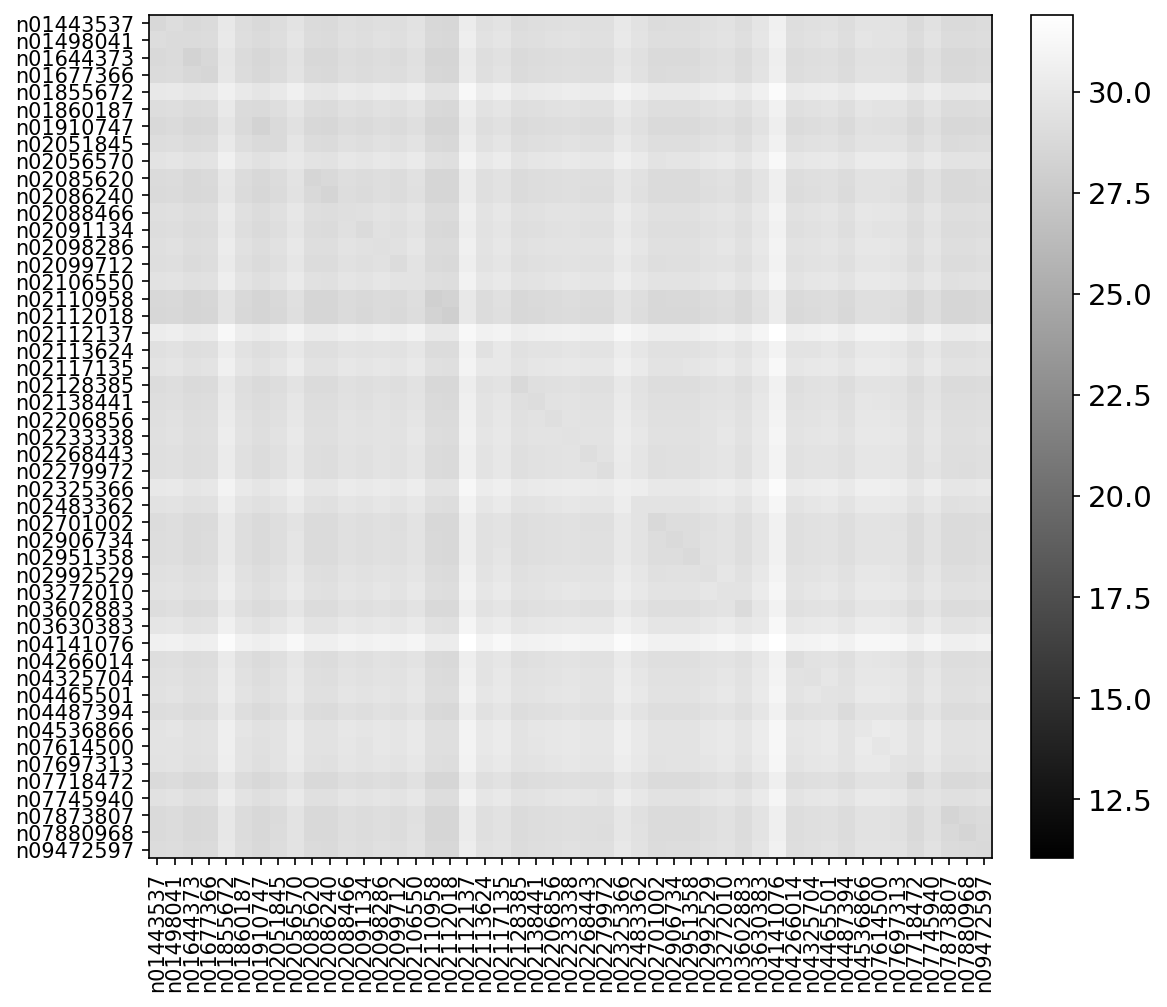}
                    \caption{Random Labels}
                \end{subfigure}
                \begin{subfigure}{0.32\textwidth}
                    \includegraphics[width=\linewidth]{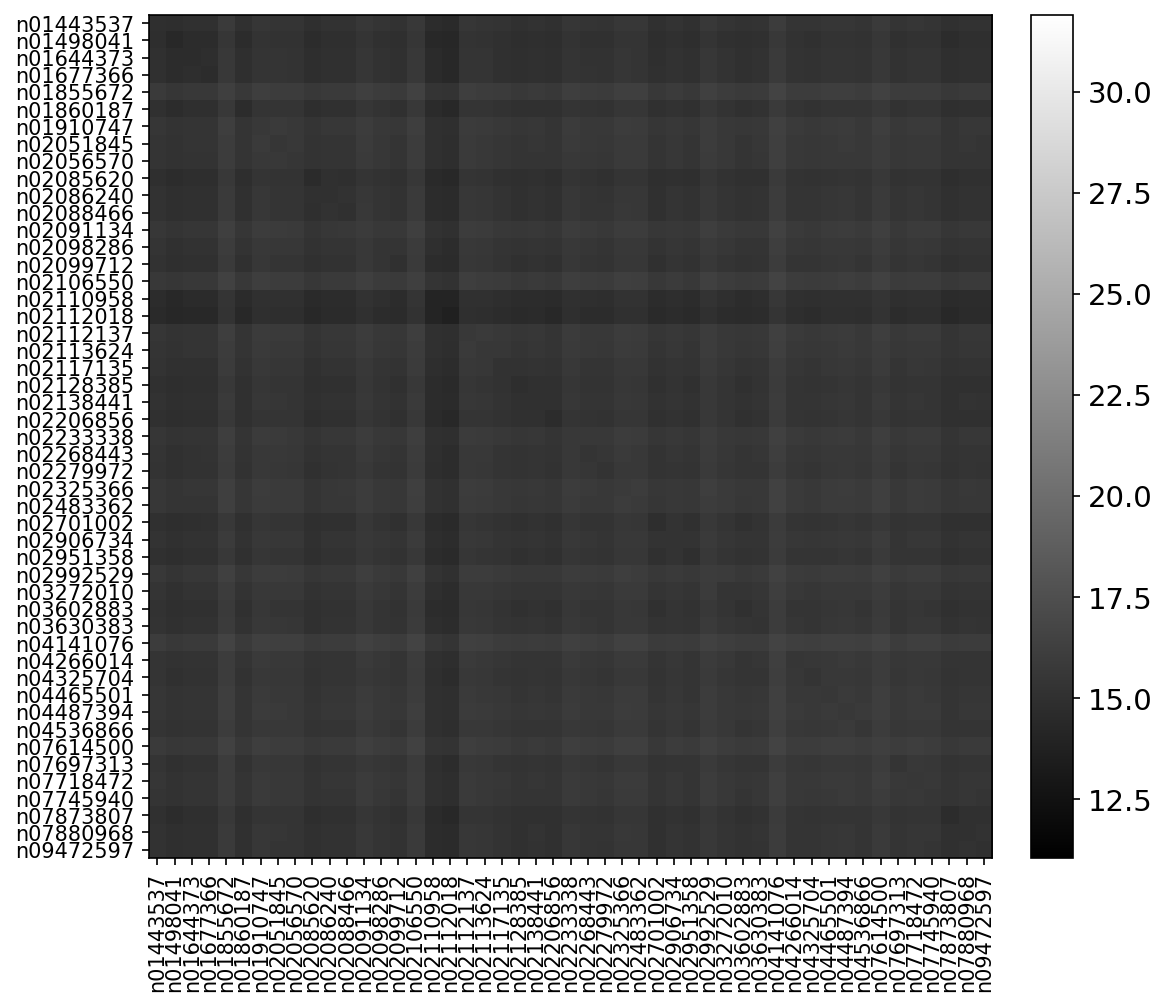}
                    \caption{Normal Labels}
                \end{subfigure}
                \caption{Pairwise Energy Distance Heatmaps for InceptionV3 on ImageNet-R. The random labels case shows a much larger inter-class distance than the initialised network, which is the opposite of what a meaningful diagnostic would show}
                \label{fig:inceptionv3_energy}
            \end{figure}

            \begin{figure}[hbtp]
                \centering
                \begin{subfigure}{0.40\textwidth}
                    \includegraphics[width=\linewidth]{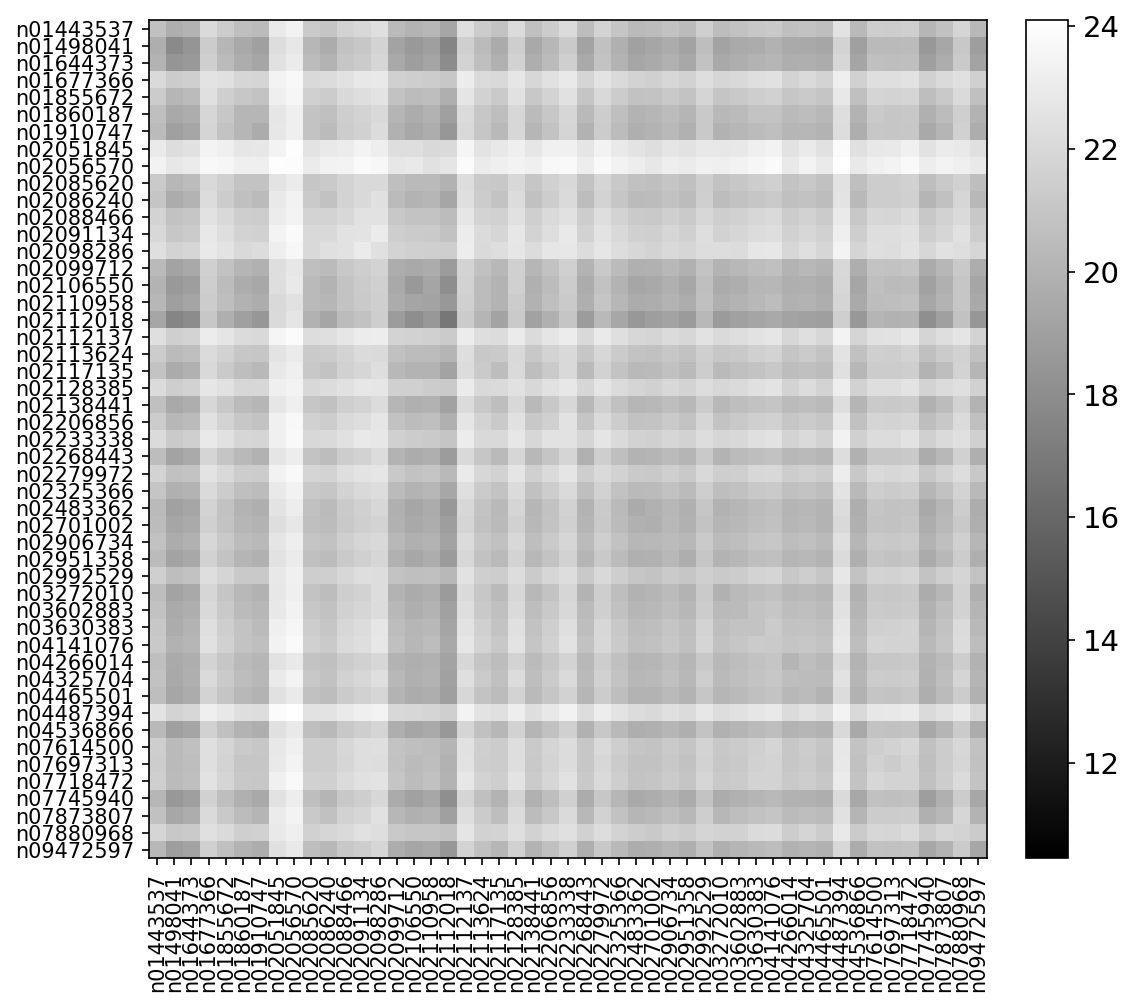}
                    \caption{Resnet50-Untrained}
                \end{subfigure}
                \begin{subfigure}{0.40\textwidth}
                    \includegraphics[width=\linewidth]{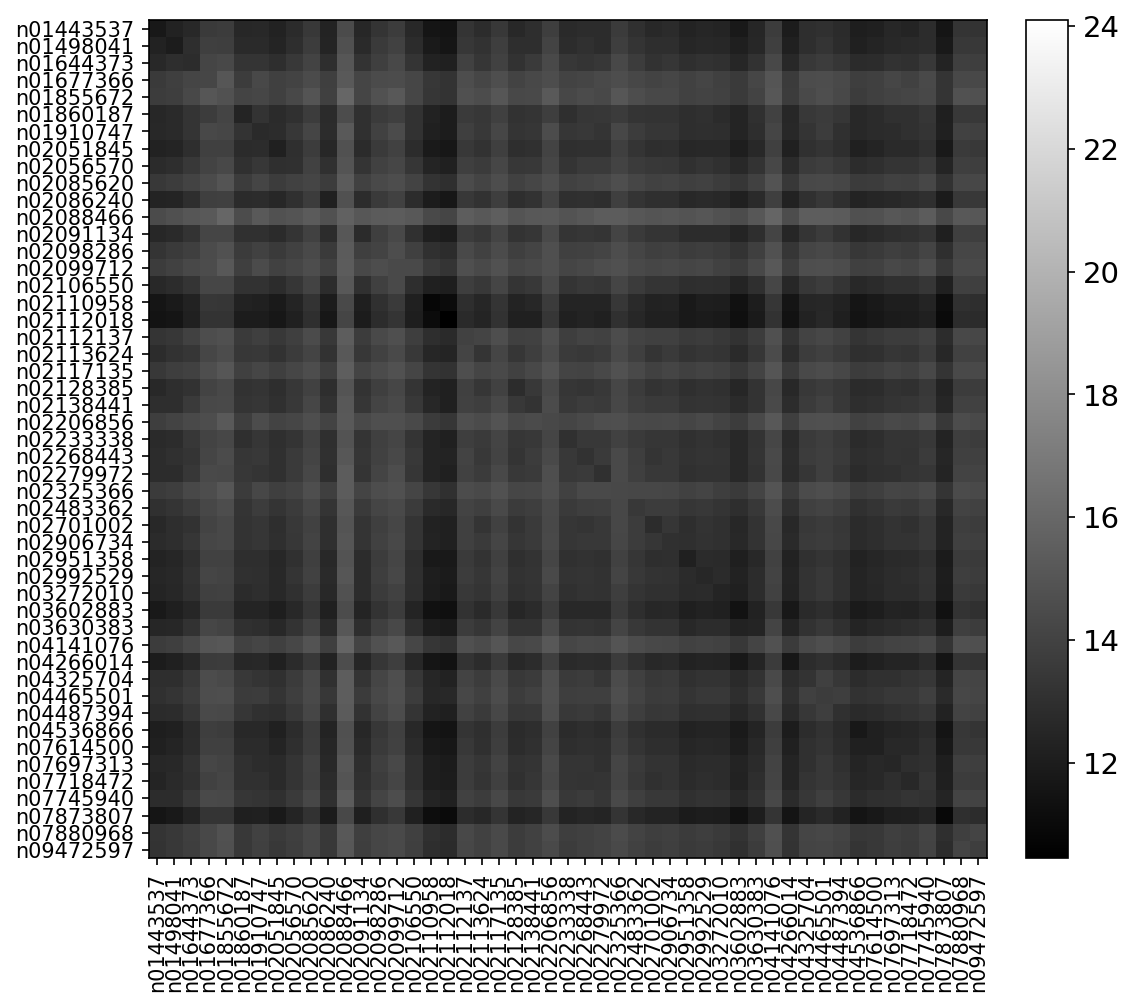}
                    \caption{Resnet50-Normal Labels}
                \end{subfigure}
                \begin{subfigure}{0.40\textwidth}
                    \includegraphics[width=\linewidth]{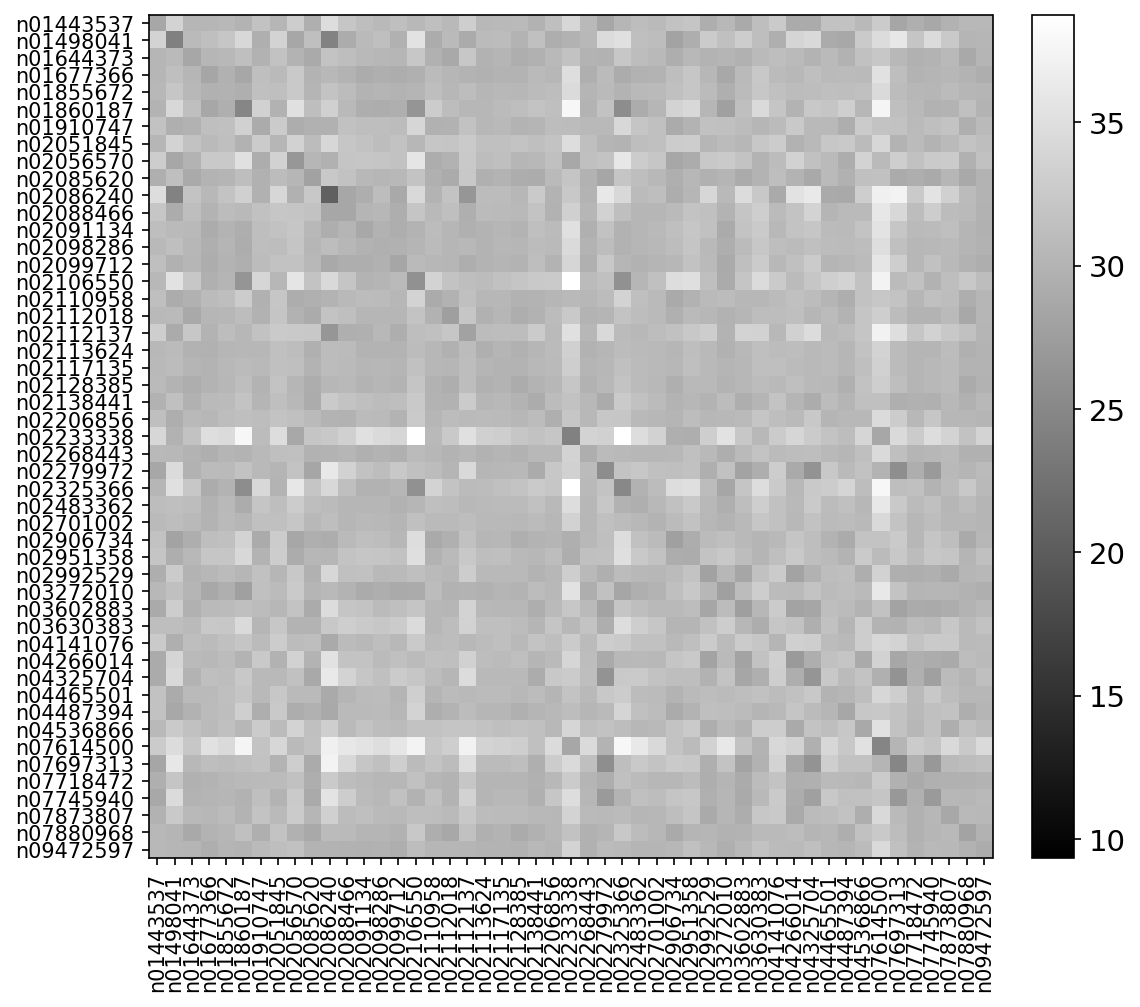}
                    \caption{ViT-B/32-Untrained}
                \end{subfigure}
                \begin{subfigure}{0.40\textwidth}
                    \includegraphics[width=\linewidth]{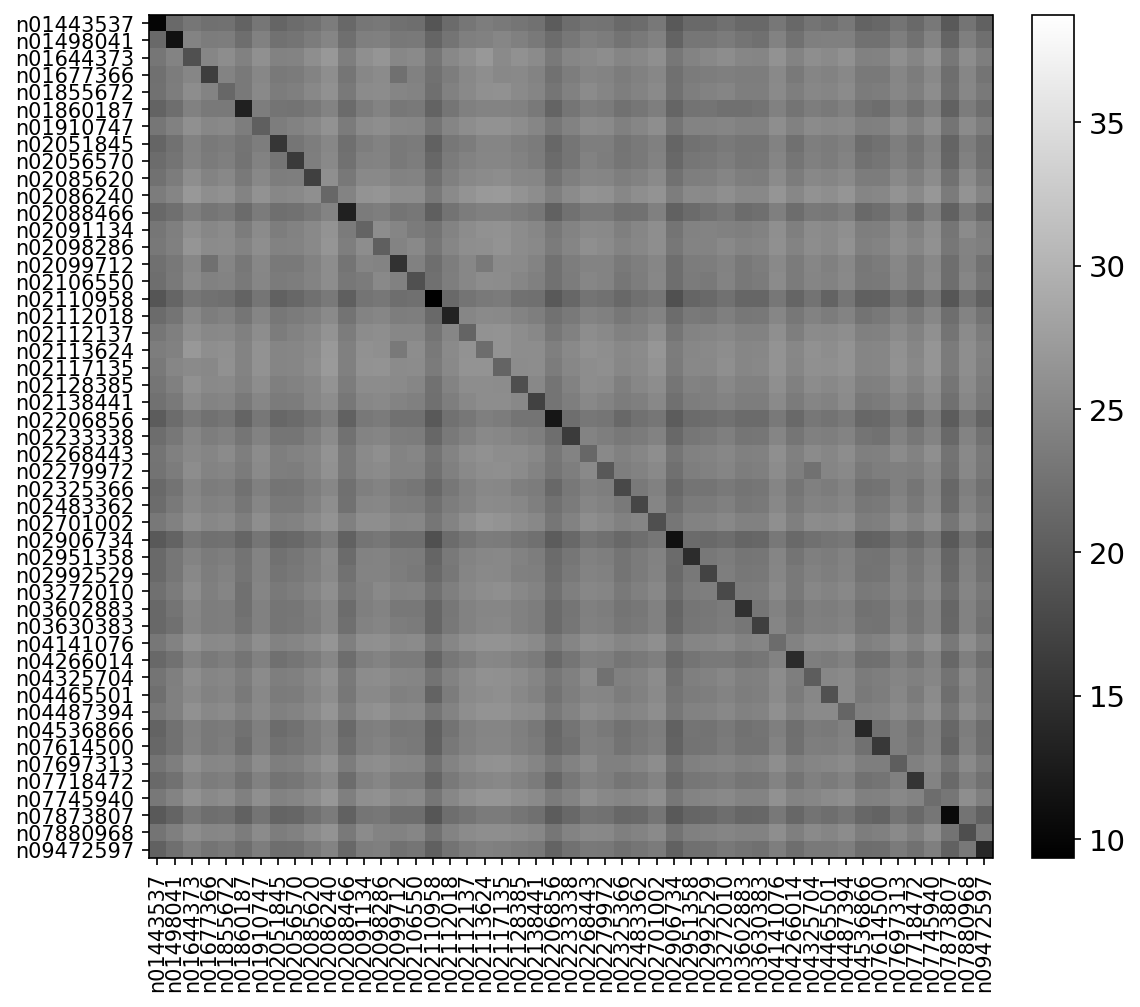}
                    \caption{ViT-B/32-Normal Labels}
                \end{subfigure}
                \caption{Pairwise Energy Distance Heatmaps for ResNet-50 and ViT-B/32 on ImageNet-R. The untrained network shows a much larger inter-class distance than the trained network}
                \label{fig:resnet_vit_b_32_energy}
            \end{figure}
            
            \begin{figure}[hbtp]
                \centering
                \begin{subfigure}{0.32\textwidth}
                    \includegraphics[width=\linewidth]{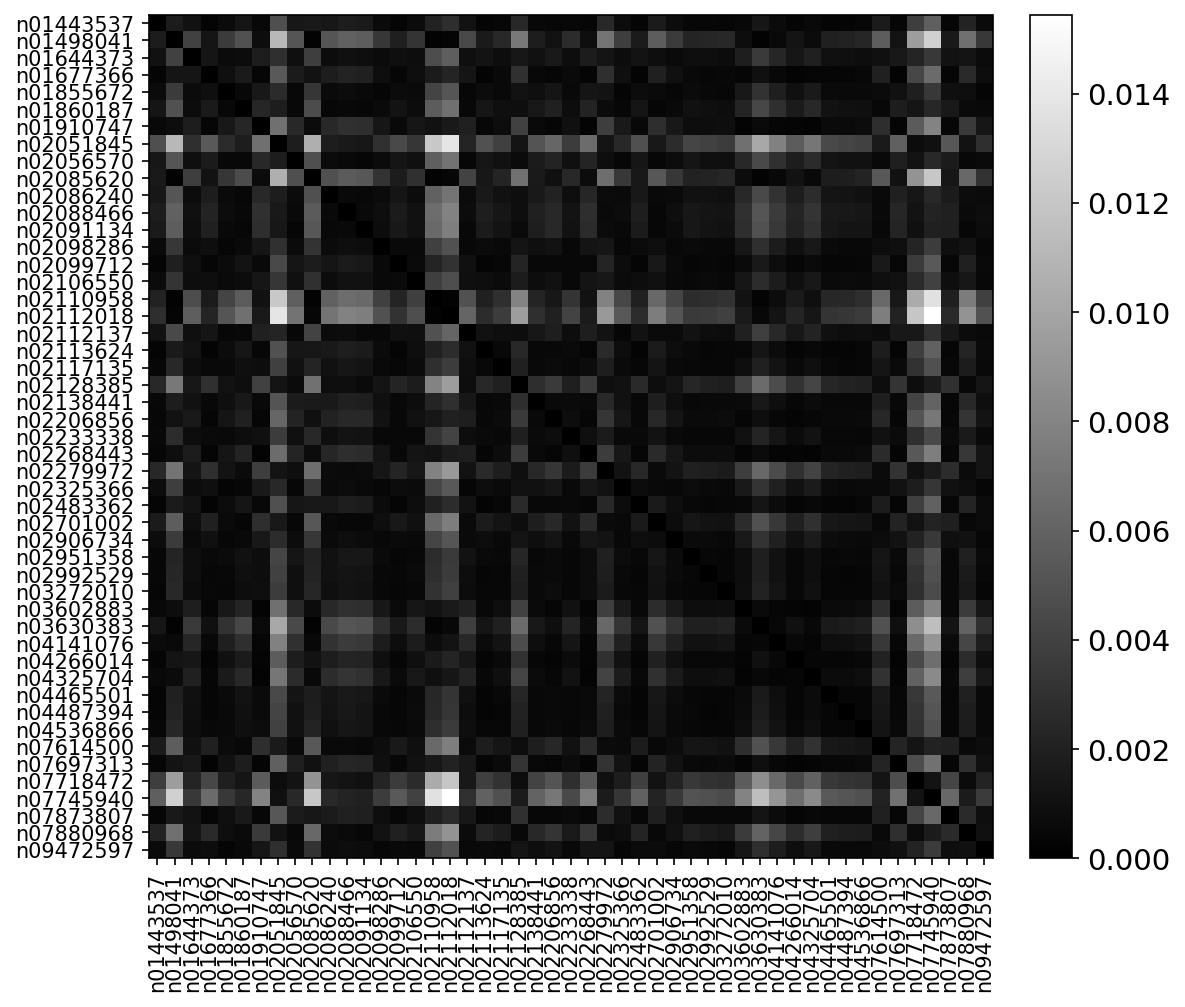}
                    \caption{Untrained}
                \end{subfigure}
                \begin{subfigure}{0.32\textwidth}
                    \includegraphics[width=\linewidth]{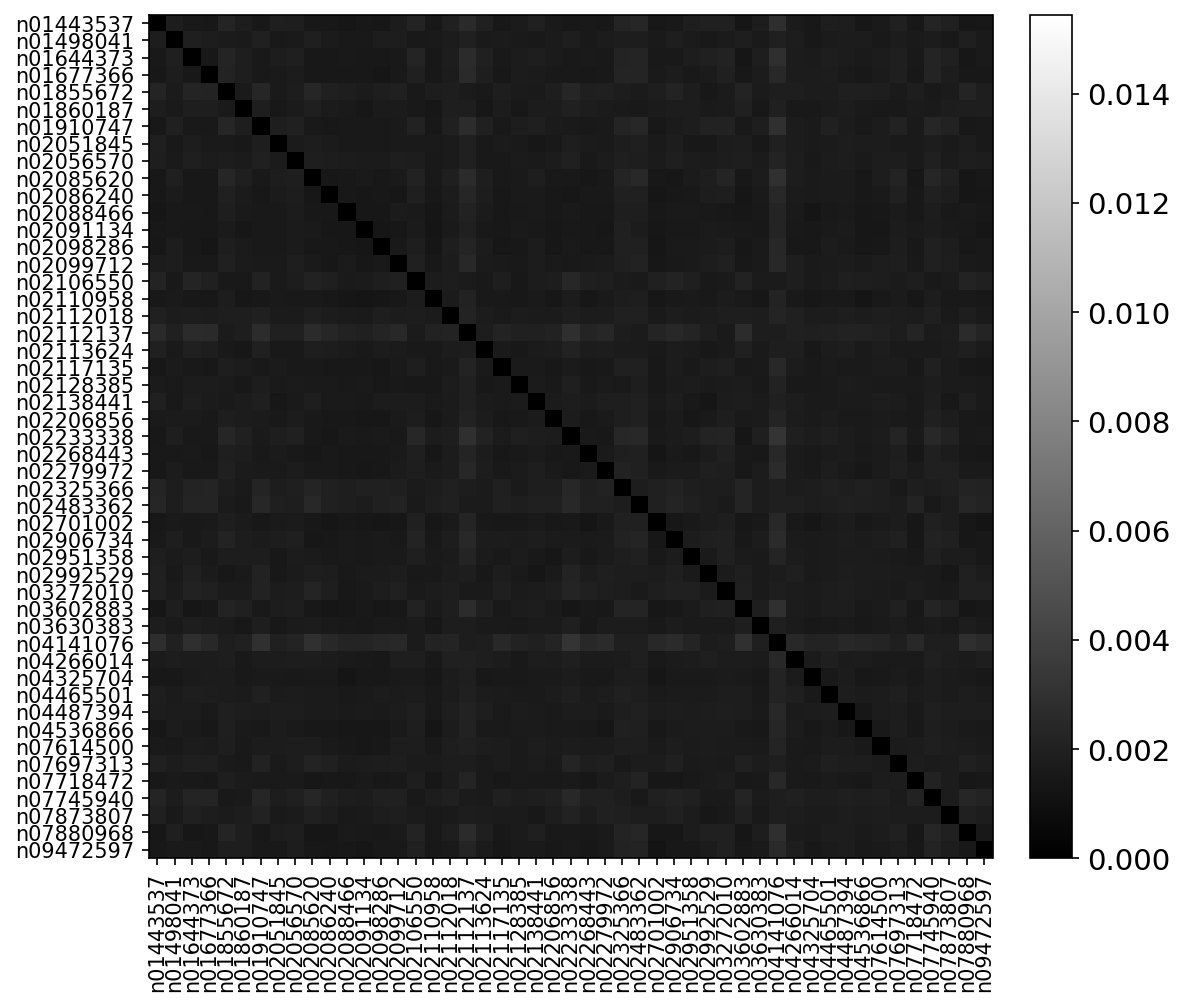}
                    \caption{Random Labels}
                \end{subfigure}
                \begin{subfigure}{0.32\textwidth}
                    \includegraphics[width=\linewidth]{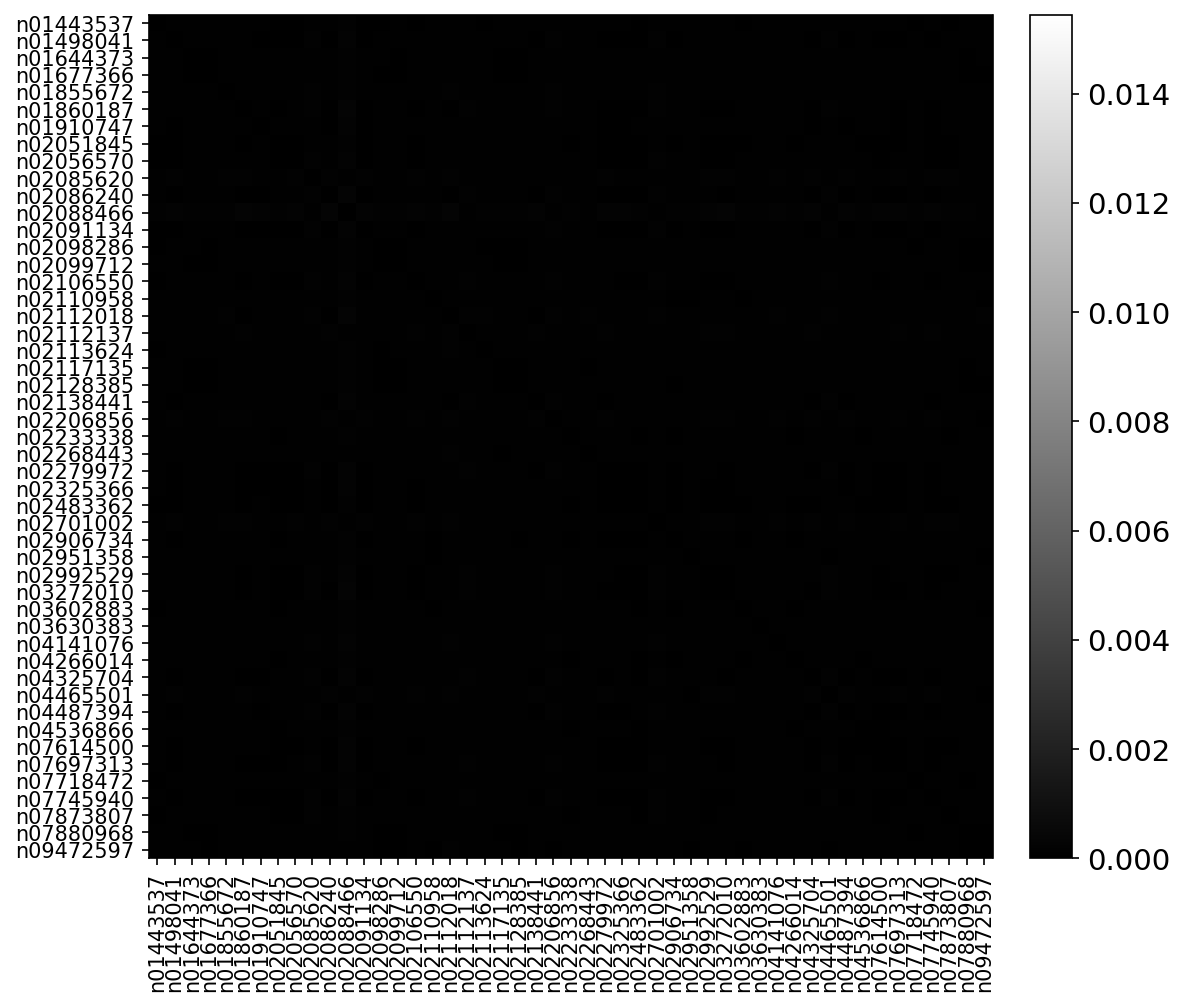}
                    \caption{Normal Labels}
                \end{subfigure}
                \caption{Heatmaps of the pairwise KL divergence between softmaxed prototype neuron-weight interactions for InceptionV3 on ImageNet-R. The untrained network has the largest inter-class distance, while the random labels network has a larger inter-class distance than the normal labels network}
                \label{fig:inceptionv3_softmax_Kl}
            \end{figure}

            \begin{figure}[hbtp]
                \centering
                \begin{subfigure}{0.40\textwidth}
                    \includegraphics[width=\linewidth]{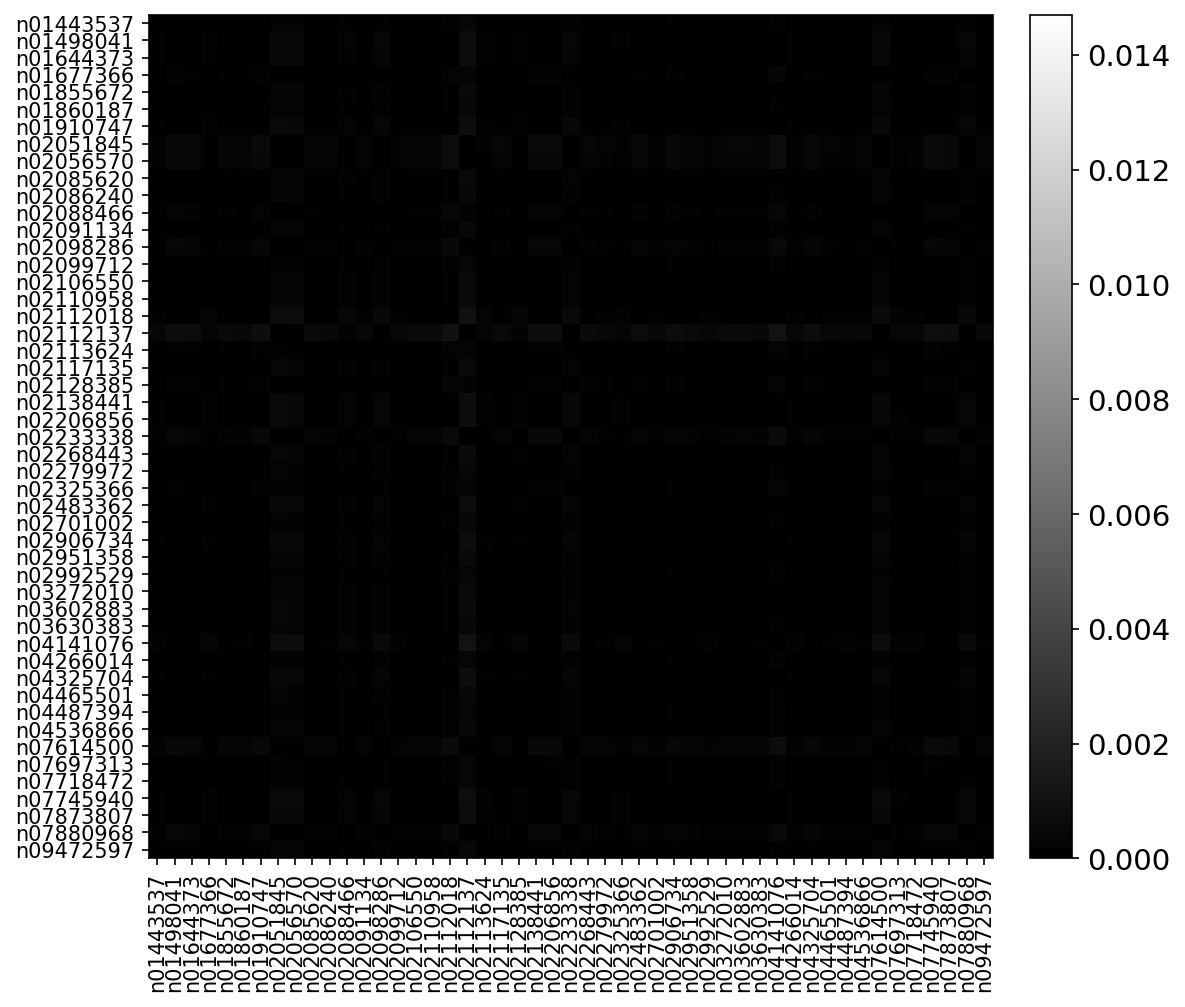}
                    \caption{ResNet50-Untrained}
                \end{subfigure}
                \begin{subfigure}{0.40\textwidth}
                    \includegraphics[width=\linewidth]{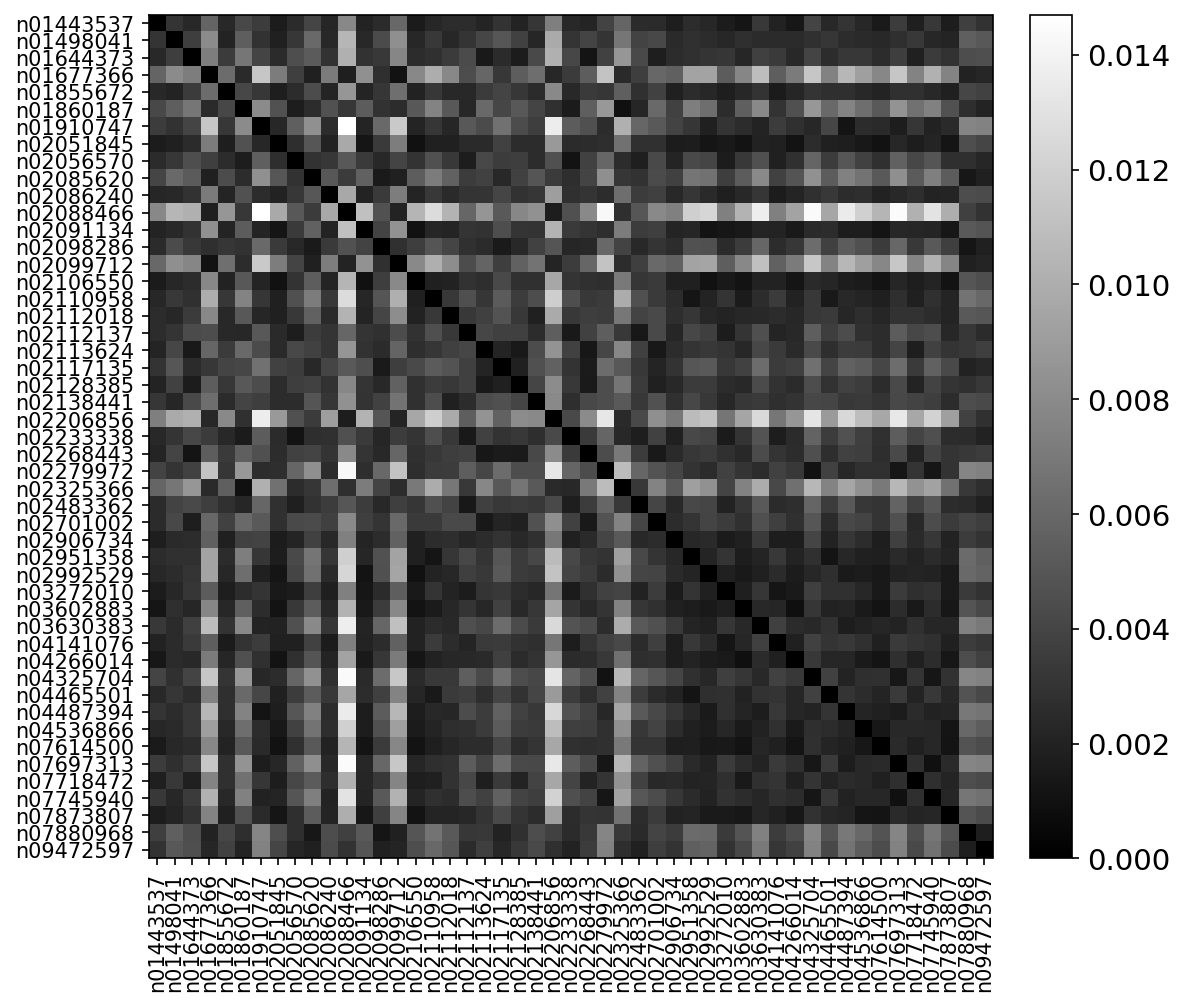}
                    \caption{ResNet50-Normal Labels}
                \end{subfigure}
                \begin{subfigure}{0.40\textwidth}
                    \includegraphics[width=\linewidth]{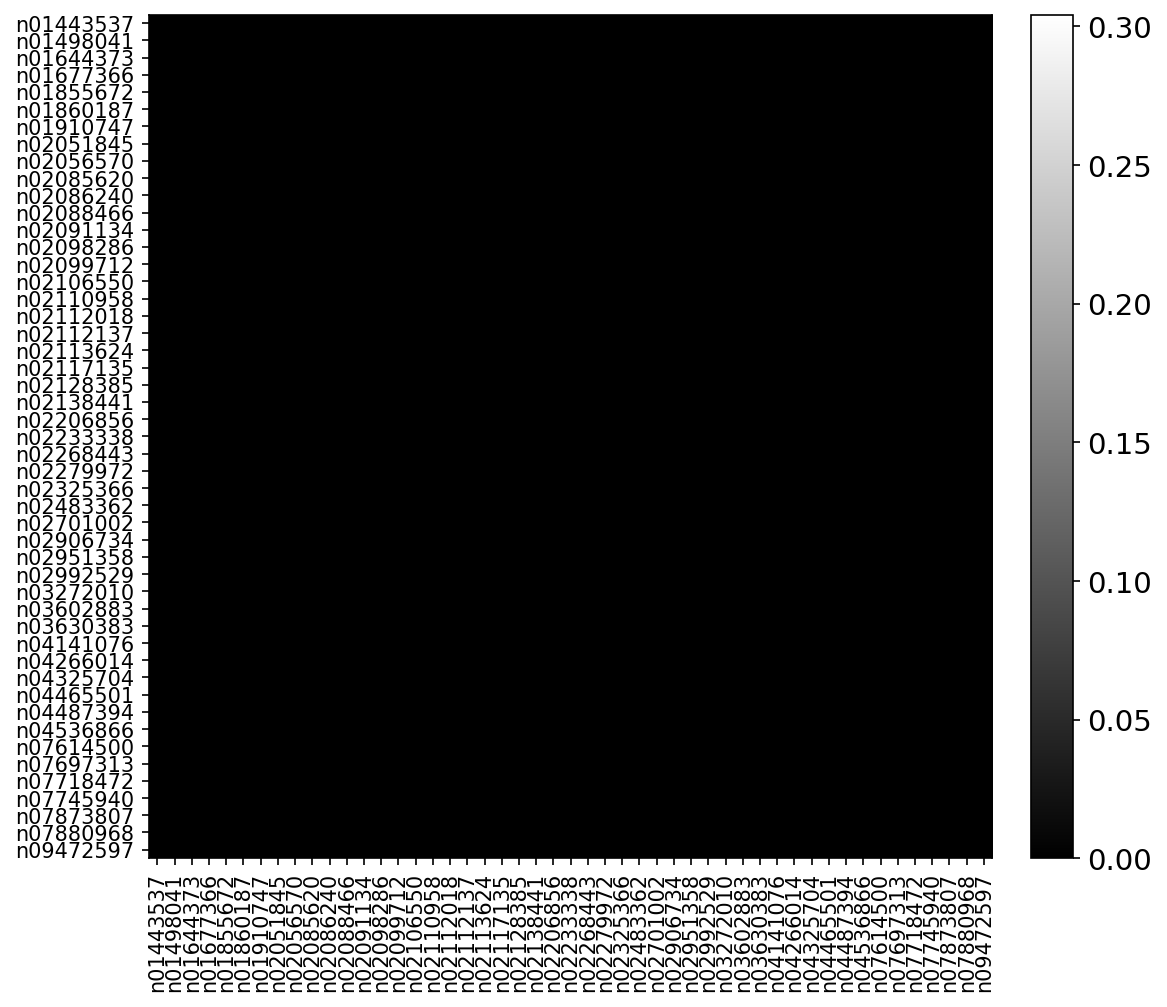}
                    \caption{ViT-B/32-Untrained}
                \end{subfigure}
                \begin{subfigure}{0.40\textwidth}
                    \includegraphics[width=\linewidth]{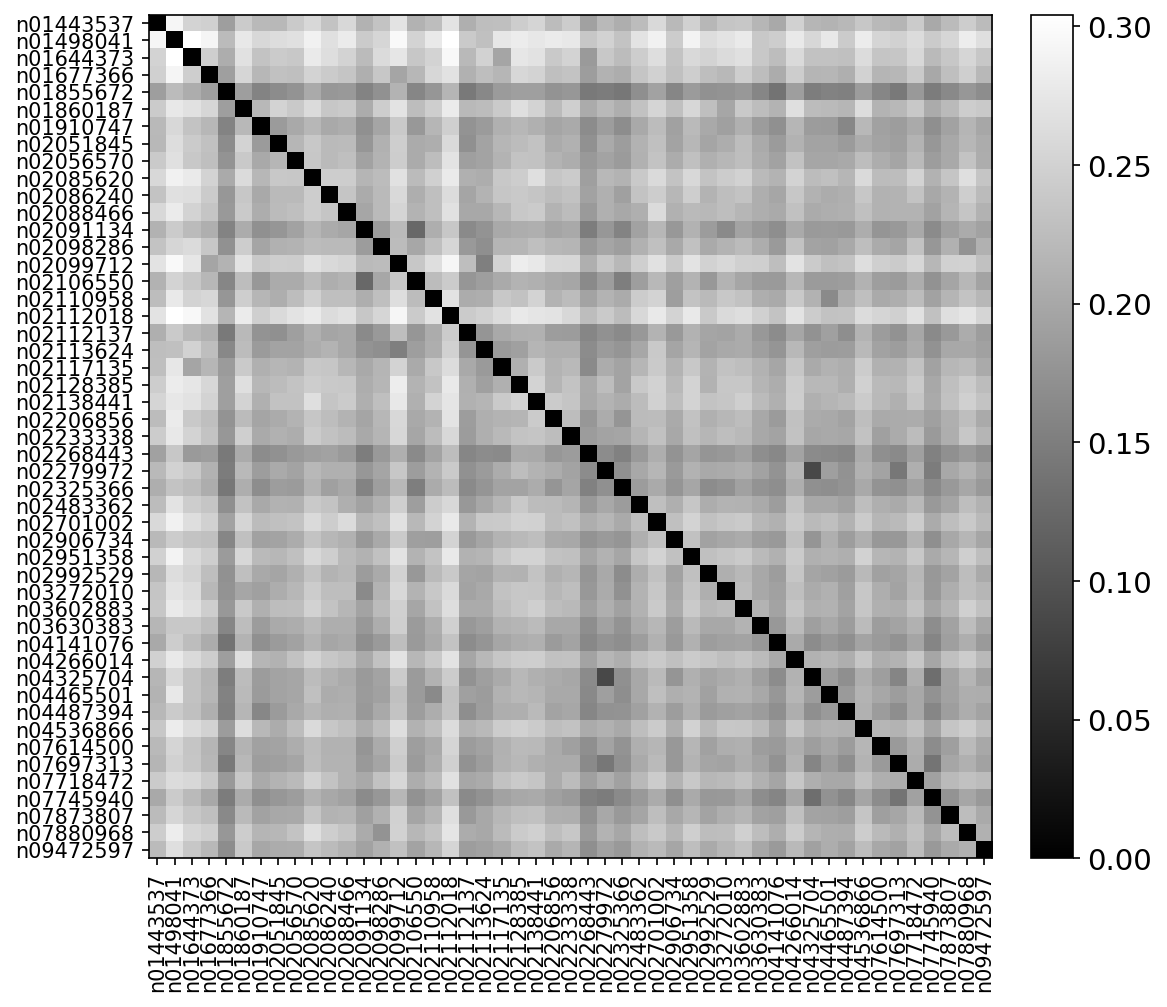}
                    \caption{ViT-B/32-Normal Labels}
                \end{subfigure}
                \caption{Heatmaps of the pairwise KL divergence between softmaxed prototype neuron-weight interactions for ResNet50 and ViT-B/32 on ImageNet-R. Though both of these networks show an increase in inter-class distance through training, the fact that InceptionV3 does not show the same pattern suggests that this metric is not a reliable diagnostic of robustness.}
                \label{fig:resnet50_vit_b_32_softmax_Kl}
            \end{figure}
    
    \section{Results for CIFAR10}
\begin{figure}[hbtp]
            \centering
            \begin{subfigure}{0.32\textwidth}
                \includegraphics[width=\linewidth]{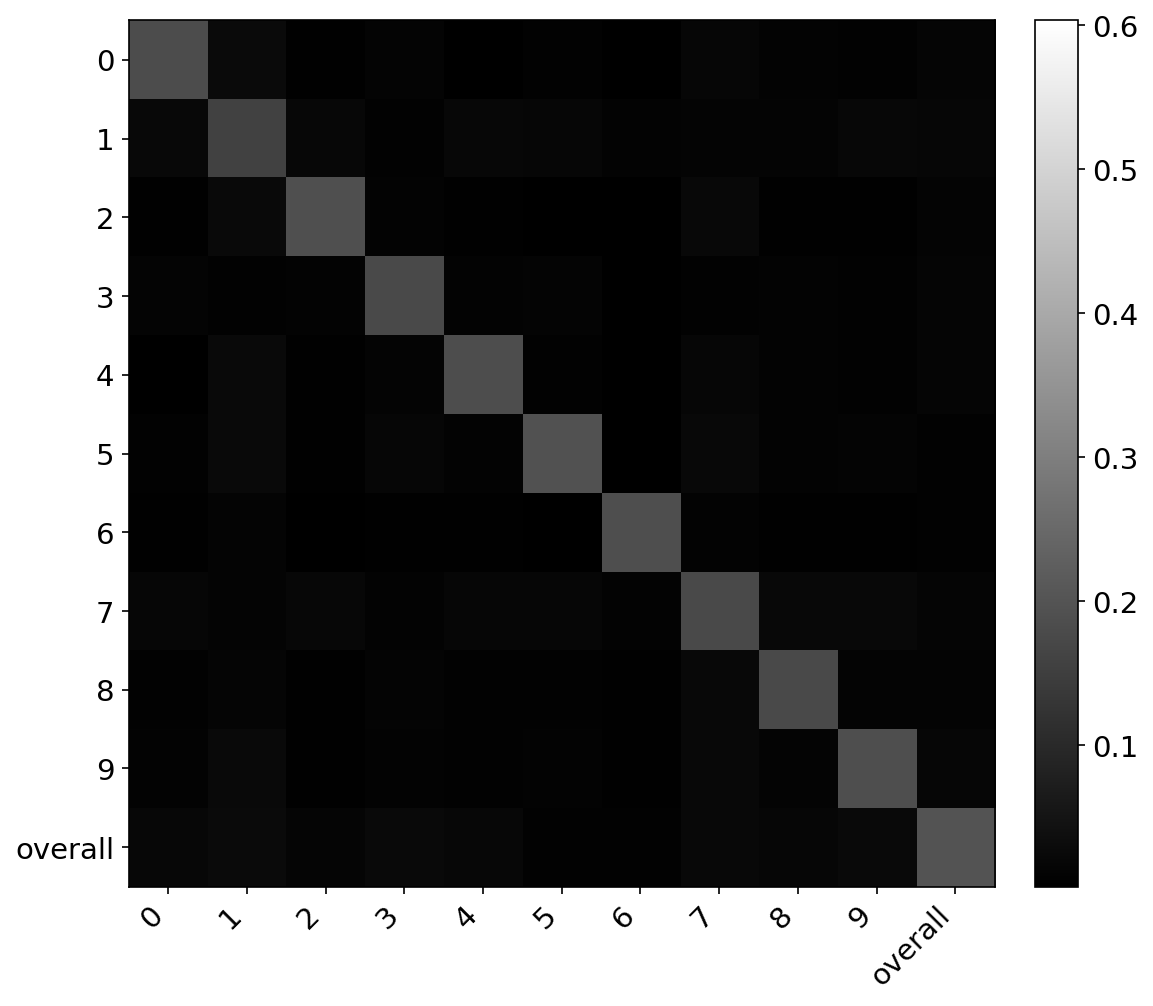}
                \caption{Initialisation}
            \end{subfigure}
            \begin{subfigure}{0.32\textwidth}
                \includegraphics[width=\linewidth]{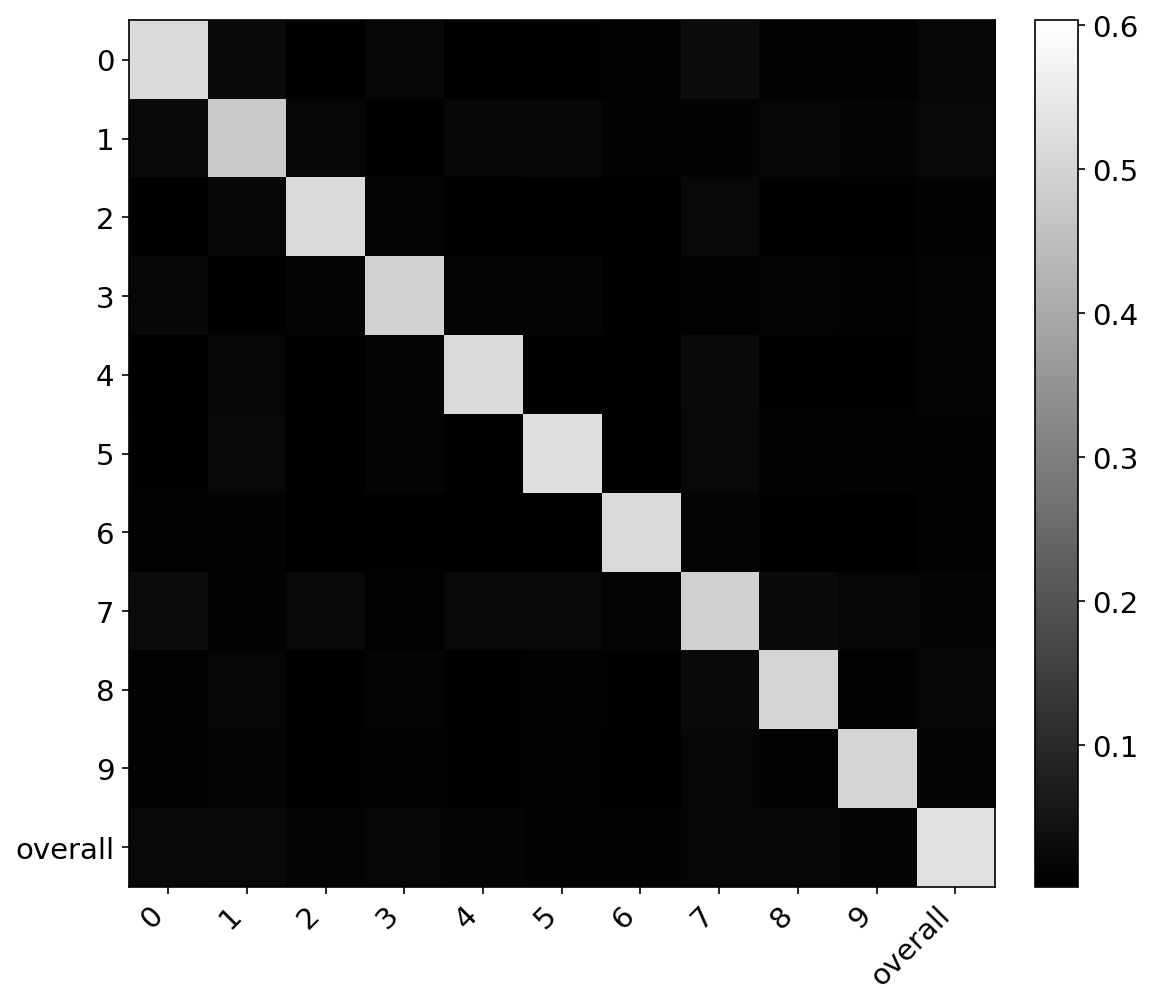}
                \caption{Random Labels}
            \end{subfigure}
            \begin{subfigure}{0.32\textwidth}
                \includegraphics[width=\linewidth]{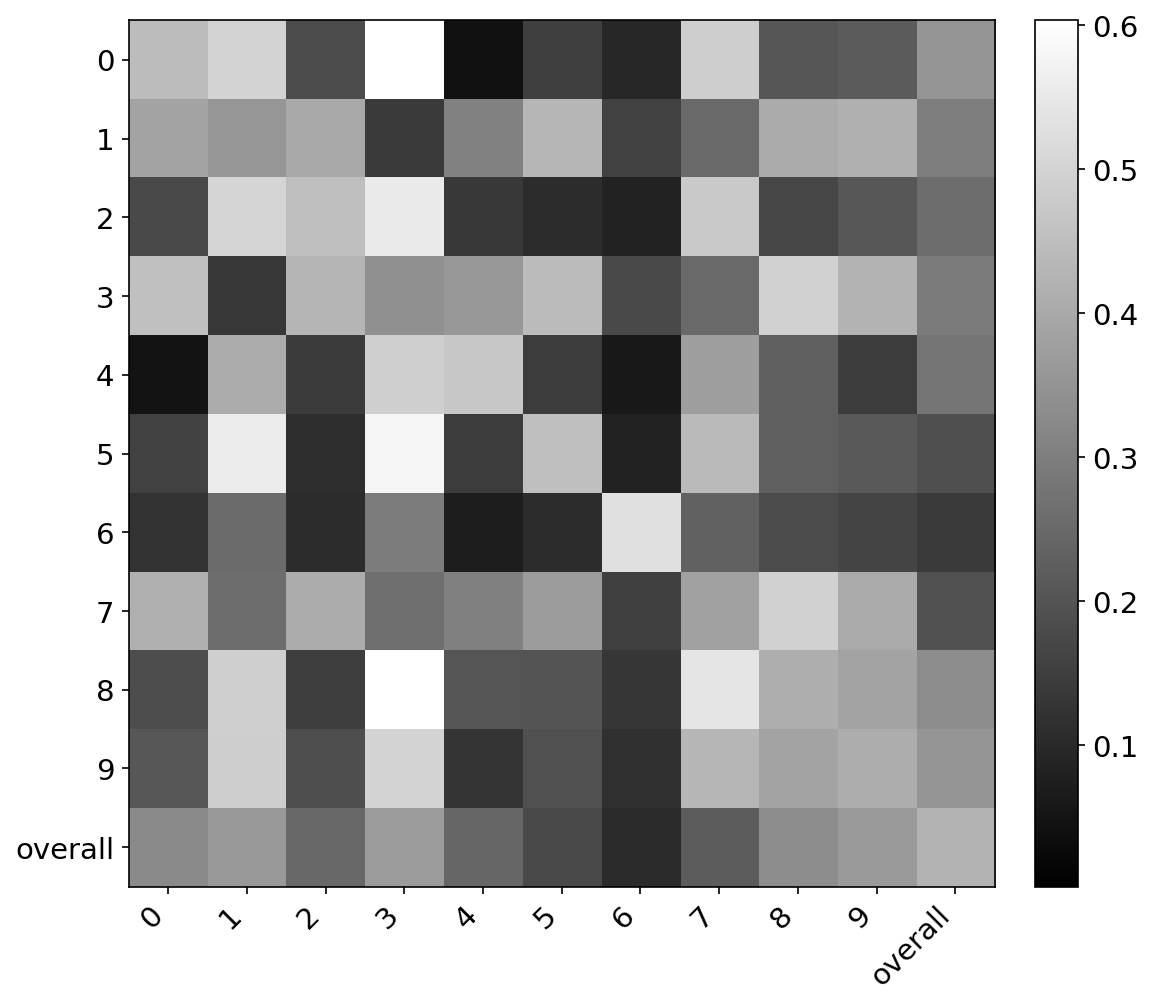}
                \caption{Normal Labels}
            \end{subfigure}
            \caption{Pairwise KL Divergence Heatmaps for Modified Alexnet on CIFAR10}
            \label{fig:cifar10_results}
        \end{figure}

\begin{figure}[hbtp]
            \centering
            \begin{subfigure}{0.32\textwidth}
                \includegraphics[width=\linewidth]{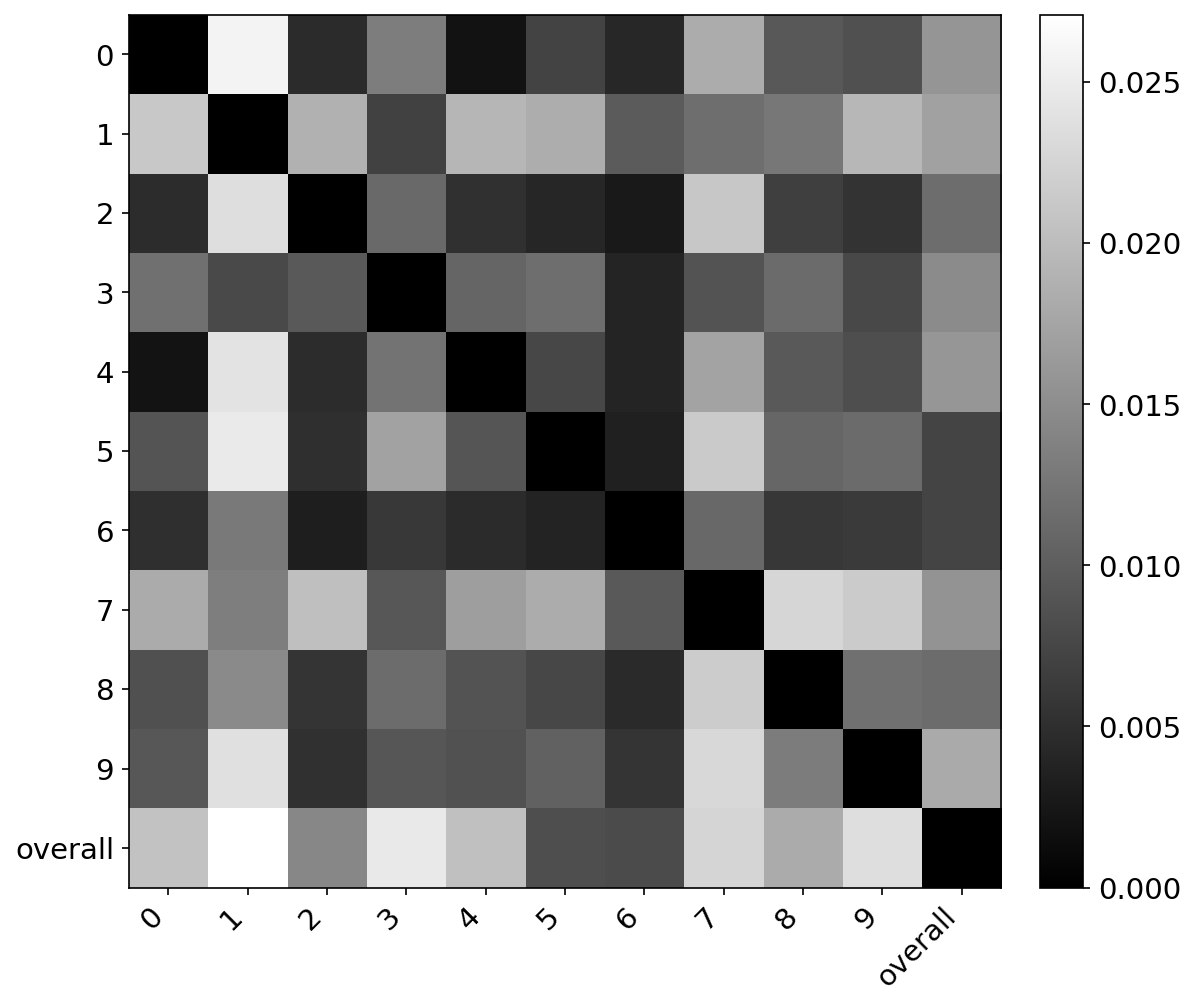}
                \caption{Initialisation}
            \end{subfigure}
            \begin{subfigure}{0.32\textwidth}
                \includegraphics[width=\linewidth]{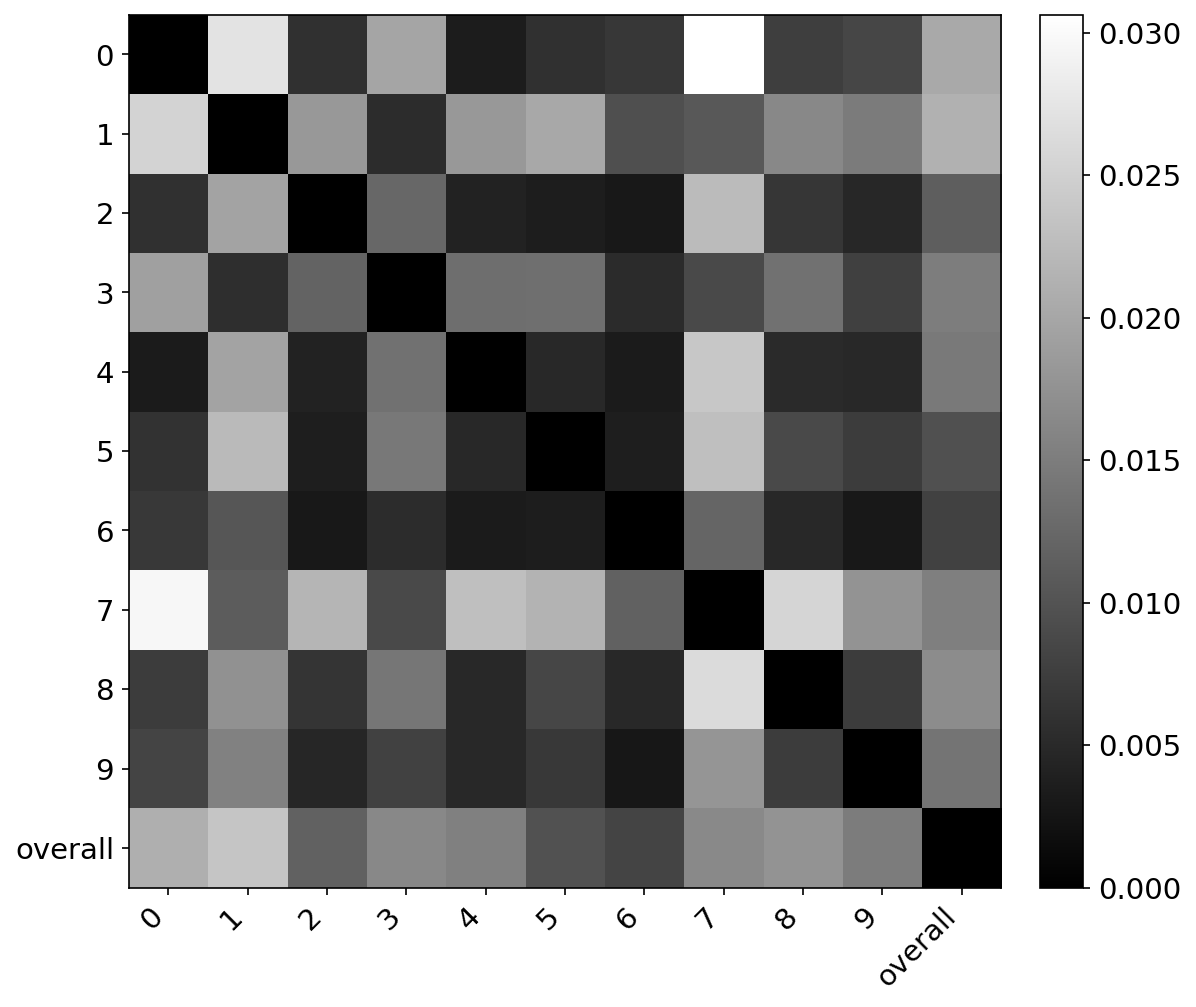}
                \caption{Random Labels}
            \end{subfigure}
            \begin{subfigure}{0.32\textwidth}
                \includegraphics[width=\linewidth]{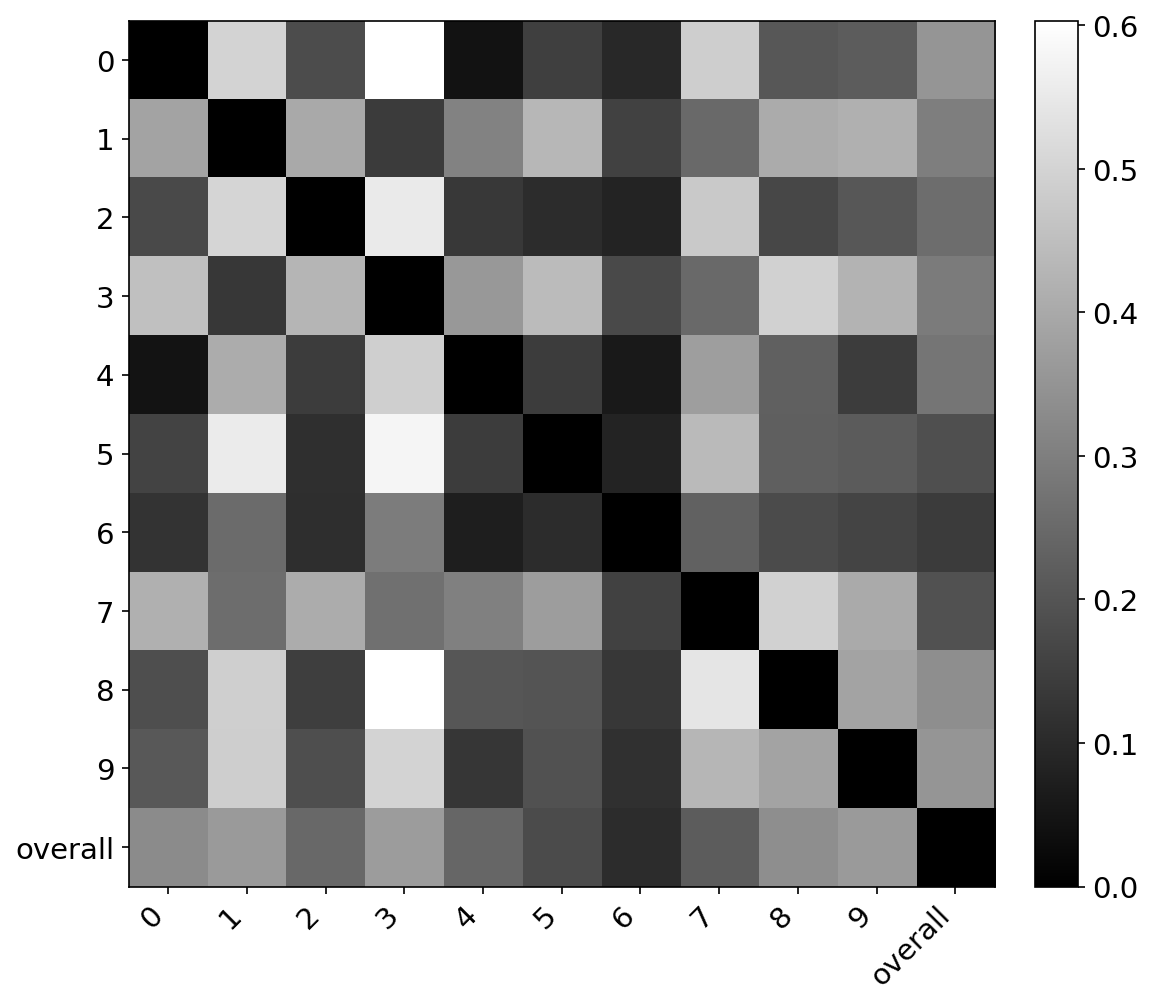}
                \caption{Normal Labels}
            \end{subfigure}
            \caption{Unnormalised Pairwise KL Divergence Heatmaps for Modified Alexnet on CIFAR10}
            \label{fig:cifar10_results_unnorm}
        \end{figure}

        The comparison heatmaps for CIFAR10 are shown in figure \ref{fig:cifar10_results}. Once again, we see that the random labels case is almost identical to the initialised network, while the normal labels case shows a marked increase in inter-class distances alongside a decrease in entropy. In the unnormalised case (Figure \ref{fig:cifar10_results_unnorm}), we see that the inter-class distances are much smaller than in the ImageNet case, which may be a result of the smaller number of classes and the smaller size of the network. The unnormalised datasets show a much more pronounced version of the ImageNet results, where classes which are close at initialisation remain close at convergence, while those which are far apart remain far apart. Class 6 is a lot darker than the other classes, which is a result of the class having high entropy. That makes it closer to the other classes in the sense of the KL divergence.
        Table \ref{tab:cifar10_Bernoulli_summary} provides the corresponding scalar summaries at the last and second-last layers. As in ImageNet, the trained model exhibits much larger mean inter-class KL than untrained and random-label checkpoints. On TinyImageNet, the inter-class mean is lower than the in-distribution trained model, consistent with reduced separation under distribution shift.

        \begin{table}[hbtp]
            \centering
            \begin{tabular}{lcc}
                \toprule
                Condition & Mean inter-class KL & Mean class entropy \\
                \midrule
                Untrained (CIFAR10) & 0.0121 & 0.1825 \\
                Random labels (CIFAR10) & 0.0119 & 0.5077 \\
                Normal labels (CIFAR10) & 0.2822 & 0.4248 \\
                \midrule
                Untrained (CIFAR10, second-last) & 0.0208 & 0.3381 \\
                Random labels (CIFAR10, second-last) & 0.0184 & 0.5341 \\
                Normal labels (CIFAR10, second-last) & 0.1228 & 0.5113 \\
                \bottomrule
            \end{tabular}
            \caption{Summary statistics for Small AlexNet Bernoulli KL divergences on CIFAR10. The key difference between the last and second last layers is the scale of the entropy, which is much higher for the second last layer relative to it's inter-class KL.}
            \label{tab:cifar10_Bernoulli_summary}
        \end{table}

        \begin{table}[hbtp]
            \centering
            \begin{tabular}{lcc}
                \toprule
                Condition & Mean inter-class KL & Mean class entropy \\
                \midrule
                Untrained (CIFAR10.1) & 0.0121 & 0.1800 \\
                Random labels (CIFAR10.1) & 0.0109 & 0.5023 \\
                Normal labels (CIFAR10.1) & 0.1933 & 0.4513 \\
                \midrule
                Untrained (CIFAR10.1, second-last) & 0.0176 & 0.3418 \\
                Random labels (CIFAR10.1, second-last) & 0.0167 & 0.5370 \\
                Normal labels (CIFAR10.1, second-last) & 0.0865 & 0.5250 \\
                \bottomrule
            \end{tabular}
            \caption{Summary statistics for Small AlexNet Bernoulli KL divergences on CIFAR10.1. The results are extremely similar to the CIFAR10 case which is to be expected given that CIFAR10.1 was intended to be a new test set for CIFAR10}
            \label{tab:cifar10.1_Bernoulli_summary}
        \end{table}

        \begin{table}[hbtp]
            \centering
            \begin{tabular}{lcc}
                \toprule
                Condition & Mean inter-class KL & Mean class entropy \\
                \midrule
                Untrained (TinyImageNet) & 0.0138 & 0.1822 \\
                Random labels (TinyImageNet) & 0.0119 & 0.4969 \\
                Normal labels (TinyImageNet) & 0.1164 & 0.4602 \\
                \midrule
                Untrained (TinyImageNet, second-last) & 0.0208 & 0.3072 \\
                Random labels (TinyImageNet, second-last) & 0.0150 & 0.5452 \\
                Normal labels (TinyImageNet, second-last) & 0.0610 & 0.4992 \\
                \bottomrule
            \end{tabular}
            \caption{Summary statistics for Small AlexNet Bernoulli KL divergences on TinyImageNet, mapped to CIFAR10. The results are extremely similar to the TinyImageNet case which is to be expected given that TinyImageNet was intended to be a new test set for TinyImageNet}
            \label{tab:tinyimagenet_Bernoulli_summary}
        \end{table}

\begin{figure}[hbtp]
            \centering
            \begin{subfigure}{0.32\textwidth}
                \includegraphics[width=\linewidth]{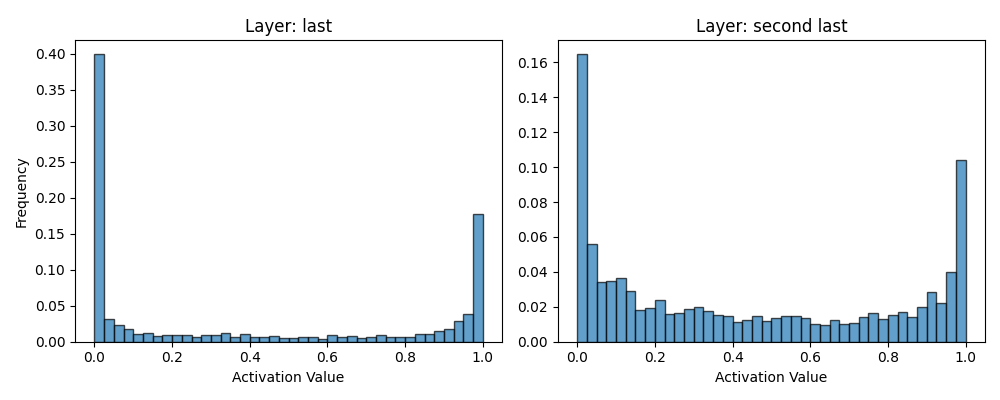}
                \caption{Initialisation}
            \end{subfigure}
            \begin{subfigure}{0.32\textwidth}
                \includegraphics[width=\linewidth]{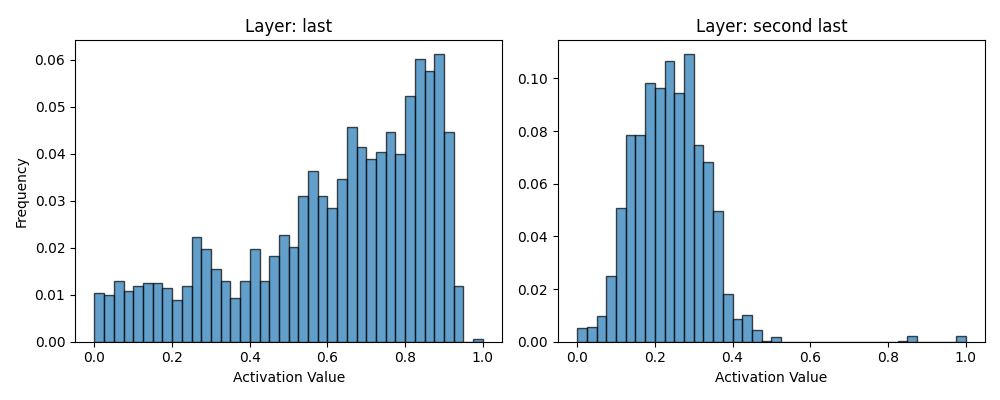}
                \caption{Random Labels}
            \end{subfigure}
            \begin{subfigure}{0.32\textwidth}
                \includegraphics[width=\linewidth]{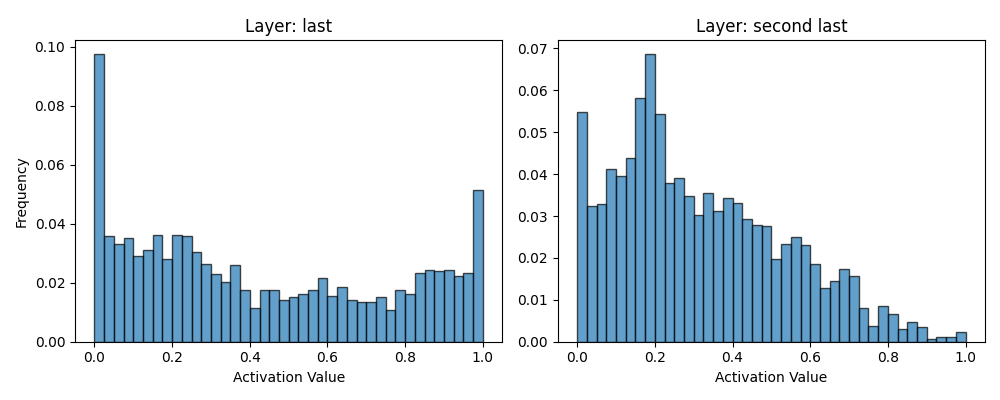}
                \caption{Normal Labels}
            \end{subfigure}
            \caption{Activation Sparsity Histograms for Alexnet on CIFAR10 for the class 'airplane'}
            \label{fig:alexnet_sparsity}
        \end{figure}

\begin{figure}[hbtp]
            \centering
            \begin{subfigure}{0.32\textwidth}
                \includegraphics[width=\linewidth]{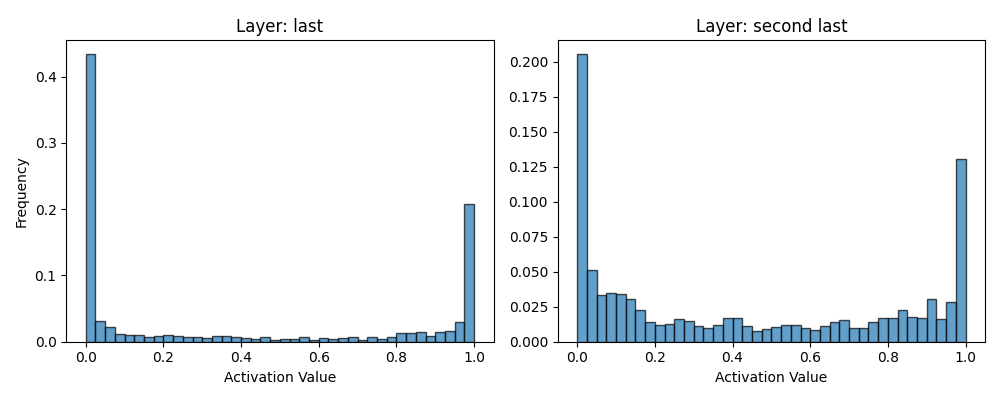}
                \caption{Initialisation}
            \end{subfigure}
            \begin{subfigure}{0.32\textwidth}
                \includegraphics[width=\linewidth]{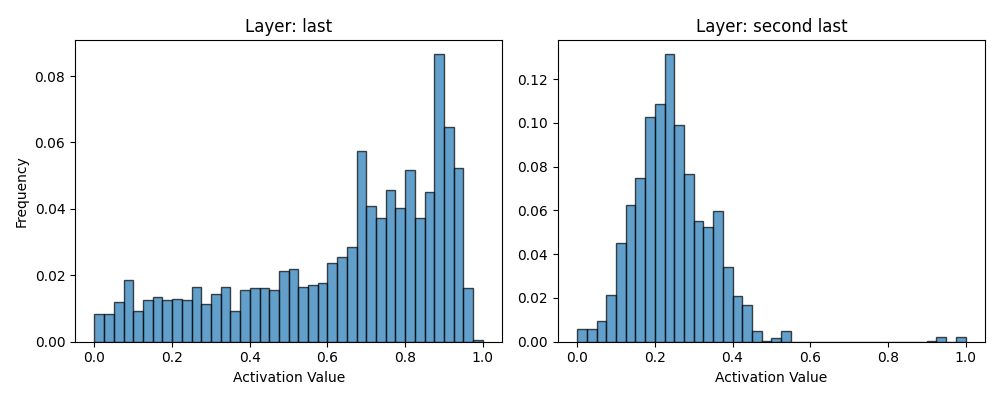}
                \caption{Random Labels}
            \end{subfigure}
            \begin{subfigure}{0.32\textwidth}
                \includegraphics[width=\linewidth]{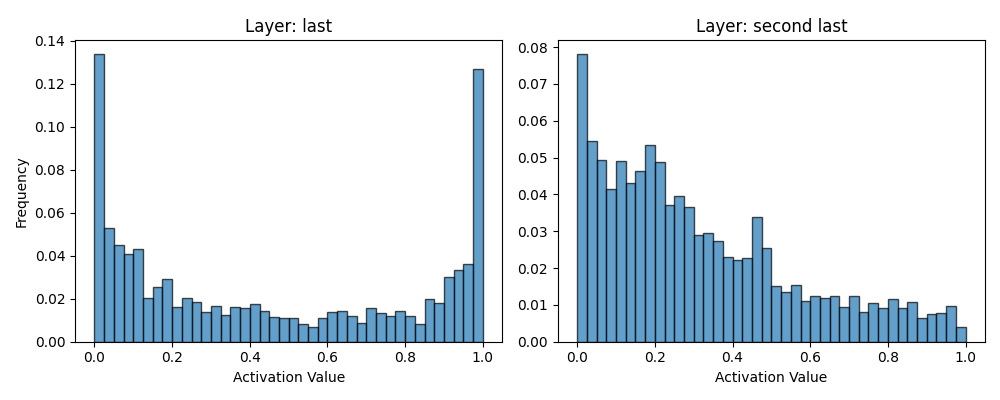}
                \caption{Normal Labels}
            \end{subfigure}
            \caption{Activation Sparsity Histograms for Alexnet on CIFAR10 for the class 'automobile'}
            \label{fig:alexnet_sparsity_auto}
        \end{figure}

        Sparsity results for CIFAR10 are shown in figures \ref{fig:alexnet_sparsity} and \ref{fig:alexnet_sparsity_auto}. The behaviour of this modified Alexnet is markedly different from InceptionV3, with some similarity maintained. The density of the network when trained on random labels is certainly higher than at initialisation, but is not nearly as pronounced as in the InceptionV3 case. Additionally, the sparsity in figures \ref{fig:alexnet_sparsity} and \ref{fig:alexnet_sparsity_auto} decreases even when training with normal labels, although the network is still extremely sparse overall, with an almost bimodal distribution with two peaks at both ends, and most of the mass concentrated in between. This may lend credence to the idea of a set of 'grandmother' paths at the last layer of the network, in that only certain neurons are of importance to any given class, while the rest are largely irrelevant.

\begin{figure}[hbtp]
            \centering
            \begin{subfigure}{0.32\textwidth}
                \includegraphics[width=\linewidth]{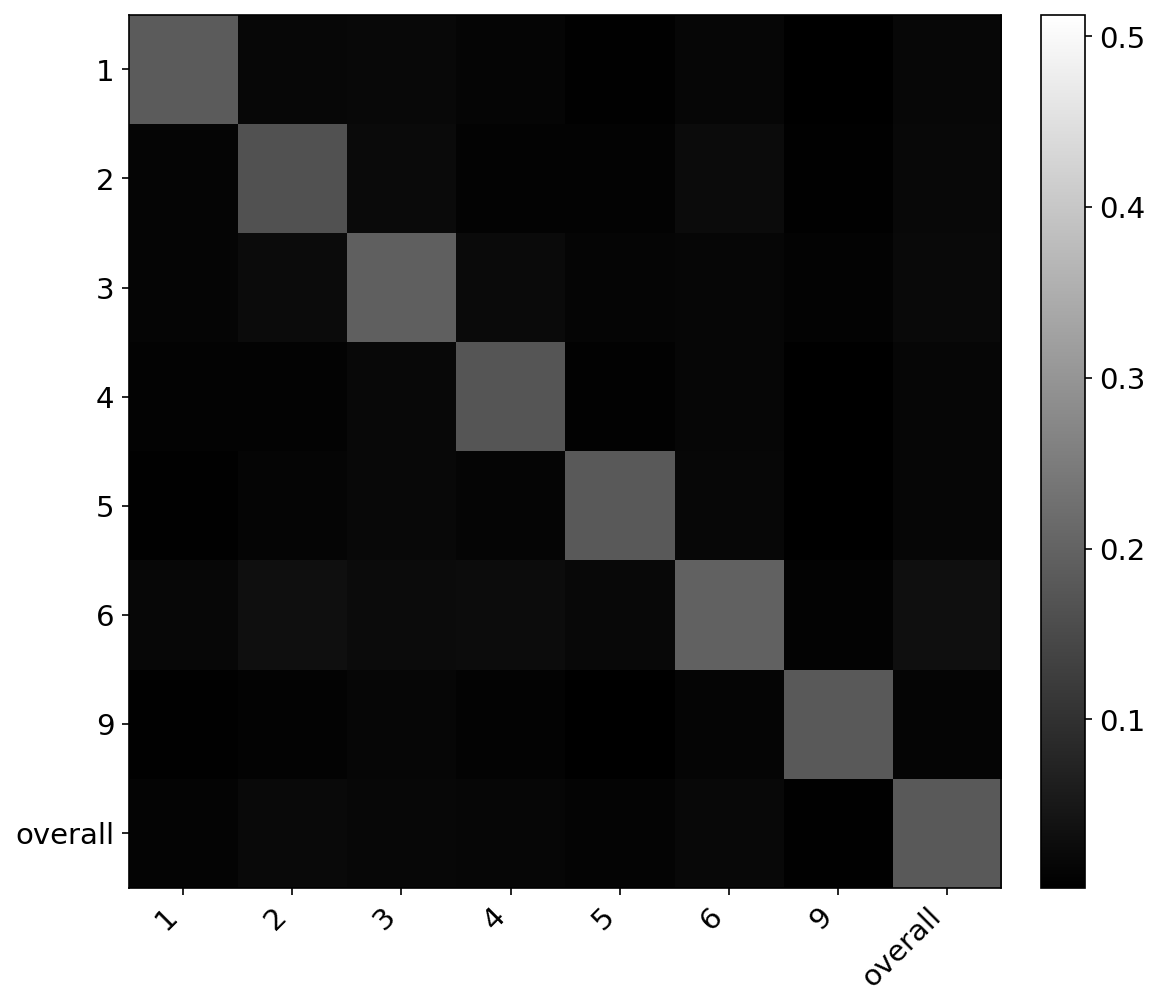}
                \caption{Initialisation}
            \end{subfigure}
            \begin{subfigure}{0.32\textwidth}
                \includegraphics[width=\linewidth]{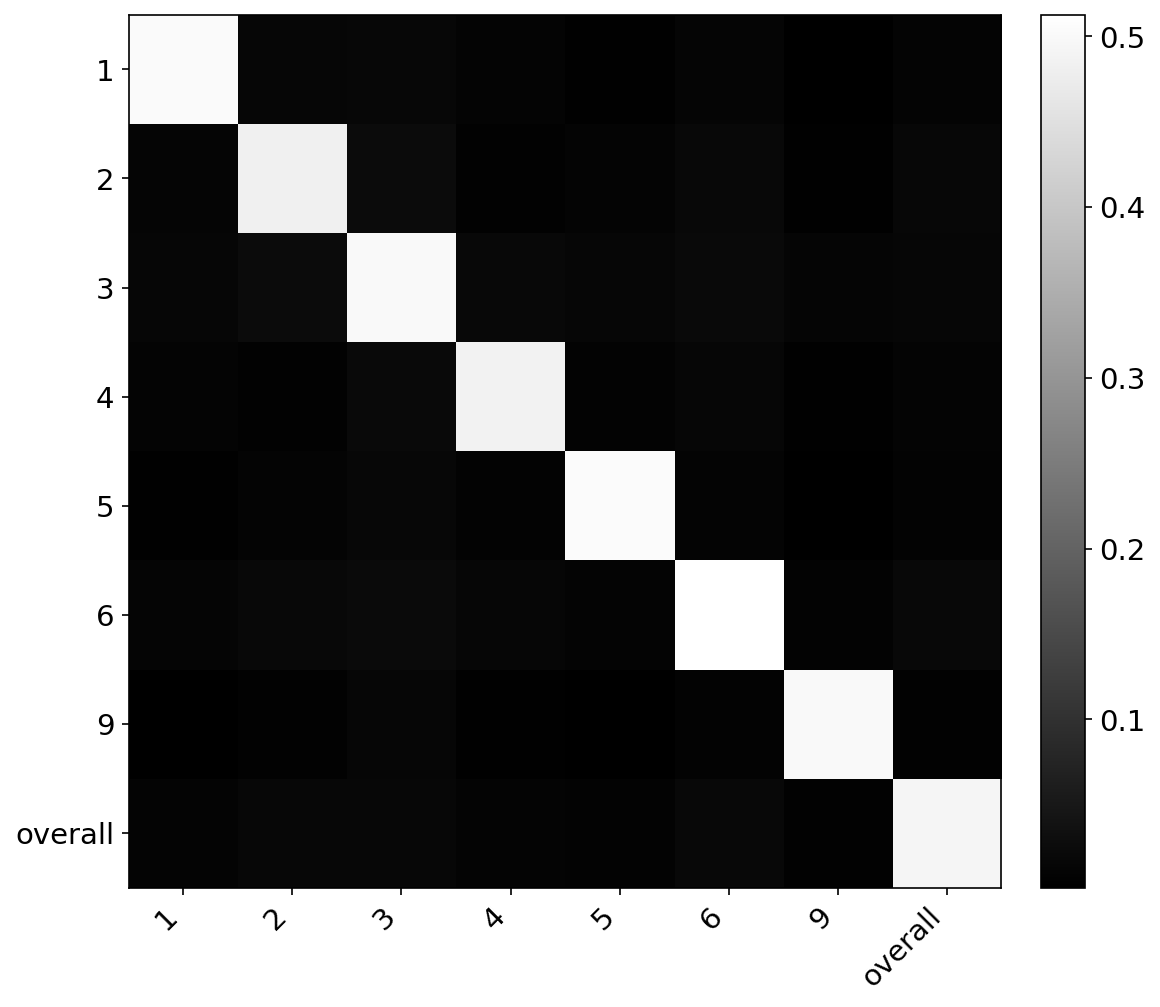}
                \caption{Random Labels}
            \end{subfigure}
            \begin{subfigure}{0.32\textwidth}
                \includegraphics[width=\linewidth]{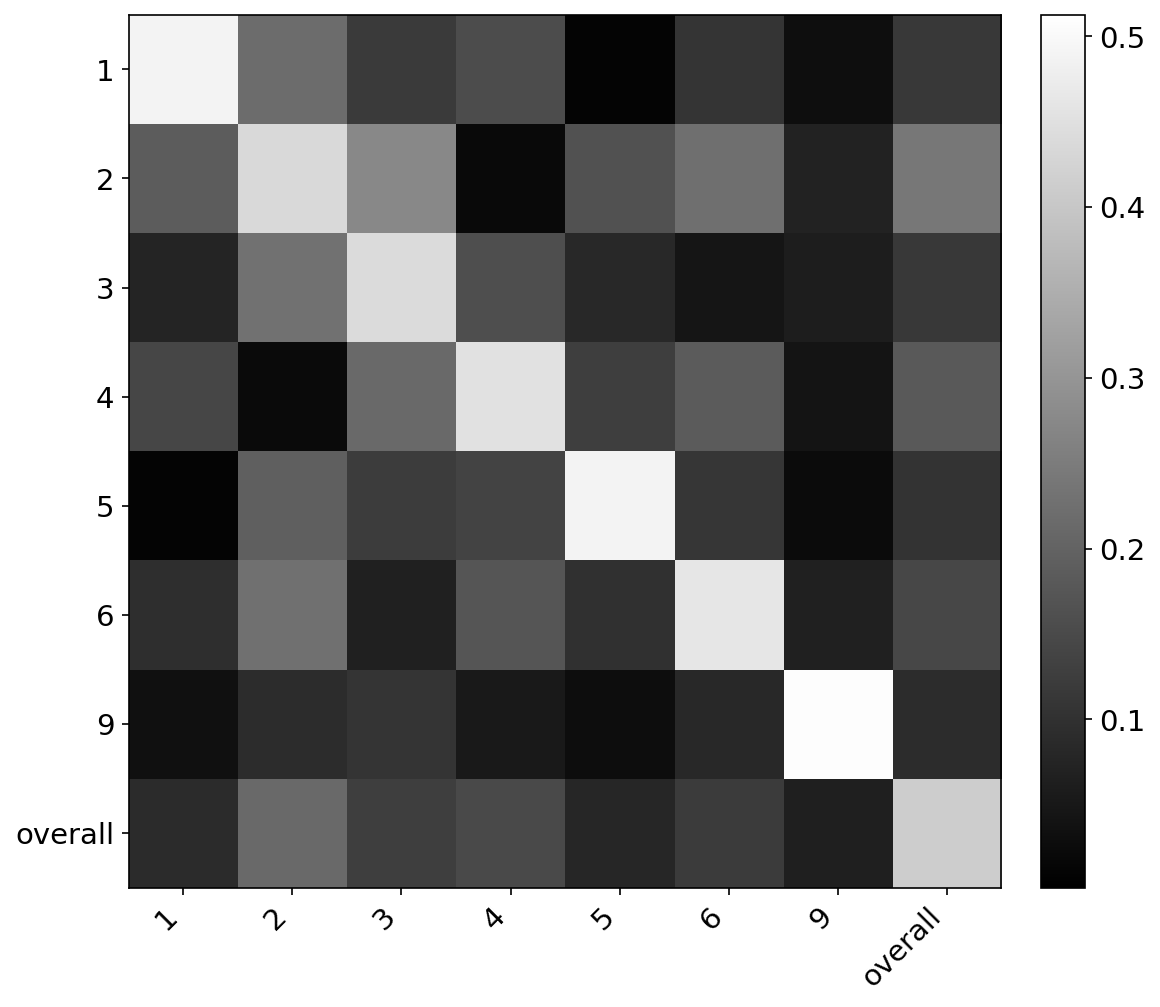}
                \caption{Normal Labels}
            \end{subfigure}
            \caption{Pairwise KL Divergence Heatmaps for Modified Alexnet on Tiny ImageNet with CIFAR10 labels}
            \label{fig:cifar10_results_OOD_ticifar10}
        \end{figure}

        The OOD results are the same as those seen on ImageNet, with modest increases in in-class separation but a much more pronounced increase in the entropy of the classes. This is shown in figure \ref{fig:cifar10_results_OOD_ticifar10}. The inter-class distances are much smaller than those for the in-distribution data, and many classes are very close to each other, indicating reduced separation under our class-conditional diagnostic.

\begin{figure}[hbtp]
            \centering
            \begin{subfigure}{0.32\textwidth}
                \includegraphics[width=\linewidth]{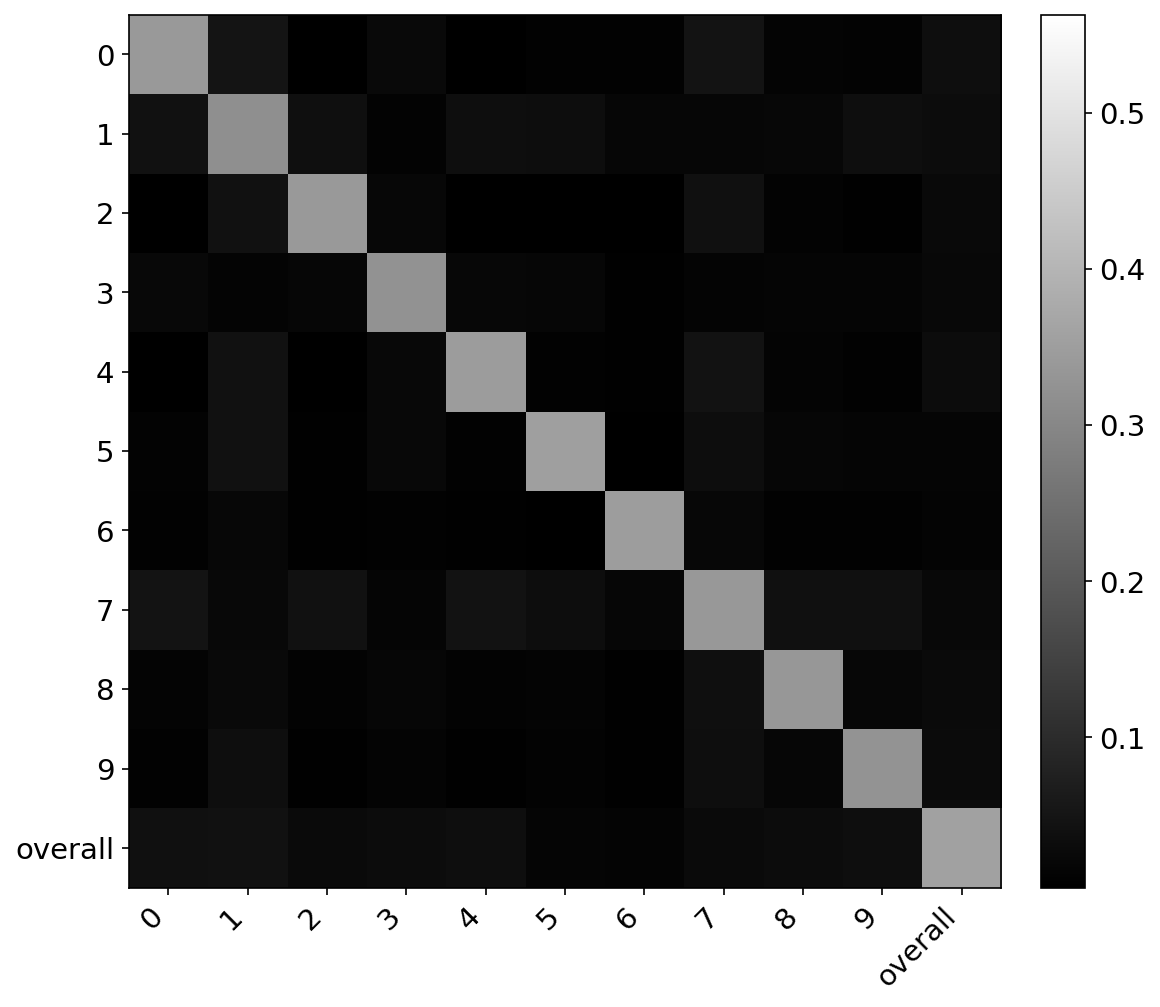}
                \caption{Initialisation}
            \end{subfigure}
            \begin{subfigure}{0.32\textwidth}
                \includegraphics[width=\linewidth]{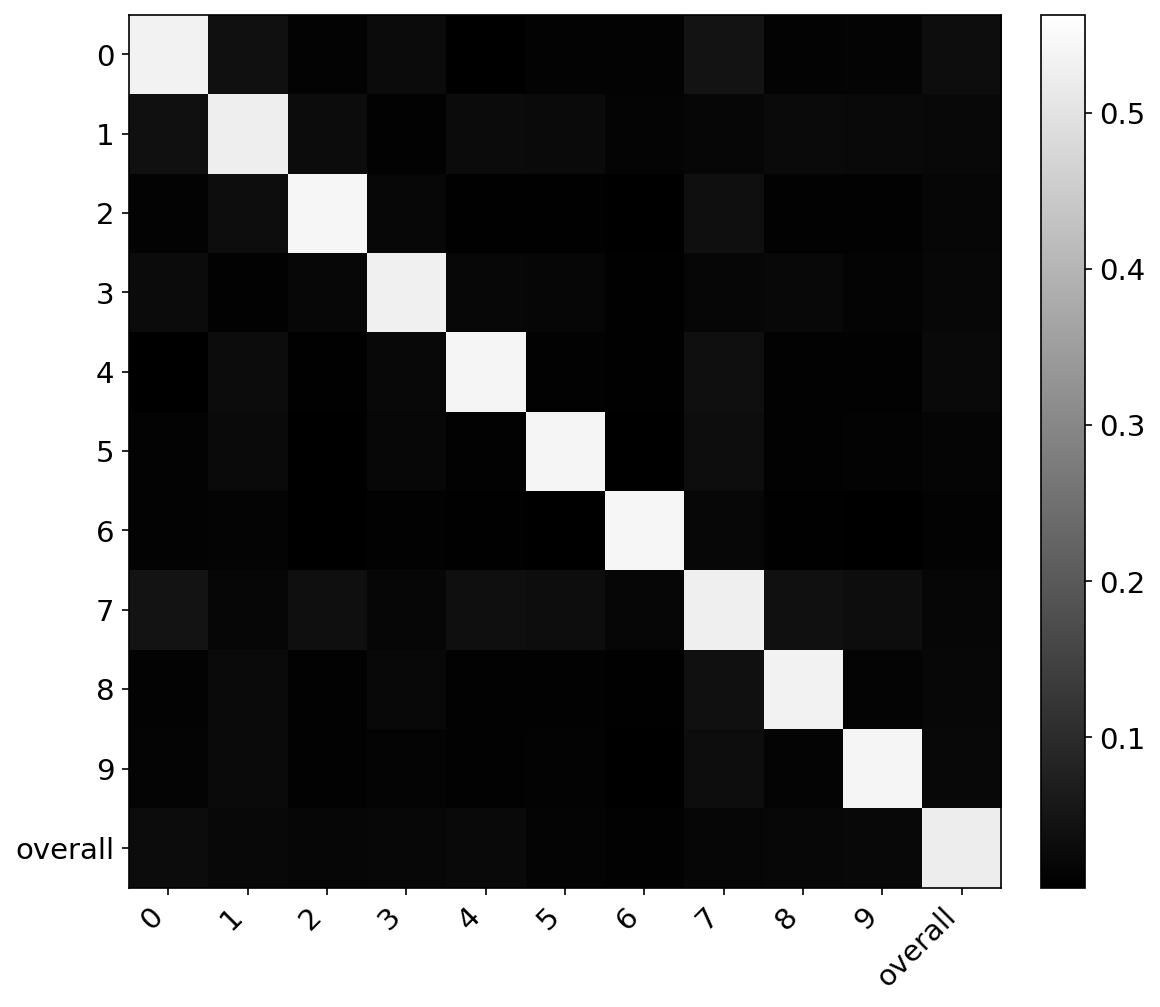}
                \caption{Random Labels}
            \end{subfigure}
            \begin{subfigure}{0.32\textwidth}
                \includegraphics[width=\linewidth]{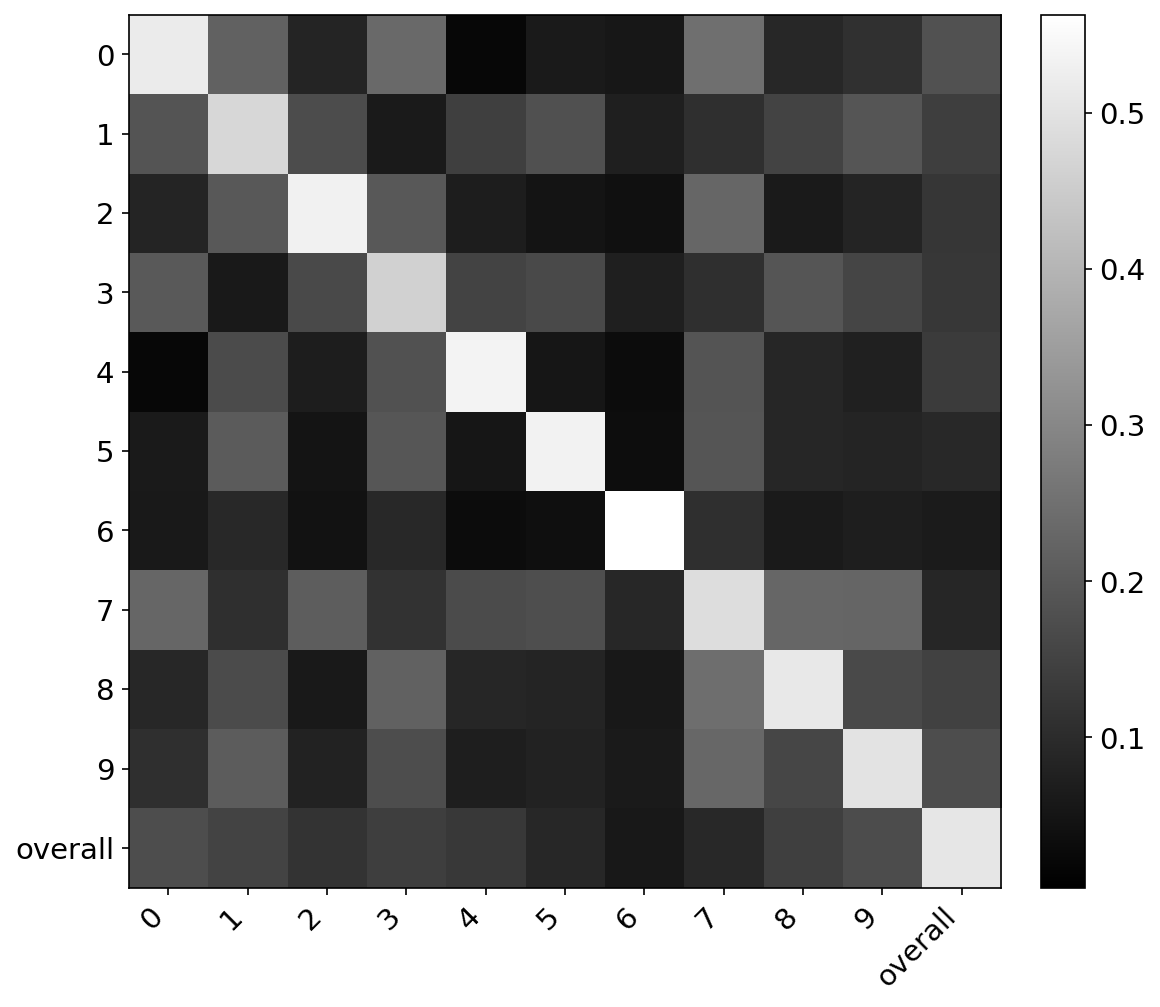}
                \caption{Normal Labels}
            \end{subfigure}
            \caption{Pairwise KL divergence heatmaps for Alexnet on CIFAR10 at the second-to-last layer}
            \label{fig:alexnet_fc2_kl}
        \end{figure}

\begin{figure}[hbtp]
            \centering
            \begin{subfigure}{0.32\textwidth}
                \includegraphics[width=\linewidth]{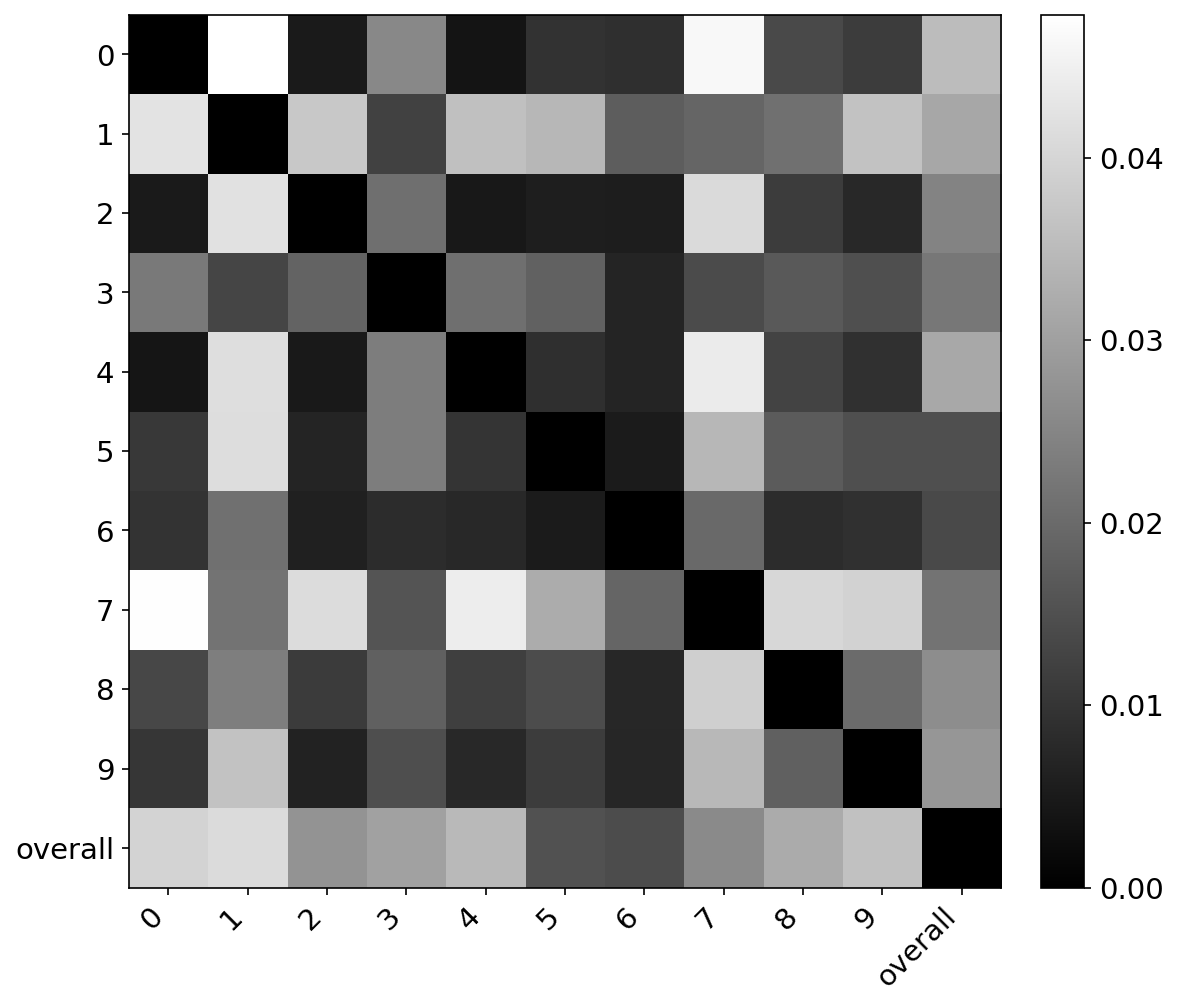}
                \caption{Initialisation}
            \end{subfigure}
            \begin{subfigure}{0.32\textwidth}
                \includegraphics[width=\linewidth]{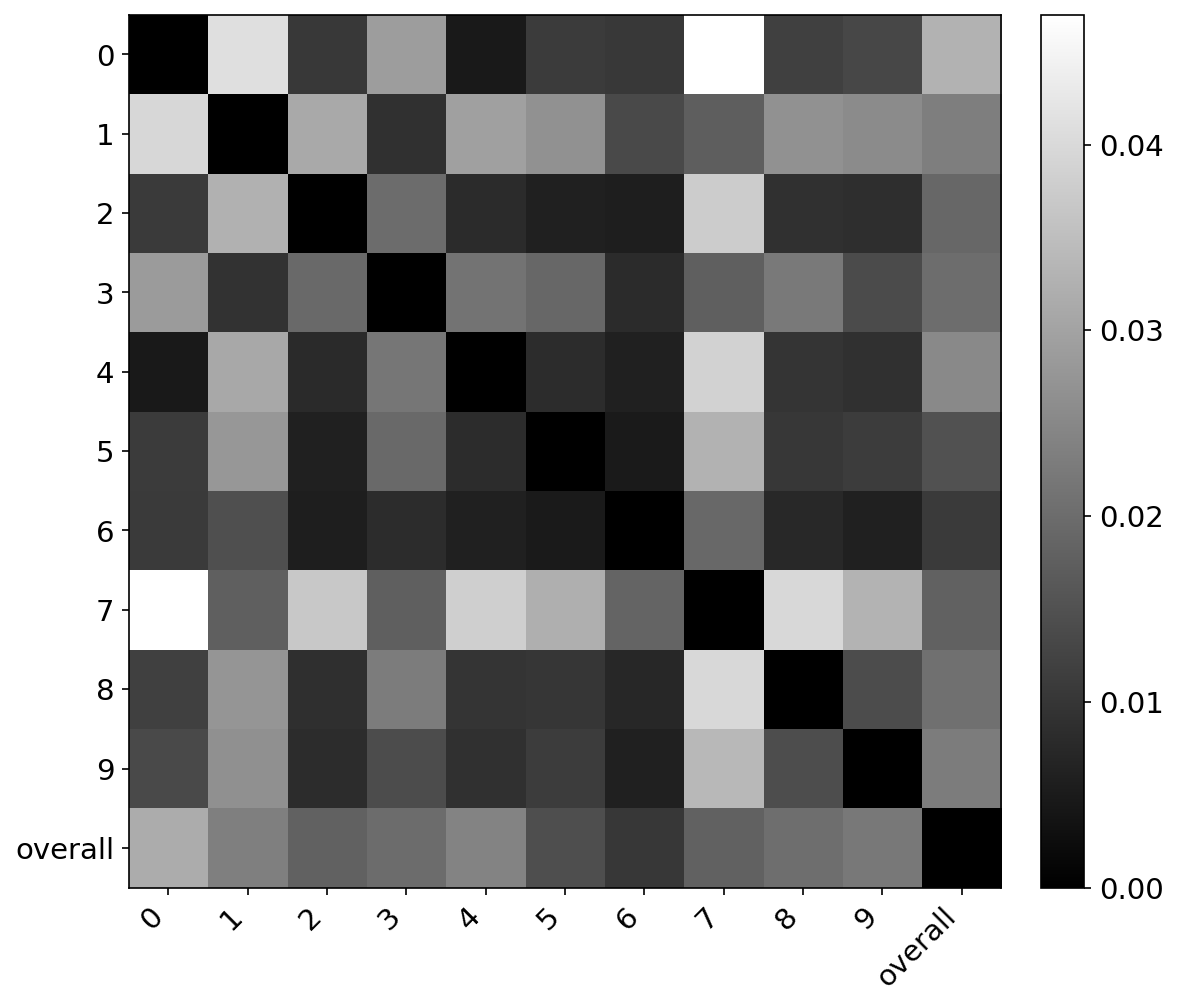}
                \caption{Random Labels}
            \end{subfigure}
            \begin{subfigure}{0.32\textwidth}
                \includegraphics[width=\linewidth]{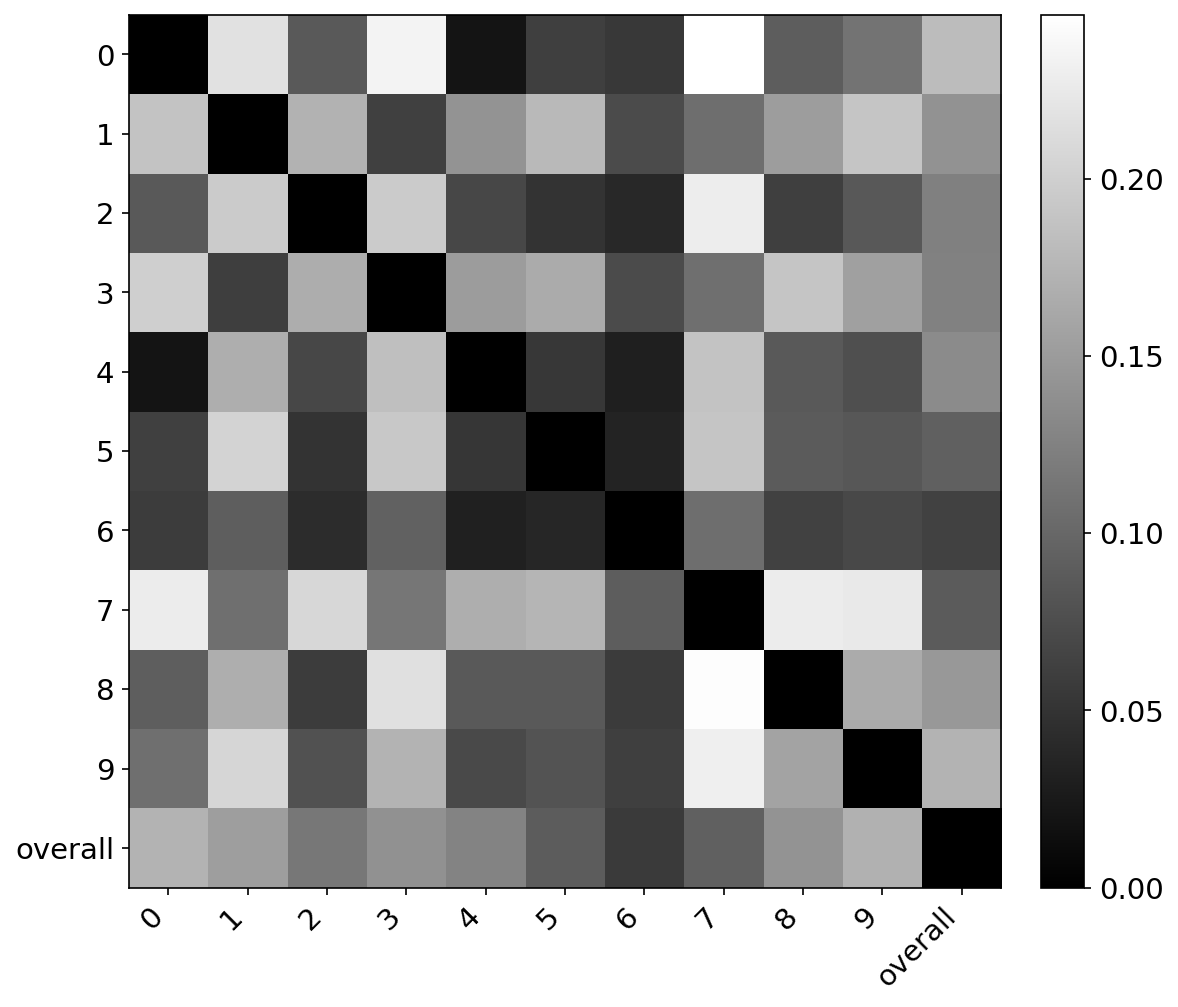}
                \caption{Normal Labels}
            \end{subfigure}
            \caption{Unnormalised Pairwise KL divergence heatmaps for Alexnet on CIFAR10 at the second-to-last layer}
            \label{fig:alexnet_fc2_kl_unnorm}
        \end{figure}

        Since the second-to-last layer of Alexnet is also fully connected, we can apply the same methodology to it as well. The results are shown in Figure \ref{fig:alexnet_fc2_kl} and the unnormalised results are shown in \ref{fig:alexnet_fc2_kl_unnorm}. Here, we can see the same trend with the inter-class distances increasing, however we also see that the representations are still highly entropic. Since this is not the final representation of the network, this itself may not reflect non-robustness. Additionally, looking at the sparsity of significant activations, we see an interesting trend where the random labels have the most sparse activation pattern, while the network at initialisation is fairly dense and the the converged network is somewhere in between. We believe that this is indicative of a certain goldilocks zone of sparsity, where too much sparsity with an absence of paths that are always significant indicates memorisation, since there is no 'grandmother' encoding semantics. On the other hand, too little sparsity indicates a lack of discriminative features being learned. Well-separated paths would also indicate that the last layer is unnecessary, since the model can discriminate between the classes even without it. 

        The same scalar summaries for the second-to-last layer are shown in Table \ref{tab:cifar10_Bernoulli_summary}, where the trained model still has larger inter-class KL than untrained and random-label checkpoints, but the magnitudes are lower than at the final layer.

\begin{figure}[hbtp]
            \centering
            \includegraphics[width=\linewidth]{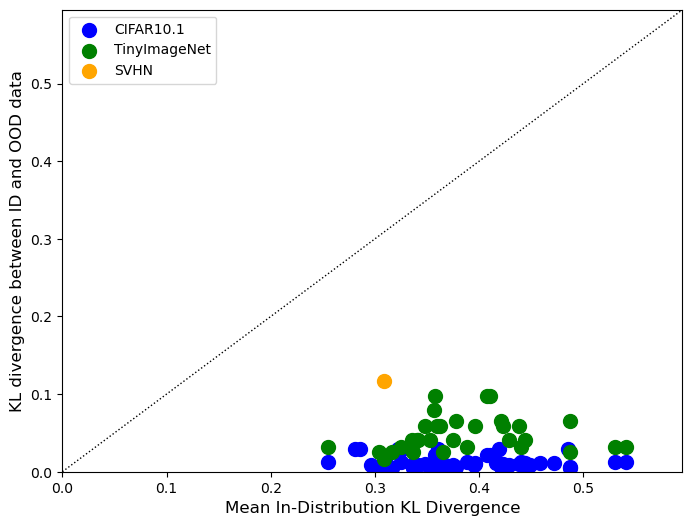}
            \caption{KL divergence for class distributions between in-distribution and OOD data for Small Alexnet}
            \label{fig:in_out_distances_cifar10}
        \end{figure}

        The pattern with in-distribution vs out-of-distribution data is seen in Figure \ref{fig:in_out_distances_cifar10}. CIFAR10.1 is expected to have a small distance from the in-distribution data as it is supposed to be a proxy for the test set itself. The same cannot be said for the classes from TinyImageNet and SVHN, which are expected to be much further away. However, we see that the distances between the different datasets are all quite small, with the OOD datasets being only slightly further away from the in-distribution data than CIFAR10.1. This suggests that these class-conditional path statistics are better interpreted as diagnostics of representation shift than as a standalone OOD detector.

        \section{Heatmaps and Histograms for ImageNet}
\begin{figure}[hbtp]
                \centering
                \begin{subfigure}{0.48\textwidth}
                    \includegraphics[width=\linewidth]{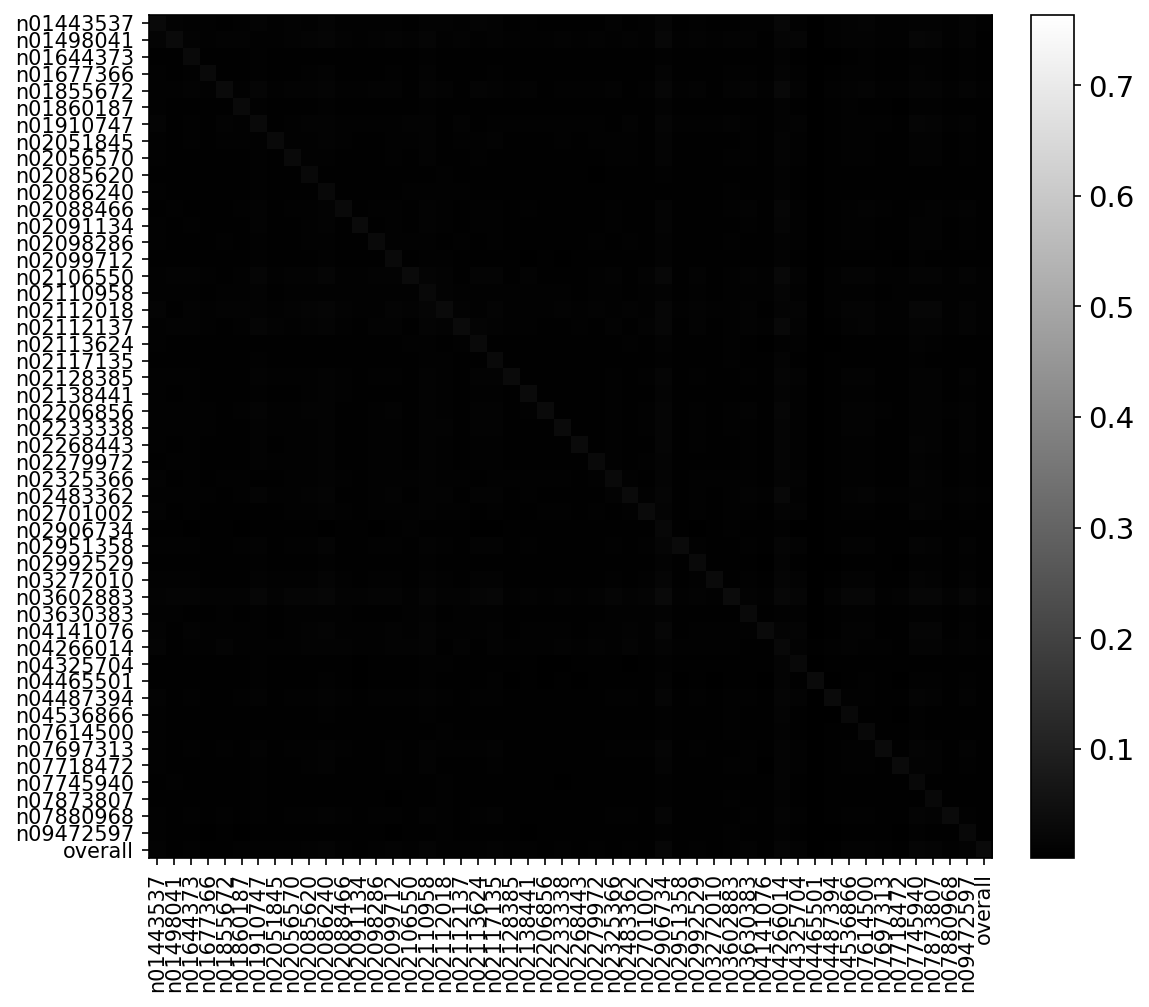}
                    \caption{Initialisation}
                \end{subfigure}
                \begin{subfigure}{0.48\textwidth}
                    \includegraphics[width=\linewidth]{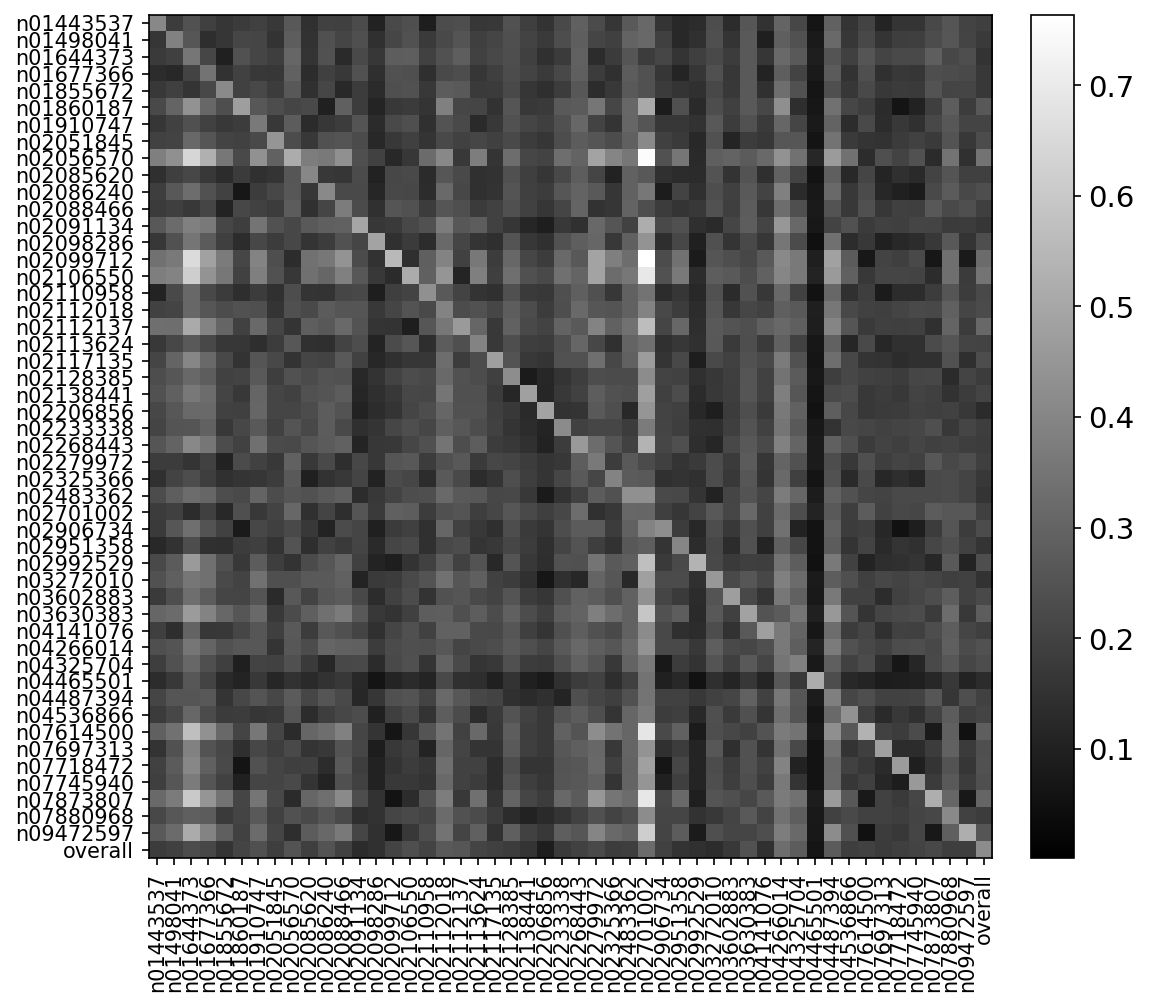}
                    \caption{Normal Labels}
                \end{subfigure}
                \caption{Pairwise KL Divergence Heatmaps for ResNet50 on ImageNet}
                \label{fig:imagenet_results_resnet50}
            \end{figure}

\begin{figure}[hbtp]
                \centering
                \begin{subfigure}{0.48\textwidth}
                    \includegraphics[width=\linewidth]{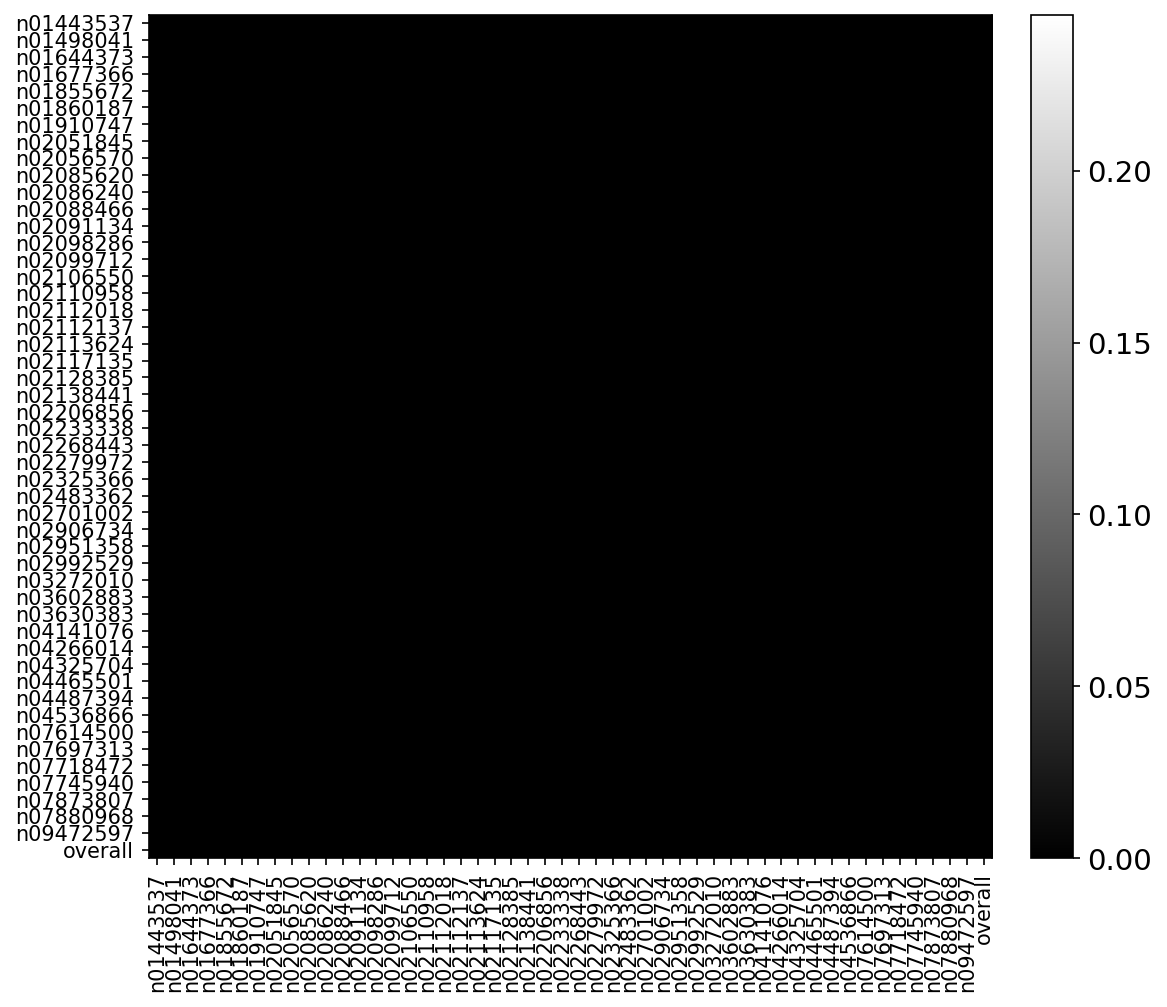}
                    \caption{Initialisation}
                \end{subfigure}
                \begin{subfigure}{0.48\textwidth}
                    \includegraphics[width=\linewidth]{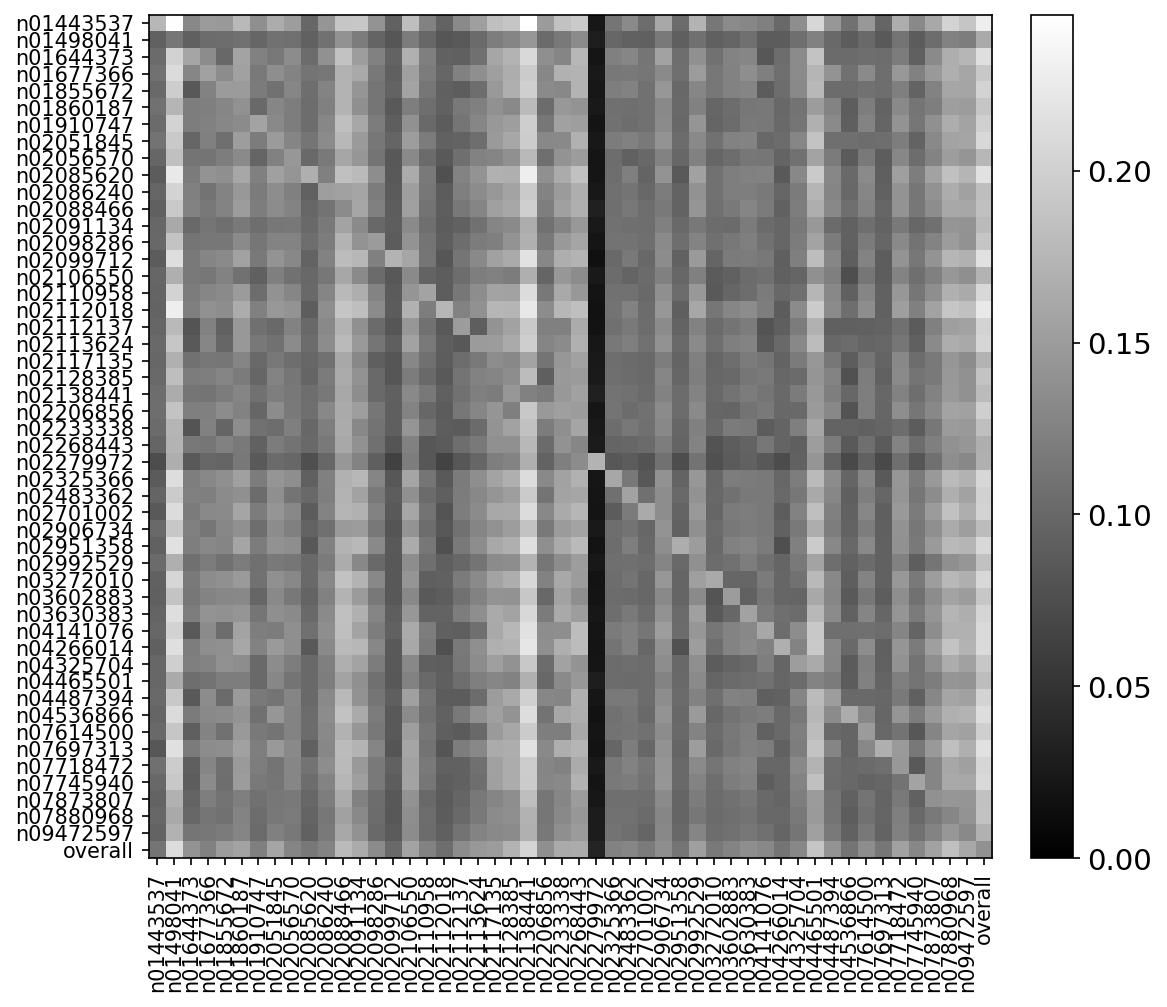}
                    \caption{Normal Labels}
                \end{subfigure}
                \caption{Pairwise KL Divergence Heatmaps for ViT/B-32 on ImageNet}
                \label{fig:imagenet_results_vit_b_32}
            \end{figure}

\begin{figure}[hbtp]
                \centering
                \begin{subfigure}{0.32\textwidth}
                    \includegraphics[width=\linewidth]{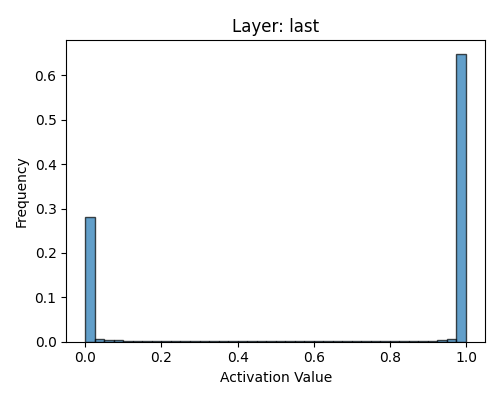}
                    \caption{Initialisation}
                \end{subfigure}
                \begin{subfigure}{0.32\textwidth}
                    \includegraphics[width=\linewidth]{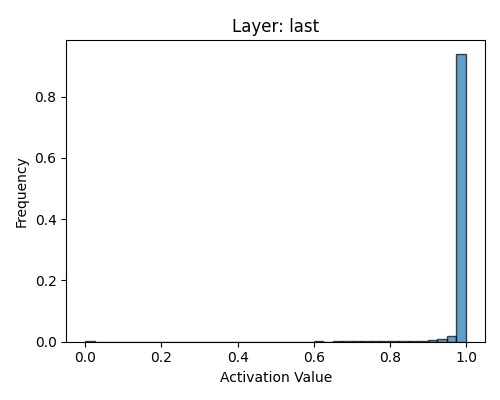}
                    \caption{Random Labels}
                \end{subfigure}
                \begin{subfigure}{0.32\textwidth}
                    \includegraphics[width=\linewidth]{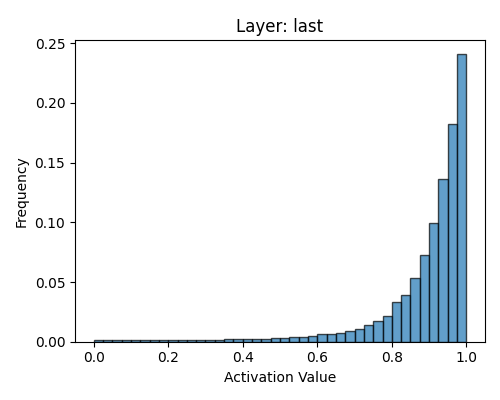}
                    \caption{Normal Labels}
                \end{subfigure}
                \caption{Activation Sparsity Histograms for InceptionV3 on ImageNet-r for the class 'cheeseburger' (n07697313)}
                \label{fig:inceptionv3_sparsity_cheeseburger_OOD}
            \end{figure}

\begin{figure}[hbtp]
                \centering
                \begin{subfigure}{0.32\textwidth}
                    \includegraphics[width=\linewidth]{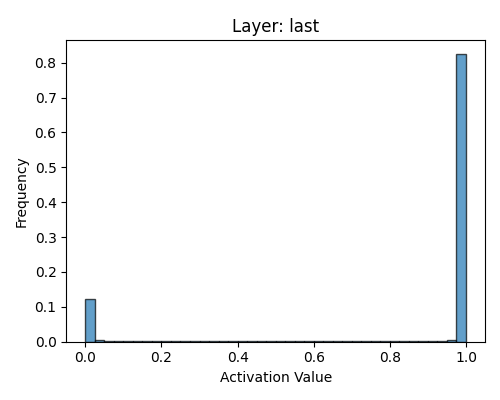}
                    \caption{Initialisation}
                \end{subfigure}
                \begin{subfigure}{0.32\textwidth}
                    \includegraphics[width=\linewidth]{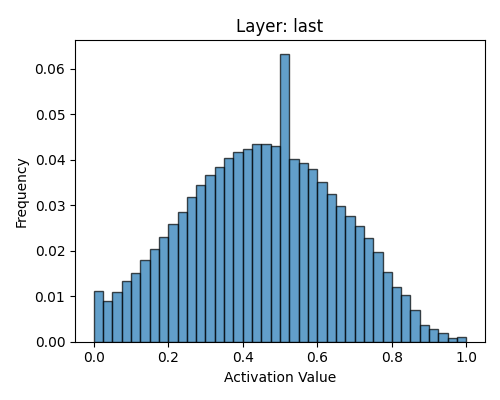}
                    \caption{Normal Labels}
                \end{subfigure}
                \caption{Activation Sparsity Histograms for Resnet50 on ImageNet for the class 'king penguin' (n02056570)}
                \label{fig:resnet50_sparsity_king_penguin_OOD}
            \end{figure}

\begin{figure}[hbtp]
                \centering
                \begin{subfigure}{0.32\textwidth}
                    \includegraphics[width=\linewidth]{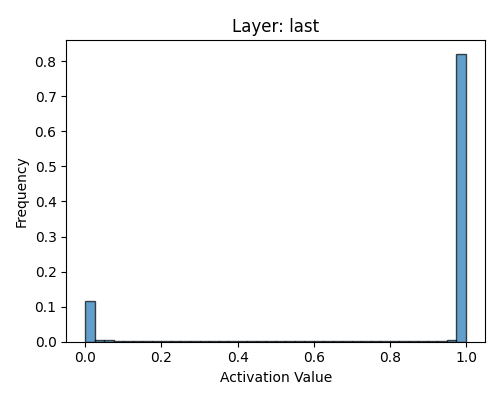}
                    \caption{Initialisation}
                \end{subfigure}
                \begin{subfigure}{0.32\textwidth}
                    \includegraphics[width=\linewidth]{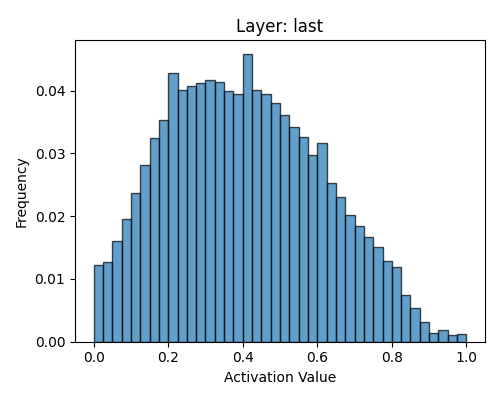}
                    \caption{Normal Labels}
                \end{subfigure}
                \caption{Activation Sparsity Histograms for Resnet50 on ImageNet for the class 'cheeseburger' (n07697313)}
                \label{fig:resnet50_sparsity_cheeseburger_OOD}
            \end{figure}

            We put a collections on the ablations in Figures \ref{fig:imagenet_results_resnet50}, \ref{fig:imagenet_results_vit_b_32}, \ref{fig:inceptionv3_sparsity_cheeseburger_OOD}, \ref{fig:resnet50_sparsity_king_penguin_OOD} and \ref{fig:resnet50_sparsity_cheeseburger_OOD}. The same trends are observed as in the main results, with the random labels case being almost identical to the initialised case, and the normal labels case showing increased inter-class distances and reduced entropy. The OOD results show a much more pronounced increase in entropy, with only modest increases in inter-class distances.

        \section{Sparsity Results on ImageNet}

            \begin{figure}[hbtp]
                \centering
                \begin{subfigure}{0.32\textwidth}
                    \includegraphics[width=\linewidth]{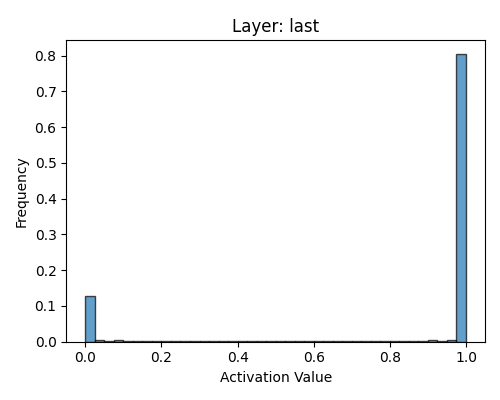}
                    \caption{Initialisation}
                \end{subfigure}
                \begin{subfigure}{0.32\textwidth}
                    \includegraphics[width=\linewidth]{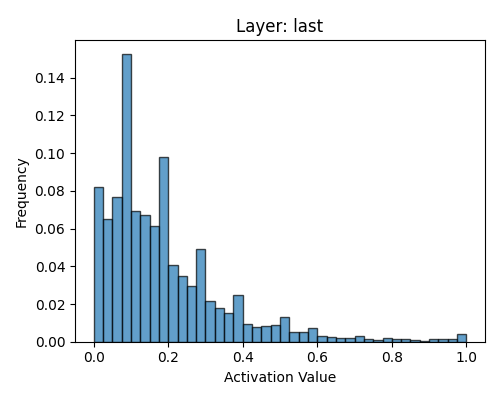}
                    \caption{Normal Labels}
                \end{subfigure}
                \caption{Activation Sparsity Histograms for Resnet50 on ImageNet for the class king penguin (n02056570)}
                \label{fig:resnet50_sparsity_king_penguin}
            \end{figure}
            
            \begin{figure}[hbtp]
                \centering
                \begin{subfigure}{0.32\textwidth}
                    \includegraphics[width=\linewidth]{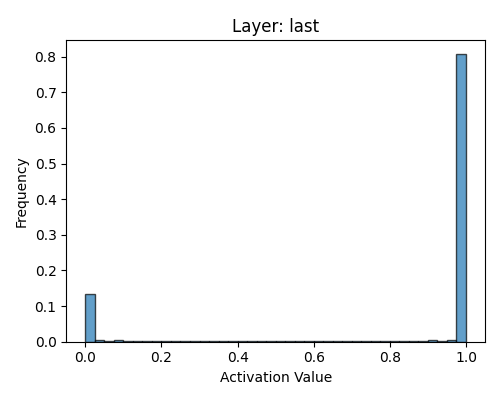}
                    \caption{Initialisation}
                \end{subfigure}
                \begin{subfigure}{0.32\textwidth}
                    \includegraphics[width=\linewidth]{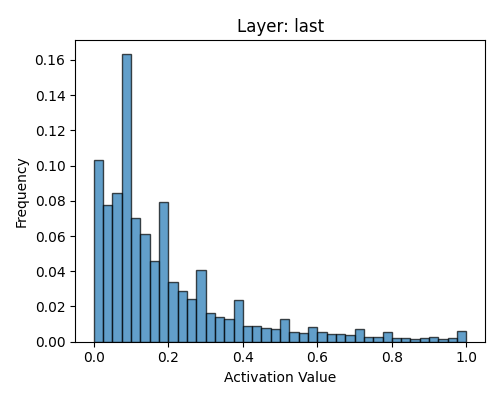}
                    \caption{Normal Labels}
                \end{subfigure}
                \caption{Activation Sparsity Histograms for Resnet50 on ImageNet for the class 'cheeseburger' (n07697313)}
                \label{fig:resnet50_sparsity_cheeseburger}
            \end{figure}

            \begin{figure}[hbtp]
                \centering
                \begin{subfigure}{0.32\textwidth}
                    \includegraphics[width=\linewidth]{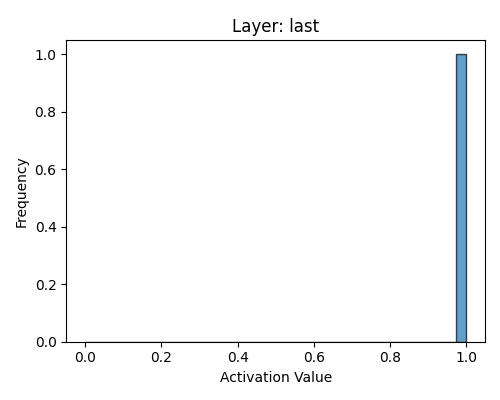}
                    \caption{Initialisation}
                \end{subfigure}
                \begin{subfigure}{0.32\textwidth}
                    \includegraphics[width=\linewidth]{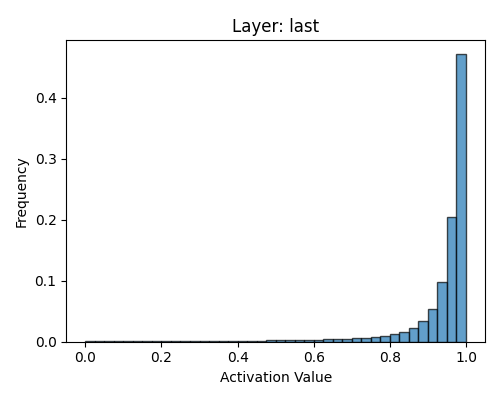}
                    \caption{Normal Labels}
                \end{subfigure}
                \caption{Activation Sparsity Histograms for ViT-B/32 ImageNet for the class 'king penguin' (n02056570)}
                \label{fig:vit_b_32_sparsity_king_penguin_OOD}
            \end{figure}
            
            \begin{figure}[hbtp]
                \centering
                \begin{subfigure}{0.32\textwidth}
                    \includegraphics[width=\linewidth]{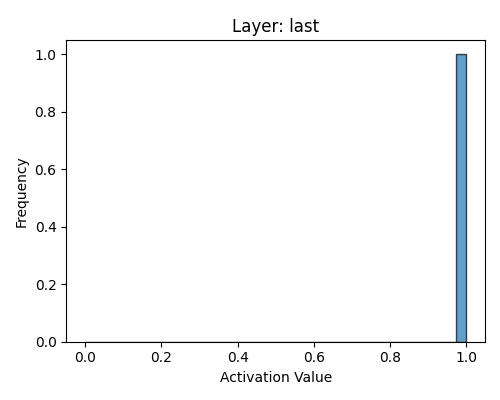}
                    \caption{Initialisation}
                \end{subfigure}
                \begin{subfigure}{0.32\textwidth}
                    \includegraphics[width=\linewidth]{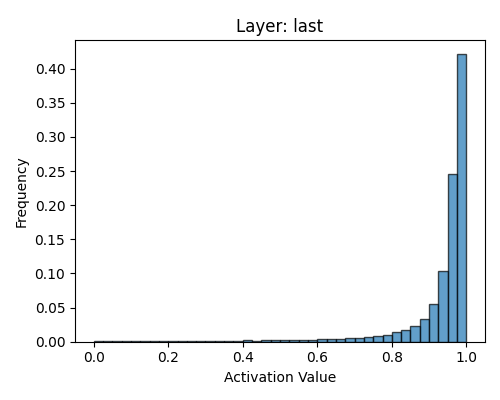}
                    \caption{Normal Labels}
                \end{subfigure}
                \caption{Activation Sparsity Histograms for ViT-B/32 ImageNet for the class 'cheeseburger' (n07697313)}
                \label{fig:vit_b_32_sparsity_cheeseburger_OOD}
            \end{figure}
    
            \begin{figure}[hbtp]
                \centering
                \begin{subfigure}{0.32\textwidth}
                    \includegraphics[width=\linewidth]{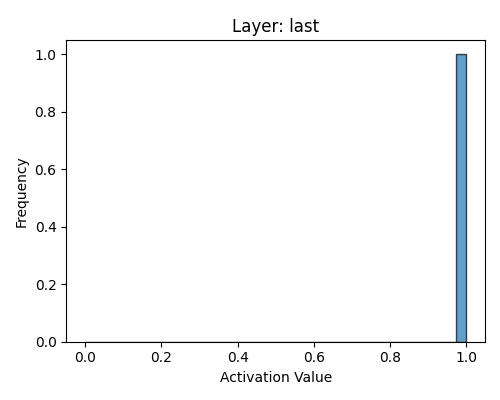}
                    \caption{Initialisation}
                \end{subfigure}
                \begin{subfigure}{0.32\textwidth}
                    \includegraphics[width=\linewidth]{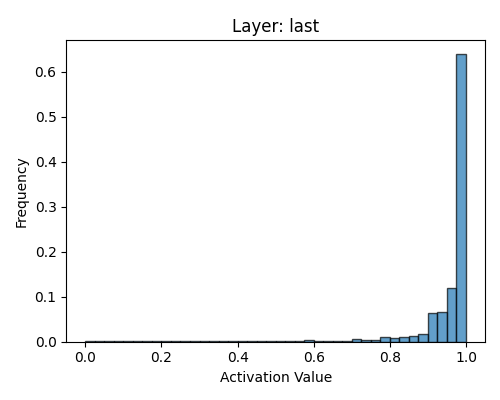}
                    \caption{Normal Labels}
                \end{subfigure}
                \caption{Activation Sparsity Histograms for ViT-B/32 ImageNet for the class king penguin (n02056570)}
                \label{fig:vit_b_32_sparsity_king_penguin}
            \end{figure}
            
            \begin{figure}[hbtp]
                \centering
                \begin{subfigure}{0.32\textwidth}
                    \includegraphics[width=\linewidth]{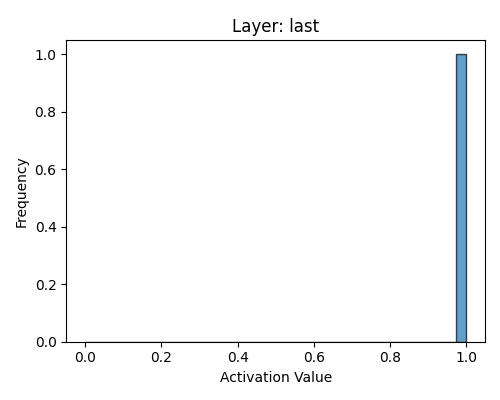}
                    \caption{Initialisation}
                \end{subfigure}
                \begin{subfigure}{0.32\textwidth}
                    \includegraphics[width=\linewidth]{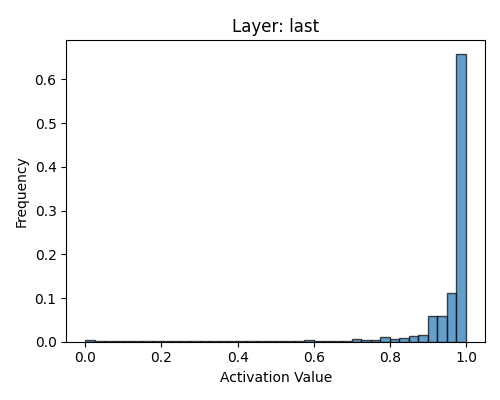}
                    \caption{Normal Labels}
                \end{subfigure}
                \caption{Activation Sparsity Histograms for ViT-B/32 ImageNet for the class 'cheeseburger' (n07697313)}
                \label{fig:vit_b_32_sparsity_cheeseburger}
            \end{figure}
\end{document}